\documentclass[twoside,11pt]{article}

\usepackage{jmlr2e}

\usepackage{amsmath,amsfonts,bm,bbm}
\usepackage{color}
\usepackage{hyperref,url}
\usepackage{algorithm,algorithmic,multirow,hhline}
\usepackage{subfigure} 
\input{mysymbol.sty}
\newcommand{\QED}{\hfill\ensuremath{\blacksquare}}

\newcommand{\closure}[2][3]{{}\mkern#1mu\overline{\mkern-#1mu#2}}
\newtheorem{assumption}{\hspace{0pt}\bf Assumption}
\jmlrheading{VV}{20YY}{PPP-PPP}{11/2016}{MM/20YY}{Alec Koppel, Garrett Warnell, Ethan Stump and Alejandro Ribeiro}
\newcommand{\INDSTATE}[1][1]{\STATE\hspace{3mm}}
\newcommand{\INDSTATED}[1][1]{\STATE\hspace{6mm}}
\ShortHeadings{Parsimonious Online Learning with Kernels}{Koppel, Warnell, Stump, and Ribeiro.}
\firstpageno{1}

\begin{document}

\title{Parsimonious Online Learning with Kernels \\via Sparse Projections in Function Space}
\author{\name Alec Koppel \email akoppel@seas.upenn.edu \\
       \addr Department of Electrical and Systems Engineering\\
       University of Pennsylvania\\
       Philadelphia, PA 19104, USA
       \AND
       \name Garrett Warnell \email garrett.a.warnell.civ@mail.mil \\
       \addr Computational and Information Sciences Directorate\\
       U.S. Army Research Laboratory\\
       Adelphi, MD 20783, USA
       \AND
       \name Ethan Stump \email ethan.a.stump2.civ@mail.mil\\
       \addr Computational and Information Sciences Directorate\\
       U.S. Army Research Laboratory\\
       Adelphi, MD 20783, USA
       \AND
       \name Alejandro Ribeiro \email aribeiro@seas.upenn.edu \\
       \addr Department of Electrical and Systems Engineering\\
       University of Pennsylvania\\
       Philadelphia, PA 19104, USA
       }
\editor{}
\maketitle

\begin{abstract}%
%
%
%
Despite their attractiveness, popular perception is that techniques for nonparametric function approximation do not scale to streaming data due to an intractable growth in the amount of storage they require. 
%
%
To solve this problem in a memory-affordable way, we propose an online technique based on functional stochastic gradient descent in tandem with supervised sparsification based on greedy function subspace projections.
The method, called \emph{parsimonious online learning with kernels} (POLK), provides a controllable tradeoff between its solution accuracy and the amount of memory it requires.
We derive conditions under which the generated function sequence converges almost surely to the optimal function, and we establish that the memory requirement remains finite.
%
%
%
We evaluate POLK for kernel multi-class logistic regression and kernel hinge-loss classification on three canonical data sets: a synthetic Gaussian mixture model, the MNIST hand-written digits, and the Brodatz texture database.
On all three tasks, we observe a favorable trade-off of objective function evaluation, classification performance, and complexity of the nonparametric regressor extracted the proposed method.
\end{abstract}


\section{Introduction}\label{sec:intro}

Reproducing kernel Hilbert spaces (RKHS) provide the ability to approximate functions using nonparametric functional representations. Although the structure of the space is determined by the choice of kernel, the set of functions that can be represented is still sufficiently rich so as to permit close approximation of a large class of functions. This resulting expressive power makes RKHS an appealing choice in many learning problems where we want to estimate an unknown function that is specified as optimal with respect to some empirical risk. When learning these optimal function representations in a RKHS, the representer theorem is used to transform the search over functions into a search over parameters, where the number of parameters grows with each new observation that is processed \citep{wheeden1977measure,norkin2009stochastic}. This growth is what endows the representation with expressive power. However, this growth also results in function descriptions that are as complex as the number of processed observations, and, more importantly, in training algorithms that exhibit a cost per iteration that grows with each new iterate \citep{Pontil05erroranalysis,Kivinen2004}. The resulting unmanageable training cost renders RKHS learning approaches inefficient for large data sets and outright inapplicable in streaming applications. This is a well-known limitation which has motivated the development of several heuristics to reduce the growth in complexity. These heuristics typically result in suboptimal functional approximations \citep{richard2009online}.

This paper proposes a new technique for learning nonparametric function approximations in a RKHS that respects optimality and ameliorates the complexity issues described above. We accomplish this by: (i) shifting the goal from that of finding an approximation that is optimal to that of finding an approximation that is optimal within a class of parsimonious (sparse) kernel representations; (ii) designing a training method that follows a trajectory of intermediate representations that are also parsimonious. The proposed technique, \emph{parsimonious online learning with kernels} (POLK), provides a controllable tradeoff between complexity and optimality and we provide theoretical guarantees that neither factor becomes untenable.

Formally, we propose solving expected risk minimization problems, where the goal is to learn a regressor that minimizes a loss function quantifying the merit of a statistical model averaged over a data set. We focus on the case when the number of training examples, $N$, is either very large, or the training examples arrive sequentially. Further, we assume that these input-output examples, $(\bbx_n, \bby_n)$, are i.i.d. realizations drawn from a stationary joint distribution over the random pair $(\bbx, \bby) \in \ccalX \times \ccalY$. This problem class is popular in many fields and particularly ubiquitous in text \citep{Sampson:1990:NLA:104905.104911}, image \citep{Mairal2007}, and genomic \citep{tacsan2014selecting} processing. Here, we consider finding regressors that are not vector valued parameters, but rather functions $f \in \ccalH$ in a hypothesized function class $\ccalH$. This function estimation task allows one to learn nonlinear statistical models and is known to yield better results in applications where linearity of a given statistical model is overly restrictive such as computer vision and object recognition \citep{mukherjee1996automatic,li2014kernel}. The adequacy of the regressor function $f$ is evaluated by the convex loss function $\ell:\ccalH \times \ccalX \times \ccalY \rightarrow \reals$ that quantifies the merit of the estimator $f(\bbx)$ evaluated at feature vector $\bbx$. This loss is averaged over all possible training examples to define the statistical loss $L(f) : = \mbE_{\bbx, \bby}{[ \ell(f(\bbx), y)}]$, which we combine with  a Tikhonov regularizer to construct the regularized loss $R(f) : = \argmin_{f \in \ccalH} L(f) +(\lambda/2)\|f \|^2_{\ccalH}$ \citep{shalev2010learnability, evgeniou2000regularization}.
We then define the optimal function as
\begin{align}\label{eq:kernel_stoch_opt}
f^*=\argmin_{f \in \ccalH} R(f) :& = \argmin_{f \in \ccalH}\mbE_{\bbx, \bby}\Big[ \ell(f\big(\bbx), y\big)\Big] +\frac{\lambda}{2}\|f \|^2_{\ccalH} 
\end{align}
The optimization problem in \eqref{eq:kernel_stoch_opt} is intractable in general. 
 However, when $\ccalH$ is equipped with a \emph{reproducing kernel} $\kappa : \ccalX \times \ccalX \rightarrow \reals$, a nonparametric function estimation problem of the form \eqref{eq:kernel_stoch_opt} may be reduced to a parametric form via the representer theorem \citep{wheeden1977measure,norkin2009stochastic}.
This theorem states that the optimal argument of \eqref{eq:kernel_stoch_opt} is in the span of kernel functions that are centered at points in the given training data set, and it reduces the problem to that of determining the $N$ coefficients of the resulting linear combination of kernels (Section \ref{sec:problem}). This results in a function description that is data driven and flexible, alas very complex. As we consider problems with larger training sets, the representation of $f$ requires a growing number of kernels \citep{norkin2009stochastic}. In the case of streaming applications this number would grow unbounded and the kernel matrix as well as the coefficient vector would grow to infinite dimension and an infinite amount of memory would be required to represent $f$. It is therefore customary to reduce this complexity by forgetting training points or otherwise requiring that $f^*$ admit a parsimonious representation in terms of a sparse subset of kernels. This overcomes the difficulties associated with a representation of unmanageable complexity but a steeper difficulty is the {\it determination} of this optimal parsimonious representation as we explain in the following section

\subsection{Context}

To understand the challenge in determining optimal parsimonious representations, recall that kernel optimization methods borrow techniques from vector valued (i.e., without the use of kernels) stochastic optimization in the sense that they seek to optimize \eqref{eq:kernel_stoch_opt} by replacing the descent direction of the objective with that of a stochastic estimate \citep{Bottou1998,Robbins1951}.
Stochastic optimization is well understood in vector valued problems to the extent that recent efforts are concerned with improving convergence properties through the use of variance reduction \citep{schmidt2013minimizing, johnson2013accelerating, defazio2014saga}, or stochastic approximations of Newton steps \citep{schraudolph2007stochastic, bordes2009sgd, mokhtari2014res, mokhtari2014global}.
Stochastic optimization in kernel spaces, however, exhibits two peculiarities that make it more challenging:

\begin{enumerate}

\item The implementation of stochastic methods for expected risk minimization in a RKHS requires storage of kernel matrices and weight vectors that together are cubic in the iteration index. This is true even if we require that the solution $f^*$  admit a sparse representation because, while it may be true that the asymptotic solution admits a sparse representation, the intermediate iterates are not necessarily sparse; see, e.g., \cite{Kivinen2004}. \label{list1}

\item The problem in (\ref{list1}) makes it necessary to use sparse approximations of descent directions. However, these sparse approximates are not guaranteed to be valid stochastic descent directions. Consequently, there are no guarantees that a path of sparse approximation learns the optimal sparse approximation. \label{list2}

\end{enumerate}

Issue \eqref{list1} is a key point of departure between kernel stochastic optimization and its vector valued counterpart. It implies that redefining $f^*$ to encourage sparsity may make it easier to work with the RKHS representation after it has been learnt. However, the stochastic gradients that need to be computed to find such representation have a complexity that grows with the order of the iteration index \citep{Kivinen2004}. Works on stochastic optimization in a RKHS have variously ignored the intractable growth of the parametric representation of $f\in \ccalH$ \citep{1715525, liu2008kernel, Pontil05erroranalysis, dieuleveut2014non}, or have augmented the learned function to limit the memory issues associated with kernelization using online sparsification procedures. These approaches focus on limiting the growth of the kernel dictionary through the use of forgetting factors \citep{Kivinen2004}, random dropping \citep{zhang2013online}, and compressive sensing techniques \citep{1315946, richard2009online,Honeine2012}. These approaches overcome Issue (\ref{list1}) but they do so at the cost of dropping optimality [cf. Issue (\ref{list2})]. This is because these sparsification techniques introduce a bias in the stochastic gradient which nullifies convergence guarantees. 

Past works that have considered \emph{supervised} sparsification (addressing issues \eqref{list1}-\eqref{list2}) have only been developed for special cases such as online support vector machines (SVM) \citep{wang2012breaking}, off-line logistic regression \cite{Zhu2005}, and off-line SVM \citep{joachims2009sparse}. The works perhaps most similar to ours, but developed only for SVM \citep{wang2012breaking}) fixes the number of kernel dictionary elements, or the model order, in advance rather tuning the kernel dictionary to guarantee stochastic descent, i.e. determining which kernel dictionary elements are most important for representing $f^*$. Further, the analysis of the resulting bias induced by sparsification requires overly restrictive assumptions and is conducted in terms of time-average objective sub-optimality, a looser criterion than almost sure convergence.
For specialized classes of loss functions, the bias of the descent direction induced by unsupervised sparsification techniques using random sub-sampling does not prevent the derivation of bounds on the time-average sub-optimality (regret) \citep{zhang2013online}; however, this analysis omits important cases such as support vector machines and kernel ridge regression.

\subsection{Contributions}

In this work, we build upon past works which combine functional generalizations of first-order stochastic optimization methods operating in tandem with supervised sparsification. In particular, descending along the gradient of the objective in \eqref{eq:kernel_stoch_opt} is intractable when the sample size $N$ is not necessarily finite, and thus stochastic methods are necessary. In Section \ref{sec:algorithm}, we build upon \cite{Kivinen2004} in deriving the generalization of stochastic gradient method called functional SGD (Section \ref{subsec:sgd}). The complexity of this online functional iterative optimization is cubic in the iteration index, a complicating factor of kernel methods which is untenable for streaming settings. 

Thus, we project the FSGD iterates onto sparse subspaces which are constructed from the span of a small number of kernel dictionary elements (Section \ref{subsec:proj}). To find these sparse subspaces of the RKHS, we make use of greedy sampling methods based on matching pursuit \citep{Pati1993}.
 The use of this technique is motivated by: (i) The fact that kernel matrices induced by arbitrary data streams will not, in general, satisfy requisite conditions for methods that enforce sparsity through convex relaxation \citep{candes2008restricted}; (ii) That having function iterates that exhibit small model order is of greater importance than exact recovery since SGD iterates are not the goal signal but just a noisy stepping stone to the optimal $f^*$. 
 Therefore, we construct these instantaneous sparse subspaces by making use of kernel orthogonal matching pursuit \citep{Vincent2002}, a greedy search routine which, given a function and an approximation budget $\eps$, returns its a sparse approximation and guarantees its output to be in a specific Hilbert-norm neighborhood of its function input.

To guarantee stochastic descent, we tie the size of the error neighborhood induced by sparse projections to the magnitude of the functional stochastic gradient and other problem parameters, thereby keeping only those kernel dictionary elements necessary for convergence (Section \ref{sec:convergence}). The result is that we are able to conduct stochastic gradient descent using only sparse projections of the stochastic gradient, maintaining a convergence path of moderate complexity towards the optimal $f^*$ \eqref{eq:kernel_stoch_opt}.
When the data and target domains ($\ccalX$ and $\ccalY$, respectively) are compact, for a certain approximation budget depending on the stochastic gradient algorithm step-size, we show that the sparse stochastically projected FSGD sequence still converges almost surely to the optimum of \eqref{eq:kernel_batch_opt} under both attenuating and constant learning rate schemes.
Moreover, the model order of this sequence remains finite for a given choice of constant step-size and approximation budget, and is, in the worst-case, comparable to the covering number of the data domain  \citep{zhou2002covering,pontil2003note}. 

In Section \ref{sec:experiments} we present numerical results on synthetic and empirical data for large-scale kernelized supervised learning tasks. We observe stable convergence behavior of POLK comparable to vector-valued first-order stochastic methods in terms of objective function evaluation, punctuated by a state of the art trade-off between test set error and number of samples processed. Further, the proposed method reduces the complexity of training kernel regressors by orders of magnitude.
In Section \ref{sec:discussion} we discuss our main findings. In particular, we suggest that there is a path forward for kernel methods as an alternative to neural networks that provides a more interpretable mechanism for inference with nonlinear statistical models and that one may achieve high generalization capability without losing convexity, an essential component for efficient training.

%
%

%
%

%
\section{Statistical Optimization in Reproducing Kernel Hilbert Spaces}\label{sec:problem}
%
Supervised learning is often formulated as an optimization problem that computes a set of parameters $\theta \in \Theta$ to minimize the average of a loss function $l : \Theta \times \ccalX \times \ccalY \rightarrow \reals $ for training examples $(\bbx_n,\bby_n) \in \ccalX \times \ccalY$. When the number of training examples $N$ is finite, this goal is referred to as \emph{empirical risk minimization} \citep{tewari13learning}, and may be solved using batch optimization techniques. The optimal $\bbtheta$ is the one that minimizes the regularized average loss, $\tilde{R}(\theta ; \{\bbx_n, y_n \}_{n=1}^N )= \frac{1}{N}\sum_{n=1}^N l(\bbtheta ; (\bbx_n, y_n))$, over the set of training data $\ccalS = \{\bbx_n,y_n\}_{n=1}^N$, i.e.,
\begin{align}\label{eq:batch_opt}
\theta^* &= \argmin_{\bbtheta \in \Theta} \tilde{R}(\bbtheta ; \ccalS)  = \argmin_{\bbtheta \in \Theta}\frac{1}{N}\sum_{n=1}^N l(\bbtheta ; \bbx_n,y_n ) + \frac{\lambda}{2}\|\bbtheta\|^2\; .
\end{align}
We focus on the case when the inputs are vectors $\bbx \in \ccalX \subseteq \reals^p$ and the target domain $\ccalY \subseteq \{0,1\}$ in the case of classification or $\ccalY \subseteq \reals$ in the case of regression.

\subsection{Supervised Kernel Learning}
In the case of supervised kernel learning \citep{MullerAFPT13,li2014kernel}, $\Theta$ is taken to be a Hilbert space, denoted here as $\ccalH$.  Elements of $\ccalH$ are \emph{functions}, $f: \ccalX \rightarrow \ccalY$, that admit a representation in terms of elements of $\ccalX$ when $\ccalH$ has a special structure.  In particular, equip $\ccalH$ with a unique \emph{kernel function}, $\kappa: \ccalX \times \ccalX \rightarrow \reals$, such that:
\begin{align} \label{eq:rkhs_properties}
(i) \  \langle f , \kappa(\bbx, \cdot)) \rangle _{\ccalH} = f(\bbx) \quad \text{for all } \bbx \in \ccalX \; ,
\qquad (ii) \ \ccalH = \closure{\text{span}\{ \kappa(\bbx , \cdot) \}} \quad\text{for all } \bbx \in \ccalX\; .
\end{align}
where $\langle \cdot , \cdot \rangle_{\ccalH}$ denotes the Hilbert inner product for $\ccalH$. We further assume that the kernel is positive semidefinite, i.e. $\kappa(\bbx, \bbx') \geq 0$ for all $\bbx, \bbx' \in \ccalX$. Function spaces with this structure are called reproducing kernel Hilbert spaces (RKHS).

In \eqref{eq:rkhs_properties}, property (i) is called the reproducing property of the kernel, and is a consequence of the Riesz Representation Theorem \citep{wheeden1977measure}.  Replacing $f$ by $\kappa(\bbx' , \cdot) $  in \eqref{eq:rkhs_properties} (i) yields the expression $ \langle \kappa(\bbx', \cdot) , \kappa(\bbx, \cdot) \rangle_{\ccalH} = \kappa(\bbx, \bbx')$, which is the origin of the term ``reproducing kernel."  This property provides a practical means by which to access a nonlinear transformation of the input space $\ccalX$.  Specifically, denote by $\phi(\cdot)$ a nonlinear map of the feature space that assigns to each $\bbx$ the kernel function $\kappa(\cdot, \bbx)$. Then the reproducing property of the kernel allows us to write the inner product of the image of distinct feature vectors $\bbx$ and $\bbx'$ under the map $\phi$ in terms of kernel evaluations only: $\langle \phi(\bbx), \phi(\bbx') \rangle_{\ccalH} =\kappa(\bbx, \bbx')$. This is commonly referred to as the \emph{kernel trick}, and it provides a computationally efficient tool for learning nonlinear functions.

Moreover, property \eqref{eq:rkhs_properties} (ii) states that any function $f\in \ccalH$ may be written as a linear combination of kernel evaluations. For kernelized and regularized empirical risk minimization, the Representer Theorem \citep{kimeldorf1971some,scholkopfgeneralized} establishes that the optimal $f$ in the hypothesis function class $\ccalH$ may be written as an expansion of kernel evaluations \emph{only} at elements of the training set as
%
%
\begin{equation}\label{eq:kernel_expansion}
f(\bbx) = \sum_{n=1}^N w_n \kappa(\bbx_n, \bbx)\; .
\end{equation}
where $\bbw = [w_1, \cdots, w_N]^T \in \reals^N$ denotes a set of weights. The upper summand index $N$ in \eqref{eq:kernel_expansion} is henceforth referred to as the model order. Common choices $\kappa$ include the polynomial kernel and the radial basis kernel, i.e., $\kappa(\bbx,\bbx') = \left(\bbx^T\bbx'+b\right)^c $ and $\kappa(\bbx,\bbx') = \exp\left\{ -\frac{\lVert \bbx - \bbx' \rVert_2^2}{2c^2} \right\}$, respectively, where $\bbx, \bbx' \in \ccalX$.

We may now formulate the kernel variant of the empirical risk minimization problem as the one that minimizes the loss functional $L : \ccalH \times \ccalX \times \ccalY \rightarrow \reals$ plus a complexity-reducing penalty. The loss functional $L$ may be written as an average over instantaneous losses $\ell: \ccalH \times \ccalX \times \ccalY \rightarrow \reals$, each of which penalizes the average deviation between $f(\bbx_n)$ and the associated output $y_n$ over the training set $\ccalS$. We denote the data loss and complexity loss as $R:\ccalH \rightarrow \reals$, and consider the problem
\begin{equation}\label{eq:kernel_batch_opt}
f^*=\argmin_{f \in \ccalH} {R} (f ; \ccalS )= \argmin_{f \in \ccalH}\frac{1}{N}\sum_{n=1}^N \ell(f(\bbx_n), y_n) +\frac{\lambda}{2}\|f \|^2_{\ccalH}\; .
\end{equation}
The above problem, referred to as Tikhonov regularization \citep{evgeniou2000regularization}, is one in which we aim to learn a general nonlinear relationship between $\bbx_n$ and $y_n$ through a function $f$. Throughout, we assume $\ell$ is convex with respect to its first argument $f(\bbx)$. By substituting the Representer Theorem expansion in \eqref{eq:kernel_expansion} into \eqref{eq:kernel_batch_opt}, the optimization problem amounts to finding an optimal set of coefficients $\bbw$ as
\begin{align}\label{eq:kernel_batch_opt2}
f^*&=\argmin_{\bbw \in \reals^N}\frac{1}{N}\sum_{n=1}^N \ell(\sum_{m=1}^N w_m \kappa(\bbx_m, \bbx_n), y_n) +\frac{\lambda}{2}\|\sum_{n,m=1}^N w_n w_m \kappa(\bbx_m, \bbx_n) \|^2_{\ccalH}  \nonumber \\
&=\argmin_{\bbw \in \reals^N}\frac{1}{N}\sum_{n=1}^N \ell(\bbw^T \boldsymbol{\kappa}_{\bbX}(\bbx_n), y_n) +\frac{\lambda}{2} \bbw^T \bbK_{\bbX,\bbX} \bbw \; 
\end{align}
where we have defined the Gram matrix (variously referred to as the \emph{kernel matrix}) $\bbK_{\bbX,\bbX}\in \reals^{N\times N}$, with entries given by the kernel evaluations between $\bbx_m$ and $\bbx_n$ as $[\bbK_{\bbX, \bbX}]_{m,n} =\kappa(\bbx_m, \bbx_n)$.  We further define the vector of kernel evaluations $\boldsymbol{\kappa}_{\bbX}(\cdot) = [\kappa(\bbx_1,\cdot) \ldots \kappa(\bbx_N,\cdot)]^T$, which are related to the kernel matrix as $\bbK_{\bbX,\bbX} = [\boldsymbol{\kappa}_{\bbX}(\bbx_1) \ldots \boldsymbol{\kappa}_{\bbX}(\bbx_N)]$. The dictionary of training points associated with the kernel matrix is defined as  $\bbX = [\bbx_1,\ \ldots\ ,\bbx_N]$.

Observe that by exploiting the Representer Theorem, we transform a nonparametric infinite dimensional optimization problem in $\ccalH$ \eqref{eq:kernel_batch_opt} into a finite $N$-dimensional parametric problem \eqref{eq:kernel_batch_opt2}. Thus, for empirical risk minimization, the RKHS provides a principled framework to solve nonparametric regression problems as via search over $\reals^N$ for an optimal set of coefficients.
A motivating example is presented next to clarify the setting of supervised kernel learning. 

\begin{example}(Kernel Logistic Regression) \label{eg1}\normalfont
Consider the case of \emph{kernel logistic regression} (KLR), with feature vectors $\bbx_n \in \ccalX\subseteq\reals^{p}$ and binary class labels $y_n\in \{0,1\}$. We seek to learn a function $f \in \ccalH$ that allows us to best approximate the distribution of an unknown class label given a training example $\bbx$ under the assumed model
\begin{align}\label{eqn:intro_lrprob}
\mbP\left( y = 0 \ | \ \bbx\right) = \frac{\exp\left\{ f(\bbx)\right\} }{1 + \exp\left\{ f(\bbx)\right\} } \; .
\end{align}
In classical logistic regression, we assume that $f$ is linear, i.e., $f(\bbx) = \bbc^T \bbx +  b$.  In KLR, on the other hand, we instead seek a nonlinear function of the form given in \eqref{eq:kernel_expansion}.
By making use of \eqref{eqn:intro_lrprob} and \eqref{eq:kernel_expansion}, we may formulate a maximum-likelihood estimation (MLE) problem to find the optimal function $f$ on the basis of $\ccalS$ by solving for the $\bbw$ that maximizes the $\lambda$-regularized average negative log likelihood over $\ccalS$, i.e.,
\begin{align}\label{eq:intro_loss} 
f^* &= \argmin_{f \in \ccalH} \frac{1}{N}\sum_{n=1}^N \left[ -\log \mbP(y=y_n \ | \ \bbx=\bbx_n) + \frac{\lambda}{2} \lVert f \rVert^2_{\ccalH} \right] \\
&=\argmin_{f\in\ccalH} \frac{1}{N}\sum_{n=1}^N \left[ \log\left(1+\exp\{f(\bbx_n)\}\right) - \mathbbm{1}(y_n=1) - f(\bbx_n)\mathbbm{1}(y_n=0) + \frac{\lambda}{2} \lVert f \rVert^2_{\ccalH} \right] \nonumber \\
&=\!\argmin_{\bbw \in \reals^N}\! \frac{1}{N}\!\!\sum_{n=1}^N\!\!\left[ \log\!\left(\!1\!\!+\! \exp\{\!\bbw^T\!\!\boldsymbol{\kappa}_{\bbX}\!(\bbx_n\!)\!\} \!\right) \!\!-\! \!\mathbbm{1}(y_n=1)\! - \!\bbw^T\!\!\boldsymbol{\kappa}_{\bbX}(\bbx_n)\mathbbm{1}(y_n=0) \!+\! \frac{\lambda}{2}\bbw^T \bbK_{\bbX,\bbX} \bbw \right], \nonumber
\end{align}
where $\mathbbm{1}(\cdot)$ represents the indicator function.  Solving \eqref{eq:intro_loss} amounts to finding a function $f$ that, given a feature vector $\bbx$ and the model outlined by \eqref{eqn:intro_lrprob}, best represents the class-conditional probabilities that the corresponding label $y$ is either $0$ or $1$.
\end{example}

\subsection{Online Kernel Learning}
The goal of this paper is to solve problems of the form \eqref{eq:kernel_batch_opt} when training examples $(\bbx_n, \bby_n)$ either become sequentially available or their total number is not necessarily finite.  To do so, we consider the case where $(\bbx_n, \bby_n)$ are independent realizations from a stationary joint distribution of the random pair $(\bbx, \bby) \in \ccalX \times \ccalY$ \citep{slavakis2013online}.  In this case, the objective in \eqref{eq:kernel_batch_opt} may be written as an expectation over this random pair as
\begin{align}\label{eq:kernel_stoch_opt2}
f^*=\argmin_{f \in \ccalH} R(f) :& = \argmin_{f \in \ccalH}\mbE_{\bbx, \bby}{[ \ell(f(\bbx), y)}] +\frac{\lambda}{2}\|f \|^2_{\ccalH} \;\\
						& = \!\!\argmin_{\bbw\in \reals^{\ccalI}, \{\bbx_n\}_{n\in\ccalI}}\mbE_{\bbx, \bby}{[ \ell(\sum_{n\in\ccalI} w_n \kappa(\bbx_n, \bbx) , y)}]  +\frac{\lambda}{2}\|\!\! \!\sum_{n,m\in \ccalI } \!\!w_n w_m \kappa(\bbx_m, \bbx_n)  \|^2_{\ccalH} \  \;. \nonumber
\end{align}
where we define the average loss as $L(f) : = \mbE_{\bbx, \bby}{[ \ell(f(\bbx), y)}]$. In the second equality in \eqref{eq:kernel_stoch_opt}, we substitute in the expansion of $f$ given by the Representer Theorem generalized to the infinite sample-size case established in \citep{norkin2009stochastic}, with $\ccalI$ as some countably infinite indexing set. 
\section{Algorithm Development}
\label{sec:algorithm}

We turn to deriving an algorithmic solution to the kernelized expected risk minimization problem stated in \eqref{eq:kernel_stoch_opt}. To do so, two complexity bottlenecks must be overcome. The first is that in order to develop a numerical optimization scheme such as gradient descent, we must compute the functional gradient (Frech$\acute{\text{e}}$t derivative) of the expected risk $L(f)$ with respect to $f$, which requires infinitely many realizations of the random pair $(\bbx, y)$. This bottleneck is handled via stochastic approximation, as detailed in Section \ref{subsec:sgd}. 
The second issue is that when making use of the stochastic gradient method in the RKHS setting, the resulting parametric updates require memory storage whose complexity is cubic in the iteration index (the curse of kernelization), which rapidly becomes unaffordable. To alleviate this memory explosion, we introduce our sparse stochastic projection scheme based upon kernel orthogonal matching pursuit in Section \ref{subsec:proj}.

\subsection{Functional Stochastic Gradient Descent}\label{subsec:sgd}
Following \cite{Kivinen2004}, we derive the generalization of the stochastic gradient method for the RKHS setting.  The resulting procedure is referred to as \emph{functional stochastic gradient descent} (FSGD). First, given an independent realization $(\bbx_t, y_t)$ of the random pair $(\bbx, y)$, we compute the stochastic functional gradient (Frech$\acute{\text{e}}$t derivative) of $L(f)$, stated as
\begin{align}\label{eq:stochastic_grad}
\nabla_f \ell(f(\bbx_t),y_t)(\cdot) 
= \frac{\partial \ell(f(\bbx_t),y_t)}{\partial f(\bbx_t)}\frac{\partial f(\bbx_t)}{\partial f}(\cdot)
\end{align}
where we have applied the chain rule. Now, define the short-hand notation $\ell'(f(\bbx_t),y_t): ={\partial \ell(f(\bbx_t),y_t)}/{\partial f(\bbx_t)} $ for the derivative of $\ell(\bbf(\bbx_t),y_t)$ with respect to its first scalar argument $f(\bbx_t)$ evaluated at $\bbx_t$. To evaluate the second term on the right-hand side of \eqref{eq:stochastic_grad}, differentiate both sides of the expression defining the reproducing property of the kernel [cf. \eqref{eq:rkhs_properties}(i)] with respect to $f$ to obtain
\begin{align}\label{eq:stochastic_grad2}
\frac{\partial  f(\bbx_t)}{\partial f} = \frac{\partial \langle f , \kappa(\bbx_t, \cdot)) \rangle _{\ccalH}}{\partial f}
= \kappa(\bbx_t,\cdot)
\end{align}
With this computation in hand, we present the stochastic gradient method for the kernelized $\lambda$-regularized expected risk minimization problem in \eqref{eq:kernel_stoch_opt} as
\begin{align}\label{eq:sgd_hilbert}
f_{t+1} =(1-\eta_t \lambda ) f_{t} - \eta_t \nabla_f \ell (f_{t}(\bbx_t),y_t)=(1-\eta_t \lambda ) f_{t} - \eta_t \ell'(f(\bbx_t),y_t) \kappa(\bbx_t,\cdot) \; ,
\end{align}
where $\eta_t> 0$ is an algorithm step-size either chosen as diminishing with $\ccalO(1/t)$ or a small constant -- see Section \ref{sec:convergence}. We further require that, given $\lambda > 0$, the step-size satisfies $\eta_t < 1/\lambda$ and the sequence is initialized as $f_0 = 0 \in \ccalH$. Given this initialization, by making use of the Representer Theorem \eqref{eq:kernel_expansion}, at time $t$, the function $f_t$ may be expressed as an expansion in terms of feature vectors $\bbx_t$ observed thus far as
\begin{align}\label{eq:kernel_expansion_t}
f_t(\bbx) 
= \sum_{n=1}^{t-1} w_n \kappa(\bbx_n, \bbx)
= \bbw_t^T\boldsymbol{\kappa}_{\bbX_t}(\bbx) \; .
\end{align}
On the right-hand side of \eqref{eq:kernel_expansion_t} we have introduced the notation $\bbX_t = [\bbx_1, \ldots, \bbx_{t-1}]\in \reals^{p\times (t-1)}$ and $\boldsymbol{\kappa}_{\bbX_t}(\cdot) = [\kappa(\bbx_1,\cdot),\ \ldots\ ,\kappa(\bbx_{t-1},\cdot)]^T$. Moreover, observe that the kernel expansion in \eqref{eq:kernel_expansion_t}, taken together with the functional update \eqref{eq:sgd_hilbert}, yields the fact that performing the stochastic gradient method in $\ccalH$ amounts to the following parametric updates on the kernel dictionary $\bbX$ and coefficient vector $\bbw$:
\begin{align}\label{eq:param_update} 
\bbX_{t+1} = [\bbX_t, \;\; \bbx_t],\;\;\;\; \bbw_{t+1} = [ (1 - \eta_t \lambda) \bbw_t , \;\; -\eta_t\ell'(f_t(\bbx_t),y_t)]  \; ,
\end{align}
Observe that this update causes $\bbX_{t+1}$ to have one more column than $\bbX_t$. We define the \emph{model order} as number of data points $M_t$ in the dictionary at time $t$ (the number of columns of $\bbX_t$). FSGD is such that $M_t=t-1$, and hence grows unbounded with iteration index $t$.

\subsection{Model Order Control via Stochastic Projection}\label{subsec:proj}
To mitigate the model order issue described above, we shall generate an approximate sequence of functions by orthogonally projecting functional stochastic gradient updates onto subspaces $\ccalH_\bbD \subseteq \ccalH$ that consist only of functions that can be represented using some dictionary $\bbD = [\bbd_1,\ \ldots,\ \bbd_M] \in \reals^{p \times M}$, i.e., $\ccalH_\bbD = \{f\ :\ f(\cdot) = \sum_{n=1}^M w_n\kappa(\bbd_n,\cdot) = \bbw^T\boldsymbol{\kappa}_{\bbD}(\cdot) \}=\text{span}\{\kappa(\bbd_n, \cdot) \}_{n=1}^M$. For convenience we have defined $[\boldsymbol{\kappa}_{\bbD}(\cdot)=\kappa(\bbd_1,\cdot) \ldots \kappa(\bbd_M,\cdot)]$, and $\bbK_{\bbD,\bbD}$ as the resulting kernel matrix from this dictionary. We will enforce parsimony in function representation by selecting dictionaries $\bbD$ that $M_t << t$.

We first show that, by selecting $\bbD = \bbX_{t+1}$ at each iteration, the sequence \eqref{eq:sgd_hilbert} derived in Section \ref{subsec:sgd} may be interpreted as carrying out a sequence of orthogonal projections.  To see this, rewrite \eqref{eq:sgd_hilbert} as the quadratic minimization
\begin{align}\label{eq:proximal_hilbert_dictionary}
f_{t+1} &= \argmin_{f \in \ccalH}
\Big\lVert f - \Big((1-\eta_t \lambda) f_t 
- \eta_t \nabla_f\ell(f_{t}(\bbx_t),y_t) \Big) \Big\rVert_{\ccalH}^2 \nonumber \\
&= \argmin_{f \in \ccalH_{\bbX_{t+1}}} 
 \Big\lVert f - \Big( (1-\eta_t \lambda)f_t - 
 \eta_t \nabla_f\ell(f_{t}(\bbx_t),y_t) \Big) \Big\rVert_{\ccalH}^2, 
\end{align}
where the first equality in \eqref{eq:proximal_hilbert_dictionary} comes from ignoring constant terms which vanish upon differentiation with respect to $f$, and the second comes from observing that $f_{t+1}$ can be represented using only the points $\bbX_{t+1}$, using \eqref{eq:param_update}.  Notice now that \eqref{eq:proximal_hilbert_dictionary} expresses $f_{t+1}$ as the orthogonal projection of the update $(1-\eta_t \lambda) f_t - \eta_t\nabla_f\ell(f_{t}(\bbx_t),y_t)$ onto the subspace defined by dictionary $\bbX_{t+1}$.

Rather than select dictionary $\bbD=\bbX_{t+1}$, we propose instead to select a different dictionary, $\bbD=\bbD_{t+1}$, which is extracted from the data points observed thus far, at each iteration.  The process by which we select $\bbD_{t+1}$ will be discussed shortly, but is of of dimension $p \times {M}_{t+1}$, with ${M}_{t+1} << t$.  As a result, we shall generate a function sequence ${f}_t$ that differs from the functional stochastic gradient method presented in Section \ref{subsec:sgd}. The function ${f}_{t+1}$ is parameterized dictionary $\bbD_{t+1}$ and weight vector $\bbw_{t+1}$. We denote columns of $\bbD_{t+1}$ as $\bbd_n$ for $n=1,\dots,{M}_{t+1}$, where the time index is dropped for notational clarity but may be inferred from the context.

To be specific, we propose replacing the update \eqref{eq:proximal_hilbert_dictionary} in which the dictionary grows at each iteration by the stochastic projection of the functional stochastic gradient sequence onto the subspace $\ccalH_{\bbD_{t+1}}=\text{span}\{ \kappa(\bbd_n, \cdot) \}_{n=1}^{M_{t+1}}$ as
\begin{align}\label{eq:projection_hat}
{f}_{t+1}& = \argmin_{f \in \ccalH_{\bbD_{t+1}}}  \Big\lVert f - 
\Big((1-\eta_t \lambda) f_t 
- \eta_t \nabla_f\ell(f_{t}(\bbx_t),y_t) \Big)\Big\rVert_{\ccalH}^2 \nonumber \\
&:=\ccalP_{\ccalH_{\bbD_{t+1}}} \Big[ 
(1-\eta_t \lambda) f_t 
- \eta_t \nabla_f\ell(f_{t}(\bbx_t),y_t) \Big] .
\end{align}
where we define the projection operator $\ccalP$ onto subspace $\ccalH_{\bbD_{t+1}}\subset \ccalH$ by the update \eqref{eq:projection_hat}.
 
{\bf Coefficient update} The update \eqref{eq:projection_hat}, for a fixed dictionary $\bbD_{t+1} \in \reals^{p\times M_{t+1}}$, may be expressed in terms of the parameter space of coefficients only. In order to do so, we first define the stochastic gradient update without projection, given function ${f}_t$ parameterized by dictionary $\bbD_t$ and coefficients $\bbw_t$, as
\begin{align}\label{eq:sgd_tilde}
\tilde{f}_{t+1} =(1 - \eta_t\lambda ) {f}_t - \eta_t\nabla_f\ell({f}_t;\bbx_t,\bby_t).
\end{align}
This update may be represented using dictionary and weight vector
\begin{align}\label{eq:param_tilde}
\tbD_{t+1} = [\bbD_t,\;\;\bbx_t], \;\;\;\; \tbw_{t+1} = [(1-\eta_t\lambda)\bbw_t,\;\; -\eta_t\ell'({f}_t(\bbx_t),y_t)] \; .
\end{align}
%
%
Observe that $\tbD_{t+1}$ has $\tilde{M}=M_t + 1$ columns, which is also the length of $\tbw_{t+1}$. For a fixed dictionary $\bbD_{t+1}$, the stochastic projection in \eqref{eq:projection_hat} amounts to a least-squares problem on the coefficient vector. To see this, make use of the Representer Theorem to rewrite \eqref{eq:projection_hat} in terms of kernel expansions, and that the coefficient vector is the only free parameter to write
\begin{align}\label{eq:proximal_hilbert_representer}
 &\argmin_{\bbw \in \reals^{{M}_{t+1}}} 
\frac{1}{2\eta_t} \Big\lVert \sum_{n=1}^{{M}_{t+1}} {w}_n\kappa(\bbd_n,\cdot)
 - \sum_{m=1}^{\tilde{M}}\tilde{w}_m\kappa(\tbd_m,\cdot) \Big\rVert_{\ccalH}^2  \\
&\quad= \argmin_{\bbw \in \reals^{{M}_{t+1}}} \! \! 
 \frac{1}{2\eta_t} \!  \left( \! \sum_{n,{n'}=1}^{{M}_{t+1}} \!\!\!\! {w}_n{w}_{n'} \kappa(\bbd_{n},\bbd_{n'})
  - 2 \!\!\!\!  \! \sum_{n,m=1}^{{M}_{t+1},\tilde{M}}  \!\!\!{w}_n\tilde{w}_m \kappa(\bbd_n,\tbd_m)
  \!  +\!  \!\!\!\!  \sum_{m,{m'}=1}^{\tilde{M}} \tilde{w}_{m}\tilde{w}_{m'}\kappa(\tbd_{m},\tbd_{m'})\! \!  \right)  \nonumber \\
&\quad= \argmin_{\bbw \in \reals^{{M}_{t+1}}} \! \!
 \frac{1}{2\eta_t} \! \!\left(\! \bbw^T\bbK_{\bbD_{t+1},\bbD_{t+1}}\bbw
 \!-\! 2\bbw^T\bbK_{\bbD_{t+1},\tbD_{t+1}}\tbw_{t+1}
  + \tbw_{t+1}\bbK_{\tbD_{t+1},\tbD_{t+1}}\tbw_{t+1}\! \right) \!
  := \bbw_{t+1} \; . \nonumber
\end{align}
In \eqref{eq:proximal_hilbert_representer}, the first equality comes from expanding the square, and the second comes from defining the cross-kernel matrix $\bbK_{\bbD_{t+1},\tbD_{t+1}}$ whose $(n,m)^\text{th}$ entry is given by $\kappa(\bbd_n,\tbd_m)$. The other kernel matrices $\bbK_{\tbD_{t+1},\tbD_{t+1}}$ and $\bbK_{\bbD_{t+1},\bbD_{t+1}}$ are similarly defined. Note that ${M}_{t+1}$ is the number of columns in $\bbD_{t+1}$, while $\tilde{M}=M_t + 1$ is the number of columns in $\tbD_{t+1}$ [cf. \eqref{eq:param_tilde}]. The explicit solution of \eqref{eq:proximal_hilbert_representer} may be obtained by noting that the last term is a constant independent of $\bbw$, and thus by computing gradients and solving for $\bbw_{t+1}$ we obtain
\begin{equation} \label{eq:hatparam_update}
\bbw_{t+1}=  \bbK_{\bbD_{t+1} \bbD_{t+1}} [\bbK_{\bbD_{t+1} \tbD_{t+1}}]^{\dagger} \tbw_{t+1} \;,
\end{equation}
where $\dagger$ is used to denote the Moore-Penrose pseudoinverse. Given that the projection of $\tilde{f}_{t+1}$ onto the stochastic subspace $\ccalH_{\bbD_{t+1}}$, for a fixed dictionary $\bbD_{t+1}$, amounts to a simple least-squares multiplication, we turn to detailing how the kernel dictionary $\bbD_{t+1}$ is selected from the data sample path $\{\bbx_u, y_u\}_{u \leq t}$.

{\bf Dictionary Update} The selection procedure for the kernel dictionary $\bbD_{t+1}$ is based upon greedy sparse approximation, a topic studied extensively in the compressive sensing community \citep{needell2008greedy}.  The function $\tilde{f}_{t+1} =(1-\eta_t) {f}_t - \eta_t\nabla_f\ell({f}_t;\bbx_t,\bby_t)$ defined by stochastic gradient method without projection is parameterized by dictionary $\tbD_{t+1}$ [cf. \eqref{eq:param_tilde}], whose model order is $\tilde{M}={M}_t +1$. We form $\bbD_{t+1}$ by selecting a subset of $M_{t+1}$ columns from $\tbD_{t+1}$ that are best for approximating $\tilde{f}_{t+1}$ in terms of error with respect to the Hilbert norm. As previously noted, numerous approaches are possible for seeking a sparse representation. We make use of \emph{kernel orthogonal matching pursuit} (KOMP) \citep{Vincent2002} with allowed error tolerance $\epsilon_t$ to find a kernel dictionary matrix $\bbD_{t+1}$ based on the one which adds the latest sample point $\tbD_{t+1}$. This choice is due to the fact that we can tune its stopping criterion to guarantee stochastic descent, and guarantee the model order of the learned function remains finite -- see Section \ref{sec:convergence} for details.
%
\begin{algorithm}[t]
\caption{Destructive Kernel Orthogonal Matching Pursuit (KOMP) \hspace{-2mm}}
\begin{algorithmic}
\label{alg:komp}
\REQUIRE  function $\tilde{f}$ defined by dict. $\tbD \in \reals^{p \times \tilde{M}}$, coeffs. $\tbw \in \reals^{\tilde{M}}$, approx. budget  $\epsilon_t > 0$ \\
\STATE \textbf{initialize} $f=\tilde{f}$, dictionary $\bbD = \tbD$ with indices $\ccalI$, model order $M=\tilde{M}$, coeffs.  $\bbw = \tbw$.
\WHILE{candidate dictionary is non-empty $\ccalI \neq \emptyset$}
{\FOR {$j=1,\dots,\tilde{M}$}
	\STATE Find minimal approximation error with dictionary element $\bbd_j$ removed \vspace{-2mm}
	$$\gamma_j = \min_{\bbw_{\ccalI \setminus \{j\}}\in\reals^{{M}-1}} \|\tilde{f}(\cdot) - \sum_{k \in \ccalI \setminus \{j\}} w_k \kappa(\bbd_k, \cdot) \|_{\ccalH} \; .$$ \vspace{-5mm}
\ENDFOR}
	\STATE Find dictionary index minimizing approximation error: $j^* = \argmin_{j \in \ccalI} \gamma_j$
	\INDSTATE{{\bf{if }} minimal approximation error exceeds threshold $\gamma_{j^*}> \epsilon_t$}
	\INDSTATED{\bf stop} 
	\INDSTATE{\bf else} 
	
	\INDSTATED Prune dictionary $\bbD\leftarrow\bbD_{\ccalI \setminus \{j^*\}}$
	\INDSTATED Revise set $\ccalI \leftarrow \ccalI \setminus \{j^*\}$ and model order ${M} \leftarrow {M}-1$.
	\INDSTATED Compute updated weights $\bbw$ defined by current dictionary $\bbD$
 	\vspace{-2mm}$$\bbw = \argmin_{\bbw \in \reals^{{M}}} \lVert \tilde{f}(\cdot) - \bbw^T\boldsymbol{\kappa}_{\bbD}(\cdot) \rVert_{\ccalH}$$\vspace{-5mm}
	\INDSTATE {\bf end}
\ENDWHILE	
\RETURN ${f},\bbD,\bbw$ of model order $M \leq \tilde{M}$ such that $\|f - \tilde{f} \|_{\ccalH}\leq \eps_t$
\end{algorithmic}
\end{algorithm}

We now describe the variant of KOMP we propose using, called Destructive KOMP with Pre-Fitting (see \cite{Vincent2002}, Section 2.3), which is summarized in Algorithm \ref{alg:komp}. This flavor of KOMP takes as an input a candidate function $\tilde{f}$ of model order $\tilde{M}$ parameterized by its kernel dictionary $\tbD\in\reals^{p\times\tilde{M}}$ and coefficient vector $\tbw\in\reals^{\tilde{M}}$. The method then seeks to approximate $\tilde{f}$ by a parsimonious function $f\in \ccalH$ with a lower model order. Initially, this sparse approximation is the original function $f = \tilde{f} $ so that its dictionary is initialized with that of the original function $\bbD=\tbD$, with corresponding coefficients  $\bbw=\tbw$. 
Then, the algorithm sequentially removes dictionary elements from the initial dictionary $\tbD$, yielding a sparse approximation $f$ of $\tilde{f}$, until the error threshold $\|f - \tilde{f} \|_{\ccalH} \leq \eps_t $ is violated, in which case it terminates.

At each stage of KOMP, a single dictionary element $j$ of $\bbD$ is selected to be removed which contributes the least to the Hilbert-norm approximation error $\min_{f\in\ccalH_{\bbD\setminus \{j\}}}\|\tilde{f} - f \|_{\ccalH}$ of the original function $\tilde{f}$, when dictionary $\bbD$ is used. Since at each stage the kernel dictionary is fixed, this amounts to a computation involving weights $\bbw \in \reals^{M-1}$ only; that is, the error of removing dictionary point $\bbd_j$ is computed for each $j$ as 
 $\gamma_j =\min_{\bbw_{\ccalI \setminus \{j\}}\in\reals^{{M}-1}} \|\tilde{f}(\cdot) - \sum_{k \in \ccalI \setminus \{j\}} w_k \kappa(\bbd_k, \cdot) \|.$
  We use the notation $\bbw_{\ccalI \setminus \{j\}}$ to denote the entries of $\bbw\in \reals^M$ restricted to the sub-vector associated with indices $\ccalI \setminus \{j\}$. Then, we define the dictionary element which contributes the least to the approximation error as $j^*=\argmin_j \gamma_j$. If the error associated with removing this kernel dictionary element exceeds the given approximation budget $\gamma_{j^*}>\eps_t$, the algorithm terminates. Otherwise, this dictionary element $\bbd_{j^*}$ is removed, the weights $\bbw$ are revised based on the pruned dictionary as $\bbw = \argmin_{\bbw \in \reals^{{M}}} \lVert \tilde{f}(\cdot) - \bbw^T\boldsymbol{\kappa}_{\bbD}(\cdot) \rVert_{\ccalH}$, and the process repeats as long as the current function approximation is defined by a nonempty dictionary.


With Algorithm \ref{alg:komp} stated, we may summarize the key steps of the proposed method in Algorithm \ref{alg:soldd} for solving \eqref{eq:kernel_stoch_opt} while maintaining a finite model order, thus breaking the ``curse of kernelization." The method, Parsimonious Online Learning with Kernels (POLK), executes the stochastic projection of the functional stochastic gradient iterates onto sparse subspaces $\ccalH_{\bbD_{t+1}}$ stated in \eqref{eq:projection_hat}. The initial function is set to null $f_0=0$, meaning that it has empty kernel dictionary $\bbD_0=[]$ and coefficient vector $\bbw_0=[]$. The notation $[]$ is used to denote the empty matrix or vector respective size $p\times0$ or $0$. Then, at each step, given an independent training example $(\bbx_t, y_t)$ and step-size $\eta_t$, we compute the \emph{unconstrained} functional stochastic gradient iterate $\tilde{f}_{t+1}(\cdot) = (1-\eta_t\lambda){f}_t - \eta_t\ell'({f}_t(\bbx_t),\bby_t)\kappa(\bbx_t,\cdot)$ which admits the parametric representation $\tbD_{t+1}$ and $\tbw_{t+1}$ as stated in \eqref{eq:param_tilde}. These parameters are then fed into KOMP with approximation budget $\eps_t$, such that $(f_{t+1}, \bbD_{t+1}, \bbw_{t+1})= \text{KOMP}(\tilde{f}_{t+1},\tilde{\bbD}_{t+1}, \tilde{\bbw}_{t+1},\eps_t)$.

%
\begin{algorithm}[t]
\caption{Parsimonious Online Learning with Kernels (POLK)}
\begin{algorithmic}
\label{alg:soldd}
\REQUIRE $\{\bbx_t,\bby_t,\eta_t,\epsilon_t \}_{t=0,1,2,...}$
\STATE \textbf{initialize} ${f}_0(\cdot) = 0, \bbD_0 = [], \bbw_0 = []$, i.e. initial dictionary, coefficient vectors are empty
\FOR{$t=0,1,2,\ldots$}
	\STATE Obtain independent training pair realization $(\bbx_t, y_t)$
	\STATE Compute unconstrained functional stochastic gradient step [cf. \eqref{eq:sgd_tilde}]
	$$\tilde{f}_{t+1}(\cdot) = (1-\eta_t\lambda){f}_t - \eta_t\ell'({f}_t(\bbx_t),\bby_t)\kappa(\bbx_t,\cdot)$$
	
	\STATE Revise dictionary $\tbD_{t+1} = [\bbD_t,\;\;\bbx_t]$ and weights $\tbw_{t+1} \leftarrow [(1-\eta_t\lambda)\bbw_t,\;\; -\eta_t\ell'({f}_t(\bbx_t),y_t)]$
	\STATE Compute sparse function approximation via Algorithm \ref{alg:komp} 
	$$({f}_{t+1},\bbD_{t+1},\bbw_{t+1}) = \textbf{KOMP}(\tilde{f}_{t+1},\tbD_{t+1},\tbw_{t+1},\epsilon_t)$$
\ENDFOR
\end{algorithmic}
\end{algorithm}

In the next section, we discuss the analytical properties of Algorithm \ref{alg:soldd} for solving online nonparametric regression problems of the form \eqref{eq:kernel_stoch_opt}. We close here with an example algorithm derivation for the kernel logistic regression problem stated in Example \ref{eg1}.

\begin{example} (Kernel Logistic Regression) \label{eg2}\normalfont
Returning to the case of {kernel logistic regression} stated in Example \ref{eg1}, with feature vectors $\bbx_n \in \ccalX\subseteq\reals^{p}$ and binary class labels $y_n\in \{0,1\}$, we may perform sparse function estimation in $\ccalH$ that fits a a training example $\bbx$ to its associated label $y$ under the logistic model [cf. \eqref{eqn:intro_lrprob}] of the odds-ratio of the given class label. The associated $\lambda$-regularized maximum-likelihood estimation (MLE) is given as \eqref{eq:intro_loss}. Provided that a particular kernel map $\kappa(\cdot,\cdot)$, regularizer $\lambda$, and step-size $\eta_t$ have been chosen, the only specialization of Algorithm \ref{alg:soldd} to this case is the computation of $\tilde{f}_t$, which requires computing the stochastic gradient of \eqref{eqn:intro_lrprob} with respect to an instantaneous training example $(\bbx_t, y_t)$. Doing so specializes \eqref{eq:sgd_tilde} to
\begin{align}\label{eq:logistic_soldd}
	\tilde{f}_{t+1}(\cdot) = (1-\eta_t\lambda){f}_t - \eta_t
 \frac{	 \exp\{-\tilde{f}_t(\bbx_t) \}}{[1+ \exp\{- \tilde{f}_t(\bbx_t) \}]^2 }
	\kappa(\bbx_t,\cdot) \; .
\end{align}
The resulting dictionary and parameter updates implied by \eqref{eq:logistic_soldd}, given in \eqref{eq:param_tilde}, are then fed into KOMP (Algorithm \ref{alg:komp}) which returns their greedy sparse approximation for a fixed budget $\eps_t$.
\end{example}

\section{Convergence Analysis}\label{sec:convergence}
We turn to studying the theoretical performance of Algorithm \ref{alg:soldd} developed in Section \ref{sec:algorithm}. In particular, we establish that the method, when a diminishing step-size is chosen, is guaranteed to converge to the optimum of \eqref{eq:kernel_stoch_opt}. We further obtain that when a sufficiently small constant step-size is chosen, the limit infimum of the iterate sequence is within a neighborhood of the optimum. In both cases, the convergence behavior depends on the approximation budget used in the online sparsification procedure detailed in Algorithm \ref{alg:komp}.

We also perform a worst-case analysis of the model order of the instantaneous iterates resulting from Algorithm \ref{alg:soldd}, and show that asymptotically the model order depends on that of the optimal $f^*\in\ccalH$. 

%
\subsection{Iterate Convergence}\label{subsec:convergence}
As is customary in the analysis of stochastic algorithms, we establish that under a diminishing algorithm step-size scheme (non-summable and square-summable), with the sparse approximation budget selection
\begin{equation}\label{eq:dim_stepsize}
\sum_{t=1}^\infty \eta_t = \infty \;, \quad \sum_{t=1}^\infty \eta_t^2 < \infty \;, \quad \eps_t = \eta_t^2 \;, 
\end{equation}
Algorithm \ref{alg:soldd} converges exactly to the optimal function $f^*$ in stated in \eqref{eq:kernel_stoch_opt} almost surely.
%
\begin{theorem}\label{theorem_diminishing}
Consider the sequence generated $\{f_t \}$ by Algorithm \ref{alg:soldd} with $f_0 = 0$, and denote $f^*$ as the minimizer of the regularized expected risk stated in \eqref{eq:kernel_stoch_opt}. Let Assumptions \ref{as:first}-\ref{as:last} hold and suppose the step-size rules and approximation budget are diminishing as in \eqref{eq:dim_stepsize} with regularizer such that $\eta_t < 1/\lambda$ for all $t$. Then the objective function error sequence converges to null in infimum almost surely as
\begin{equation}\label{eq:liminf_objective_thm}
\liminf_{t\rightarrow \infty} R(f_t) - R(f^*) = 0 \qquad\text{ a.s.}
\end{equation}
Moreover, the sequence of functions $\{f_t \}$ converges almost surely to the optimum $f^*$ as
\begin{equation}\label{eq:sequence_conv_thm}
\lim_{t\rightarrow \infty} \| f_t - f^*\|_{\ccalH}^2 = 0 \qquad \text{ a.s.}
\end{equation}
\end{theorem}
\begin{myproof} See Appendix \ref{apx_theorem_diminishing}. \end{myproof}

The result in Theorem \ref{theorem_diminishing} states that when a diminishing algorithm step-size is chosen as, e.g. $\eta_t=\ccalO(1/t)$, and the approximation budget that dictates the size of the sparse stochastic subspaces onto which the iterates are projected is selected as $\eps_t=\eta_t^2$, we obtain exact convergence to the optimizer of the regularized expected risk in \eqref{eq:kernel_stoch_opt}. However, in obtaining exact convergence behavior, we require the approximation budget to approach null asymptotically, which means that the model order of the resulting function sequence may grow arbitrarily, unless $f^*$ is sparse and the magnitude of the stochastic gradient reduces sufficiently quickly, i.e., comparable to $\eps_t = \ccalO(1/t^2)$.

If instead we consider a constant algorithm step-size $\eta_t=\eta$ and the approximation budget $\eps_t=\eps$ is chosen as a constant which satisfies $\eps_t=\eps=\ccalO(\eta^{3/2})$, we obtain that the iterates converge in infimum to a neighborhood of the optimum, as we state next.
%
\begin{theorem}\label{theorem_constant}
Denote $\{f_t \}$ as the sequence generated by Algorithm \ref{alg:soldd} with $f_0 = 0$, and denote $f^*$ as the minimizer of the regularized expected risk stated in \eqref{eq:kernel_stoch_opt}. Let Assumptions \ref{as:first}-\ref{as:last} hold, and given regularizer $\lambda > 0$, suppose a constant algorithm step-size $\eta_t=\eta$ is chosen such that $\eta < 1/\lambda$, and the sparse approximation budget satisfies $\eps=K \eta^{3/2}=\ccalO(\eta^{3/2})$, where $K$ is a positive scaler. Then the algorithm converges to a neighborhood almost surely as
\begin{equation}\label{eq:theorem_constant}
\liminf_{t\rightarrow \infty} \|f_t - f^* \|_\ccalH \leq \frac{\sqrt{\eta}}{\lambda}\Big(K +  \sqrt{K^2 + \lambda  \sigma^2 }\Big) = \ccalO(\sqrt{\eta}) \qquad \text{ a.s. }
\end{equation}
\end{theorem}
\begin{myproof} See Appendix \ref{apx_theorem_constant}. \end{myproof}

Theorem \ref{theorem_constant} states that when a sufficiently small constant step-size is used together with a bias tolerance induced by sparsification chosen as $\eps=\ccalO(\eta^{3/2})$, Algorithm \ref{alg:soldd} converges in infimum to a neighborhood of the optimum which depends on the chosen step-size, the parsimony constant $K$ which scales the approximation budget $\eps$, the regularization parameter $\lambda$, as well as the variance of the stochastic gradient $\sigma^2$. This result again is typical of convergence results in stochastic gradient methods. However, the use of a constant learning rate allows use to guarantee the model order of the resulting function sequence is always bounded, as we establish in the following subsection.

\subsection{Model Order Control}\label{subsec:model_order}
In this subsection, we establish that the sequence of functions $\{f_t\}$ generated by Algorithm \ref{alg:soldd}, when a constant algorithm step-size is selected, is parameterized by a kernel dictionary which is guaranteed to have finitely many elements, i.e., its the model order remains bounded. We obtain that the worst-case bound on the model order of $f_t$ is depends by the topological properties of the feature space $\ccalX$, the Lipschitz constant of the instantaneous loss, and the radius of convergence $\Delta = ({\sqrt{\eta}}/{\lambda})(K +  \sqrt{K^2 + \lambda  \sigma^2 })$ defined in Theorem \ref{theorem_constant}. 
%
\begin{theorem}\label{theorem_model_order}
Denote $f_t$ as the function sequence defined by Algorithm \ref{alg:soldd} with constant step-size $\eta_t=\eta<1/\lambda$ and approximation budget $\eps=K \eta^{3/2}$ where $K>0$ is an arbitrary positive scalar. Let $M_t$ be the model order of $f_t$ i.e., the number of columns of the dictionary $\bbD_t$ which parameterizes $f_t$. Then there exists a finite upper bound $M^\infty$ such that, for all $t\geq0$, the model order is always bounded as $M_t\leq M^\infty$. Consequently, the model order of the limiting function $f^\infty=\lim_t f_t$ is finite.

\end{theorem}
\begin{myproof} See Appendix \ref{apx_theorem_model_order}. \end{myproof}
The number of kernel dictionary elements in the function sequence $f_t$ generated by Algorithm \ref{alg:soldd} is in the worst-case determined by the packing number of the kernel transformation of the feature space $\phi(\ccalX) = \kappa(\ccalX,\cdot)$, as shown in the proof of Theorem \ref{theorem_model_order}. Moreover, the online sparsification procedure induced by KOMP reduces to a condition on the scale of the packing number of $\phi(\ccalX)$ as stated in \eqref{eq:min_gamma2}.  Specifically, as the radius $\frac{K\sqrt{\eta}}{C}$ increases, the packing number of the kernelized feature space decreases, and hence the required model order to fill $\phi(\ccalX)$ decreases.
This radius depends on the constant $K$ which scales the approximation budget selection $\eta$, the learning rate $\eta$, and the constant  $C$ bounding the gradient of the regularized instantaneous loss.

\begin{table*}[]\centering\hfill
\renewcommand{\arraystretch}{1.7}
\caption{Summary of convergence results for different parameter selections.}
\label{tab2}
\begin{tabular}{llll}
\cline{1-4}  
$\qquad$ & $\qquad$ Diminishing & $\ \quad$ Constant &   $ \ $
\\
\cline{1-4} 
Step-size/Learning rate      & $ \ \qquad\eta_t = \ccalO(1/t)$       &   $\qquad \eta_t = \eta>0$   \\  
Sparse Approximation Budget       &$\ \qquad\eps_t=\eta_t^2$       &  $\qquad\eps=\ccalO(\eta^{3/2})$     \\  
Regularization Condition       & $\ \qquad\eta_t < 1/\lambda$       &  $\qquad\eta < 1/\lambda$     \\  
Convergence Result                & $\ \qquad f_t \rightarrow f^*$ a.s.             & $\qquad\liminf_t \| f_t - f^* \| = \ccalO(\sqrt{\eta})$  a.s. \\ 
Model Order Guarantee               & $\qquad$ None            &  $\qquad$Finite  \\ 
\cline{1-4}
\end{tabular}
\end{table*}

We have established that Algorithm \ref{alg:soldd} yields convergent behavior for the problem \eqref{eq:kernel_stoch_opt} in both diminishing and constant step-size regimes. When the learning rate $\eta_t$ satisfies $\eta_t < 1/\lambda$, where $\lambda>0$ is the regularization parameter, and is attenuating such that $\sum_t \eta_t = \infty$ and $\sum_t \eta_t^2<\infty$, i.e., $\eta_t = \ccalO(1/t)$, the approximation budget $\eps_t$ of Algorithm \ref{alg:komp} must satisfy $\eps_t=\eta_t^2$ [cf. \eqref{eq:dim_stepsize}]. Practically speaking, this means that asymptotically the iterates generated by Algorithm \ref{alg:soldd} may have a very large model order in the diminishing step-size regime, since the approximation budget is vanishing as $\eps_t = \ccalO(1/t^2)$. On the other hand, when a constant algorithm step-size $\eta_t = \eta$ is chosen to satisfy $\eta<1/\lambda$, then we only require the constant approximation budget $\eps_t = \eps$ to satisfy $\eps = \ccalO(\eta^{3/2})$. This means that in the constant learning rate regime, we obtain a function sequence which converges to a neighborhood of the optimal $f^*$ defined by \eqref{eq:kernel_stoch_opt} and is guaranteed to have a finite model order. These results are summarized in Table \ref{tab2}.

\begin{remark}(Sparsity of $f^*$)
\normalfont Algorithm \ref{alg:soldd} provides a method to avoid keeping an unnecessarily large number of kernel dictionary elements along the convergence path towards $f^*$ [cf. \eqref{eq:kernel_stoch_opt}], solving the classic scalability problem of kernel methods in stochastic programming. However, if the optimal function admits a low dimensional representation $|\ccalI| << \infty$, then in addition to extracting memory efficient instantaneous iterates, POLK will obtain the optimal function exactly. In Section \ref{sec:experiments}, we illustrate this property via a multi-class classification problem where the data is generated from Gaussian mixture models.
\end{remark}

\section{Experiments}
\label{sec:experiments}
In this section, we evaluate POLK by considering its performance on two supervised learning tasks trained for three streaming data sets.  The specific tasks we consider are those of (a) training a multi-class kernel logistic regressor (KLR), and (b) training a multi-class kernel support vector machine (KSVM).  The three data sets we use are (i) \texttt{multidist}, a synthetic data set we constructed using two-dimensional Gaussian mixture models; (ii) \texttt{mnist}, the MNIST handwritten digits \citep{MNIST}; and (iii) \texttt{brodatz}, image textures drawn from a subset of the Brodatz texture database \citep{Brodatz1966}.  Where possible, we compare our technique with competing methods. Specifically, for the online support vector machine case, we compare with budgeted stochastic gradient descent (BSGD) \cite{wang2012breaking}, which requires a maximum model order \emph{a priori}. For off-line (batch) KLR, we compare with the import vector machine (IVM) \citep{Zhu2005}, a sparse second-order method.  We also compare with the batch techniques of LIBSVM \citep{Chang2011}, applicable to KSVM only, and an L-BFGS solver \citep{Nocedal1980}.

\subsection{Tasks}\label{subsec:tasks}
The tasks we consider are those of multi-class classification, which is a problem that admits approaches based on probabilistic and geometric criteria.  In what follows, we use $\bbx_n \in \ccalX \subset \reals^p$ to denote the $n^\text{th}$ feature vector in a given data set, and $y_n \in \{1,\ldots,C\}$ to denote its corresponding label.

\noindent {\bf Multi-class Kernel Support Vector Machines (Multi-KSVM)}
 The first task we consider is that of training a multi-class kernel support vector machine, in which the merit of a particular regressor is defined by its ability to maximize its classification margin. In particular, define a set of class-specific activation functions $f_c : \ccalX \rightarrow \reals$, and denote them jointly as $\bbf \in \ccalH^C$.  In Multi-KSVM, points are assigned the class label of the activation function that yields the maximum response. KSVM is trained by taking the instantaneous loss $\ell$ to be the multi-class hinge function which defines the margin separating hyperplane in the kernelized feature space, i.e.,
\begin{align}\label{eq:multi_svm}
\ell(\bbf,\bbx_n,y_n) = \max(0,1+f_r(\bbx_n)-f_{y_n}(\bbx_n)) + \lambda \sum_{c'=1}^C \lVert f_{c'} \rVert_{\ccalH}^2 \; ,
\end{align}
where $r = \argmax_{c'\neq y} f_{c'}(\bbx)$. Further details may be found in \cite{Murphy2012}.

\vspace{0.1cm}

\noindent {\bf Multi-class Kernel Logistic Regression (Multi-KLR)}
The second task we consider is that of kernel logistic regression, wherein, instead of maximizing the margin which separates sample points in the kernelized feature space, we instead adopt a probabilistic model on the odds ratio that a sample point has a specific label relative to all others. Using the same notation as above for the class-specific activation functions, we adopt the probabilistic model:
\begin{align}\label{eq:multi_logistic_prob}
P(y = c \,|\, \bbx) \triangleq \frac{\exp(f_c(\bbx))}{\sum_{c^\prime} \exp( f_{c^\prime}(\bbx))}.
\end{align}
which models the odds ratio of a given sample point being in class $c$ versus all others. We use the negative log likelihood pertaining to the above model as the instantaneous loss (see, e.g., \cite{Murphy2012}), i.e.,
\begin{align}\label{eq:multi_logistic}
\ell(\bbf,\bbx_n,y_n) &= - \log P(y = y_n | \bbx_n) + \frac{\lambda}{2} \sum_{c} \| f_c \|_{\ccalH}^2 \nonumber \\
&= \log \left(\sum_{c^\prime} \exp(f_{c^\prime}(\bbx_n))\right) - f_{y_n}(\bbx_n) + \frac{\lambda}{2} \sum_{c} \| f_c \|_{\ccalH}^2 \; .
\end{align}
Observe that the loss \eqref{eq:multi_logistic} substituted into the empirical risk minimization problem in Example \ref{eg1} is its generalization to multi-class classification. For a given set of activation functions, the classification decision $\tilde{c}$ for $\bbx$ is given by the class that yields the maximum likelihood, i.e., $\tilde{c}=\argmax_{c\in\{1,\dots,C\}} f_c(\bbx)$.

\vspace{0.1cm}

\begin{figure}[!]
	\centering
	\subfigure[!][\texttt{multidist} data]{\includegraphics[width=0.32\linewidth,height=4.4cm,]{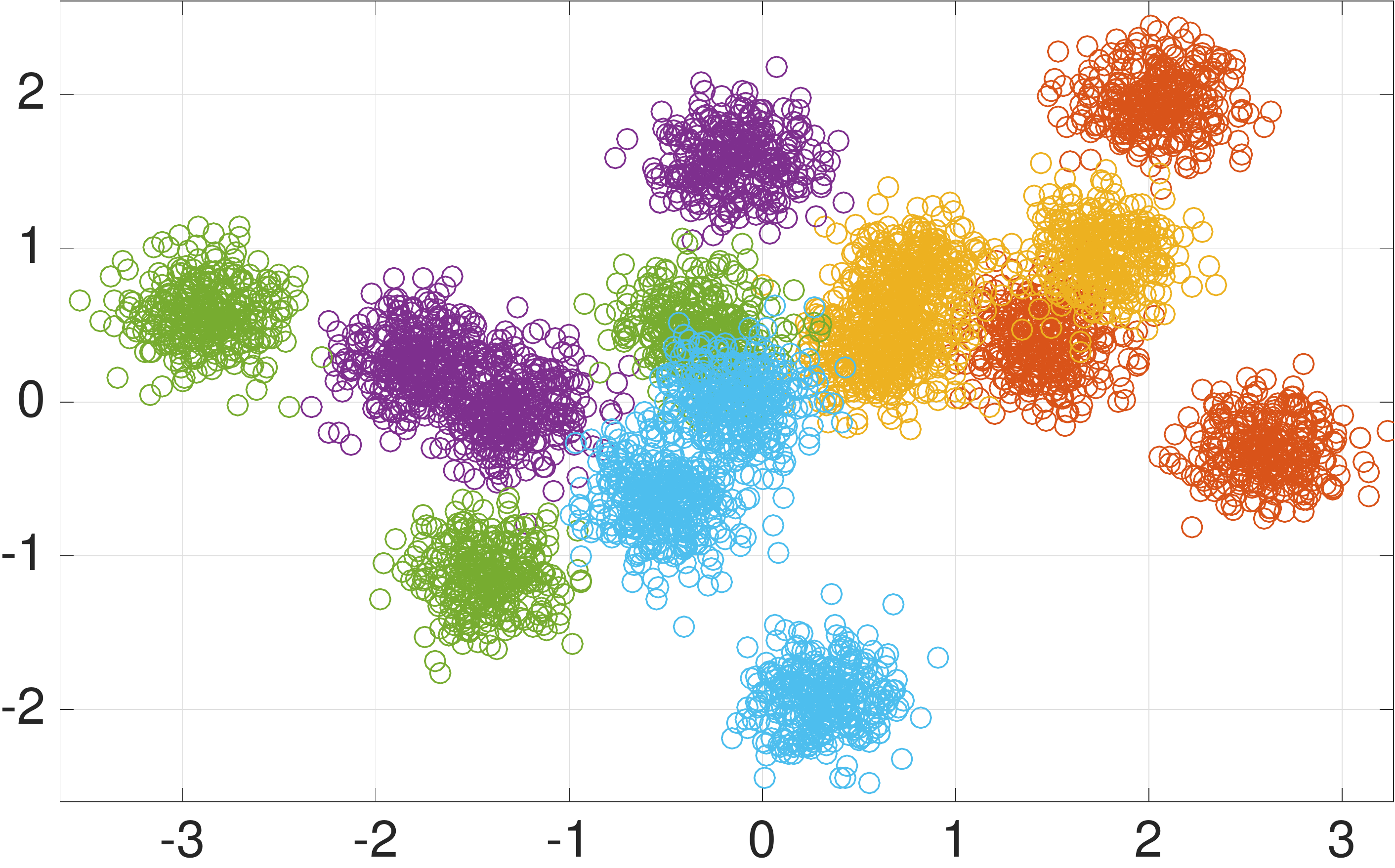}\label{subfig:gmm_pointcloud}}
	\subfigure[\texttt{brodatz} example texture]{\includegraphics[width=0.3\linewidth,height=4.4cm]{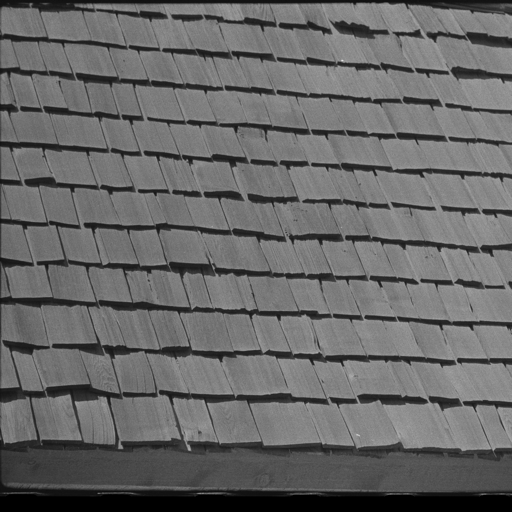}\label{subfig:brodatz2}}
	\subfigure[\texttt{mnist} examples]{\includegraphics[width=0.3\linewidth, height=4.4cm]{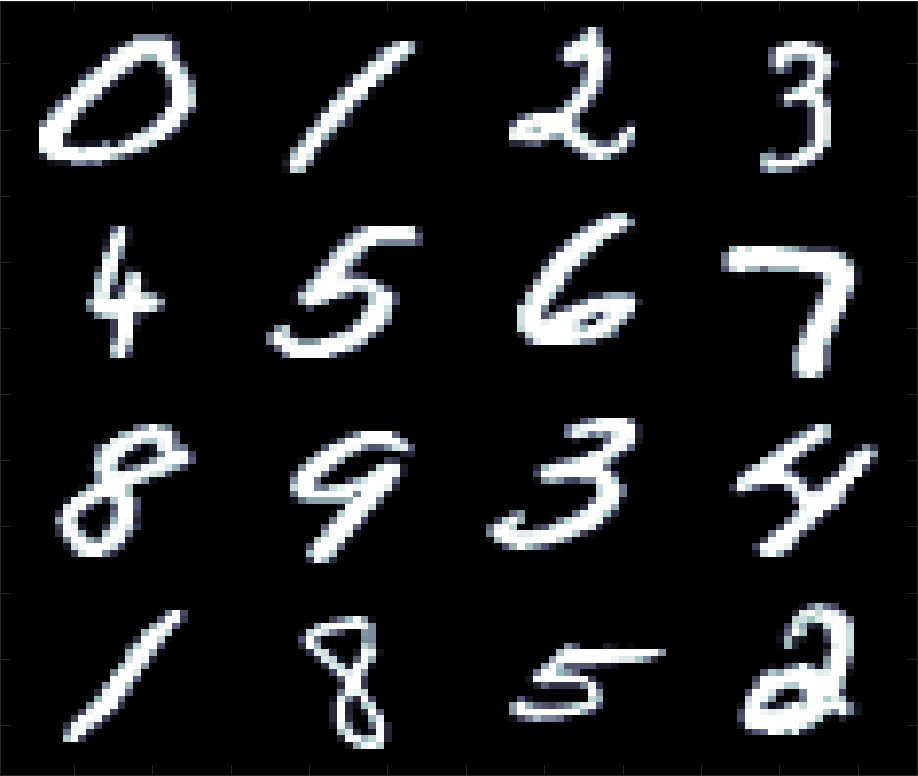}\label{subfig:mnist}}
	\caption{Visualizations of the data sets used in experiments.}
	\label{fig:ex_data}\vspace{-4mm}
\end{figure}\vspace{-4mm}

\subsection{Data Sets}\label{subsec:datasets}
We evaluate Algorithm \ref{alg:soldd} for the Multi-KLR and Multi-KSVM tasks described above using the \texttt{multidist}, \texttt{mnist}, \texttt{brodatz} data sets.

\vspace{0.2cm}

\noindent{\bf \texttt{multidist}} \\
In a manner similar to \citep{Zhu2005}, we generate the \texttt{multidist} data set using a set of Gaussian mixture models.  The data set consists $N=5000$ feature-label pairs for training and $2500$ for testing.  Each label $y_n$ was drawn uniformly at random from the label set. The corresponding feature vector $\bbx_n \in \reals^p$ was then drawn from a planar ($p=2$), equitably-weighted Gaussian mixture model, i.e., $\bbx \given y \; \sim \; (1/3) \sum_{j=1}^3 \ccalN(\boldsymbol{\mu}_{y,j}, \sigma^2_{y,j}\bbI)$ where $\sigma^2_{y,j}=0.2$ for all values of $y$ and $j$. The means $\boldsymbol{\mu}_{y,j}$ are themselves realizations of their own Gaussian distribution with class-dependent parameters, i.e., $\boldsymbol{\mu}_{y,j} \sim \ccalN( \boldsymbol{\theta}_y, \sigma^2_y\bbI )$, where $\left\{ \bbtheta_1,\ldots,\bbtheta_C\right\}$ are equitably spaced around the unit circle, one for each class label, and $\sigma_y^2=1.0$. We fix the number of classes $C=5$, meaning that the feature distribution has, in total, $15$ distinct modes.  The data points are plotted in Figure \ref{subfig:gmm_pointcloud}.

\vspace{0.2cm}

\noindent{\bf \texttt{mnist}} \\
The \texttt{mnist} data set we use is the popular MNIST data set \citep{MNIST}, which consists of $N=60000$ feature-label pairs for training and $10000$ for testing.  Feature vectors are $p=784$-dimensional, where each dimension captures a single grayscale pixel value (scaled to lie within the unit  interval) that corresponds to a unique location in a $28$-pixel-by-$28$-pixel image of a cropped, handwritten digit.  Labels indicate which digit is written, i.e., there are $C=10$ classes total, corresponding to digits $0,\dots,9$ -- examples are given in Figure \ref{subfig:mnist}.

\vspace{0.2cm}

\noindent{\bf \texttt{brodatz}} \\
We generated the \texttt{brodatz} data set using a subset of the images provided in \cite{Brodatz1966}.  Specifically, we used $13$ texture images (i.e., $C=13$), and from them generated a set of $256$ textons \citep{Leung1999}.  Next, for each overlapping patch of size $24$-pixels-by-$24$-pixels within these images, we took the feature to be the associated $p=256$-dimensional texton histogram.  The corresponding label was given by the index of the image from which the patch was selected.  When then randomly selected $N=10000$ feature-label pairs for training and $5000$ for testing.  An example texture image can be seen in Figure \ref{subfig:brodatz2}.

\subsection{Results}\label{subsec:results}
For each task and data set described above, we implemented POLK (Algorithm \ref{alg:soldd}) along with the competing methods described at the beginning of the section. For some of the tasks, only a subset of the competing methods are applicable, and in some cases such as online logistic regression, none are. Here, we shall describe the details of each experimental setting and the corresponding results.

\vspace{0.2cm}

\noindent{\bf \texttt{multidist} Results }\\
Due to the small size of our synthetic \texttt{multidist} data set, we were able to generate results for the Multi-KSVM task using each of the methods specified earlier except for IVM.  For POLK, we used the following specific parameter values: we select the Gaussian/RBF kernel with bandwidth $\tilde{\sigma}^2 = 0.6$, constant learning rate $\eta = 6.0$, parsimony constant $K \in \{10^{-4}, 0.04\}$, and regularization constant $\lambda = 10^{-6}$.  Further, we processed streaming samples in mini-batches of size $32$.  For BSGD, we used the same $\tilde{\sigma}^2$ and $\lambda$, but achieved the best results with smaller constant learning rate $\eta = 1.0$ (perhaps due, in part, to the fact that BSGD does not support mini-batching).  In order to compare with POLK, we set BSGD's pre-specified model orders to be $\{16, 129\}$, i.e., the steady-state model orders of POLK parameterized with the values of $K$ specified above.

\begin{figure}
	\centering
	\subfigure[Empirical risk $R(\bbf_t)$]{\includegraphics[width=0.32\linewidth,height=3.5cm]{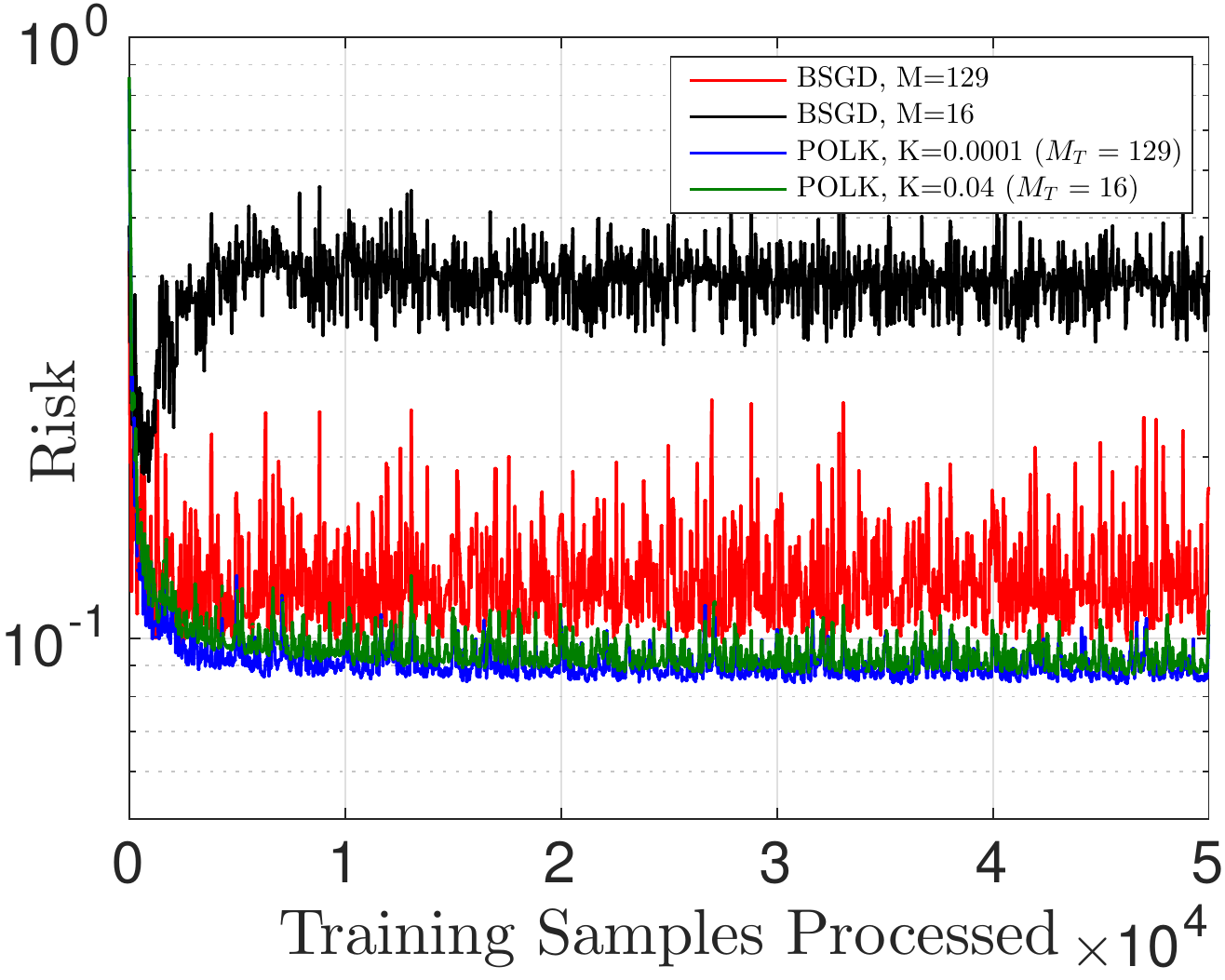}\label{subfig:objective_gmm_ksvm}}	
	\subfigure[Error rate]{\includegraphics[width=0.32\linewidth,height=3.5cm]{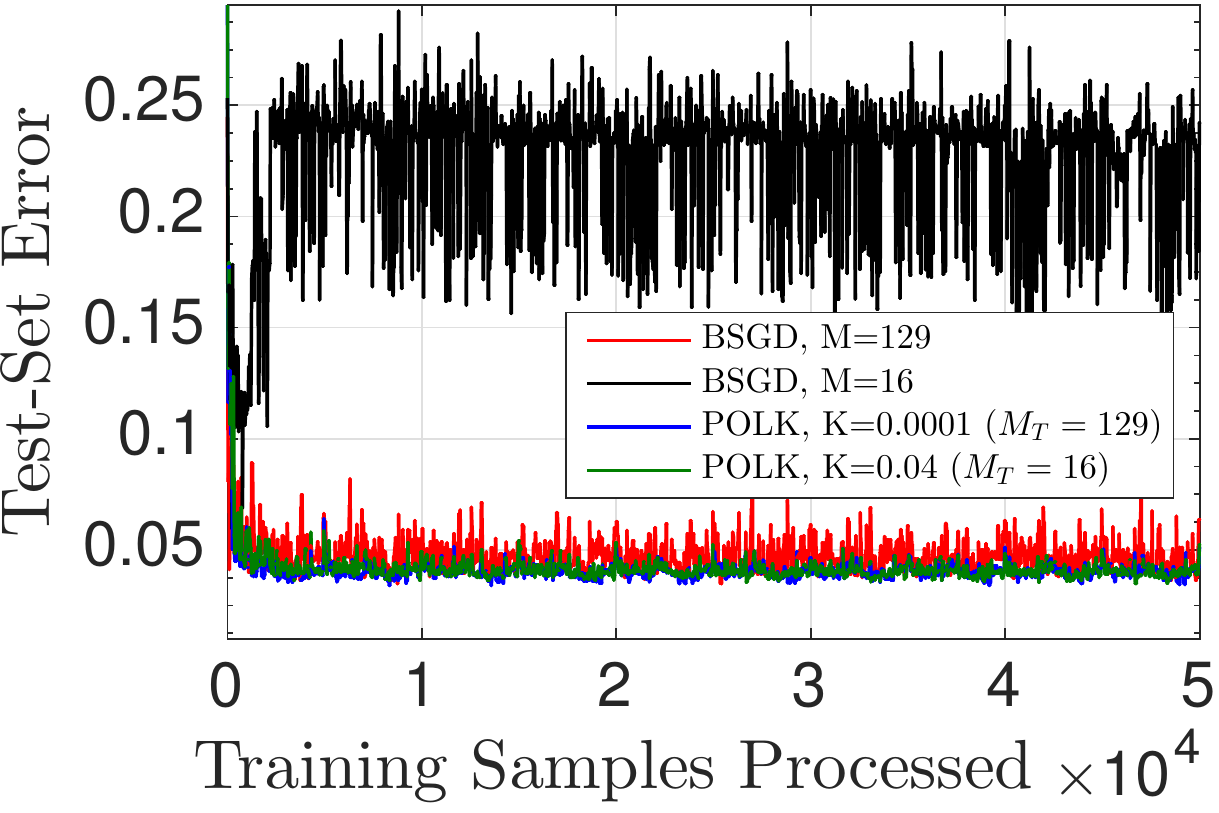}\label{subfig:error_gmm_ksvm}}
	\subfigure[Model order $M_t$]{\includegraphics[width=0.32\linewidth,height=3.5cm]{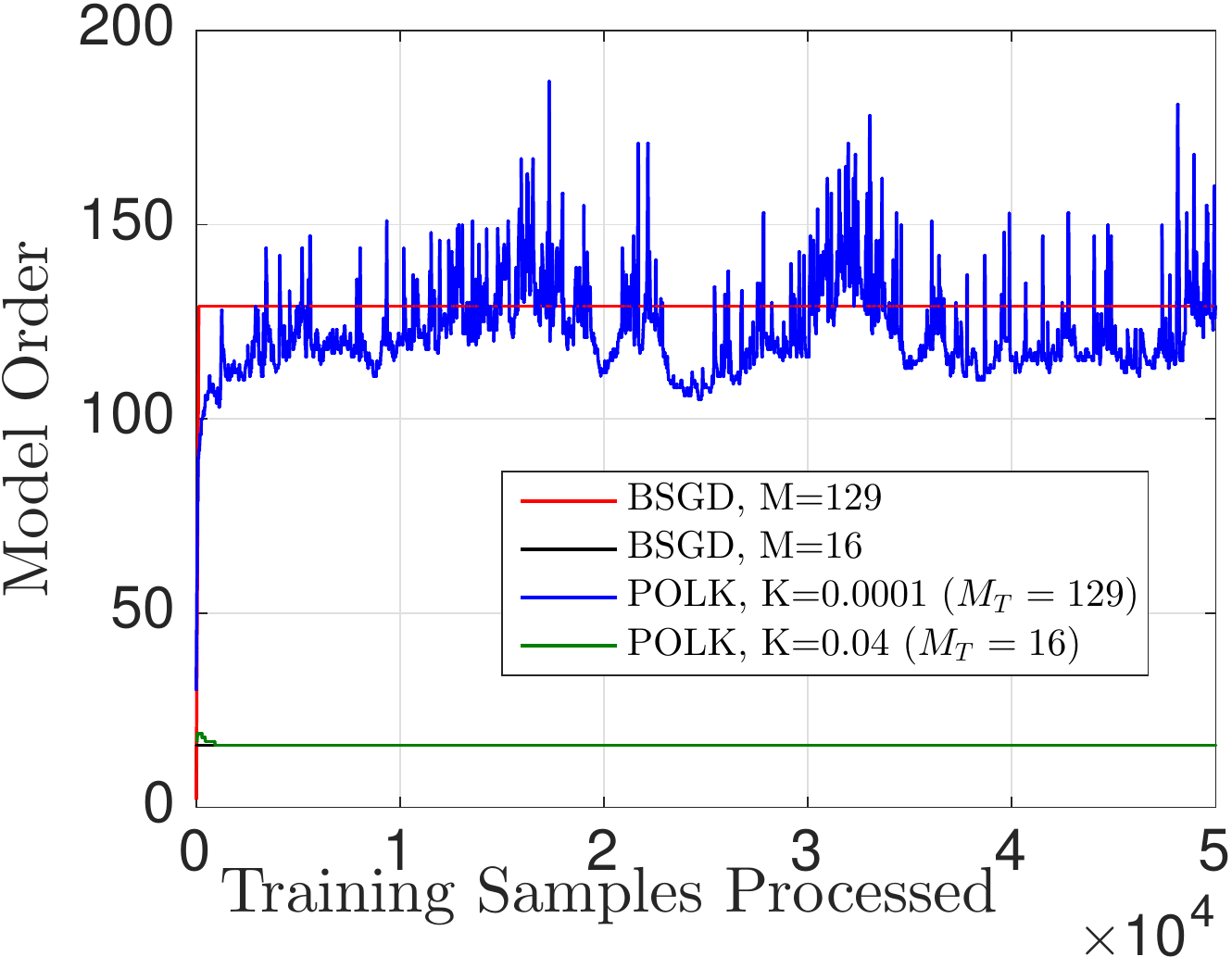}\label{subfig:order_gmm_ksvm}}
	\caption{Comparison of POLK and BSGD on the \texttt{multidist} data set for the Multi-KSVM task. Observe that POLK achieves lower risk and higher accuracy for a fixed model order. More accurate POLK regressors require use of a smaller parsimony constant $K$, although we observe diminishing benefit of increasing the model order via reducing $K$.}\vspace{-4mm}
	\label{fig:ksvm_gmm}
\end{figure}

\begin{figure}
	\centering 
	\subfigure[$\bbf_T$ (hinge loss)]{\includegraphics[width=0.4\linewidth,height=3.5cm]{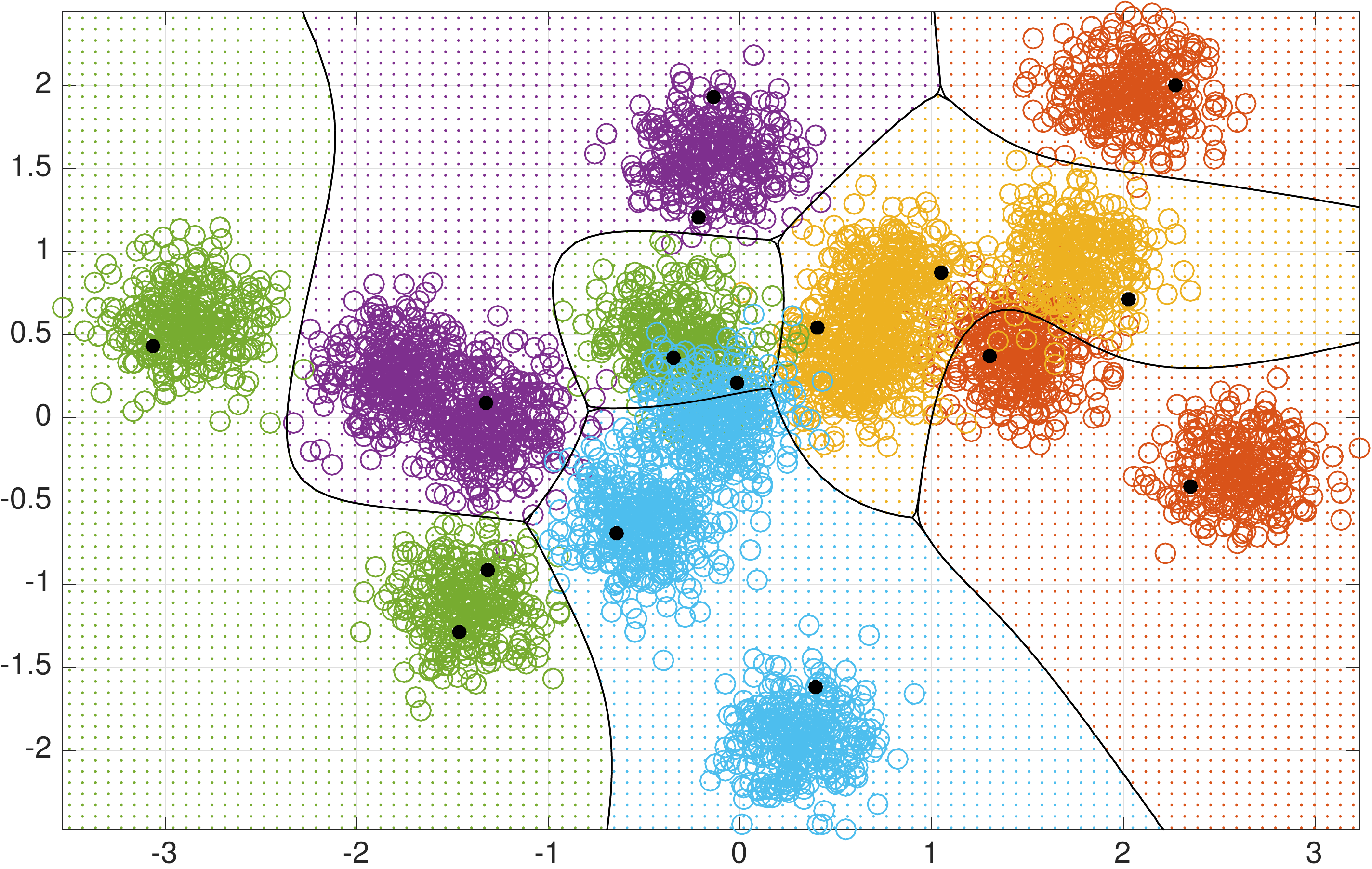}\label{subfig:decision_svm}} \hspace{8mm}
	\subfigure[$\bbf_T$ (logistic loss)]{\includegraphics[width=0.4\linewidth,height=3.5cm]{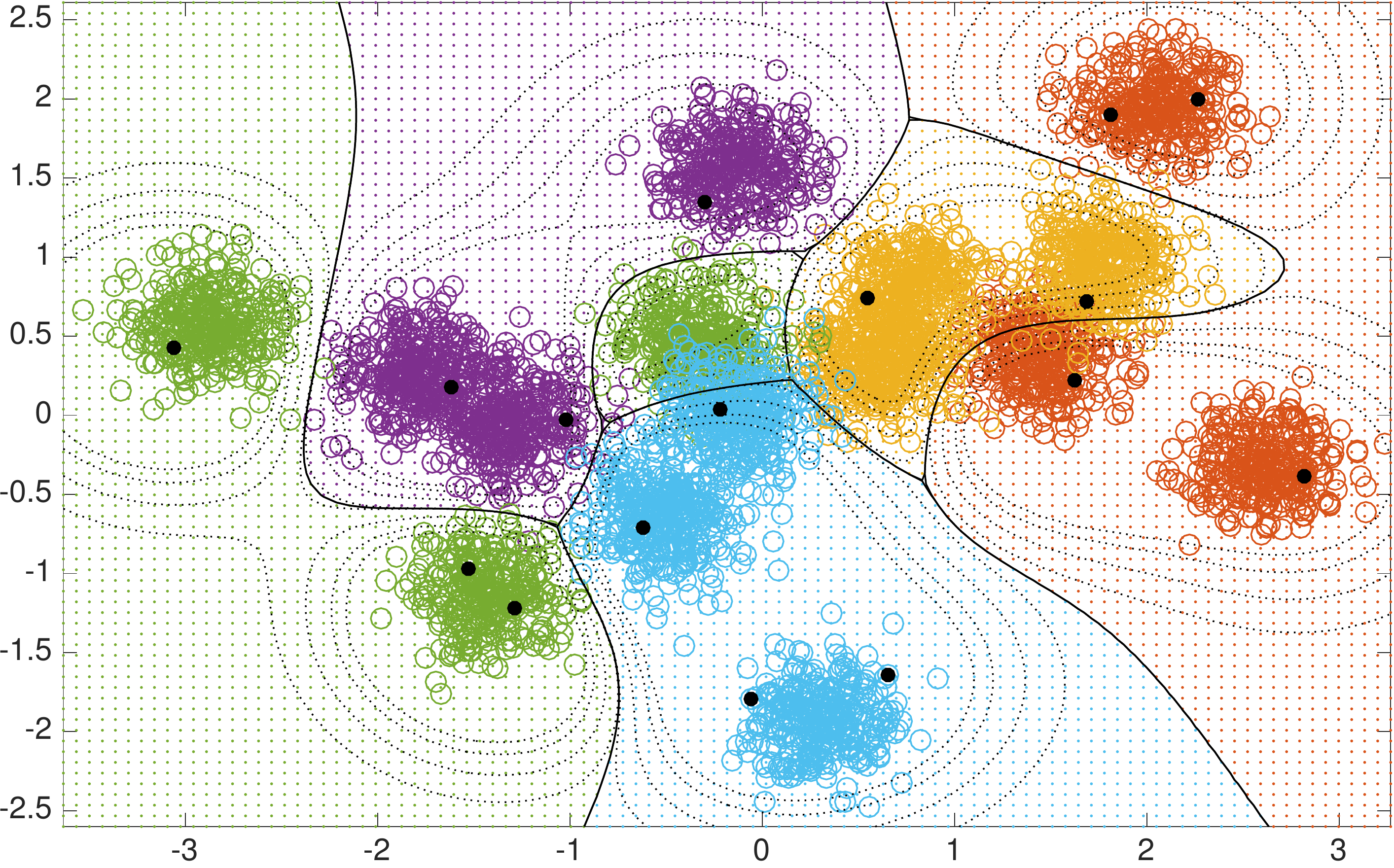}\label{subfig:decision_klr}}\hspace{-1.5cm}
	\caption{Visualization of the decision surfaces yielded by POLK for the Multi-KSVM and Multi-Logisitic tasks on the \texttt{multidist} data set. Training examples from distinct classes are assigned a unique color. Grid colors represent the classification decision by $\bbf_T$. Bold black dots are kernel dictionary elements, which concentrate at the modes of the joint data distribution. Solid lines are drawn to denote class label boundaries, and additional dashed lines in \ref{subfig:decision_klr} are drawn to denote confidence intervals.}
	\label{fig:decision}\vspace{-4mm}
\end{figure}

In Figure \ref{fig:ksvm_gmm} we plot the empirical results of this experiment for POLK and BSGD, and observe that POLK outperforms the competing method by an order of magnitude in terms of objective evaluation (Fig. \ref{subfig:objective_gmm_ksvm}) and test-set error rate (Fig \ref{subfig:error_gmm_ksvm}). Moreover, because the marginal feature density of \texttt{multidist} contains $15$ modes, the optimal model order is $M^*=15$, which is approximately \emph{learned} by POLK for $K = 0.04$ (i.e., $M_T=16$) (Fig. \ref{subfig:order_gmm_ksvm}).  The corresponding trial of BSGD, on the other hand, initialized with this parameter, does not converge. Observe that for this task POLK exhibits a state of the art trade off between test set accuracy and number of samples processed -- reaching below $4\%$ error after only $1249$ samples. The final decision surface $\bbf_T$ of this trial of POLK is shown in Fig. \ref{subfig:decision_svm}, where it can be seen that the selected kernel dictionary elements concentrate near the modes of the marginal feature density. 

We can also see from Table \ref{tab:multidist} that POLK compares favorably to the batch techniques for Multi-KSVM on the \texttt{multidist} data set.  It achieves approximately the same error rate as LIBSVM with significantly fewer model points (support vectors) and even outperforms our (dense) L-BFGS batch solver in terms of test-set error, while adding the ability to process data in an online fashion.

\begin{figure}
	\centering 
	\subfigure[Empirical risk $R(\bbf_t)$]{\includegraphics[width=0.32\linewidth,height=3.5cm]{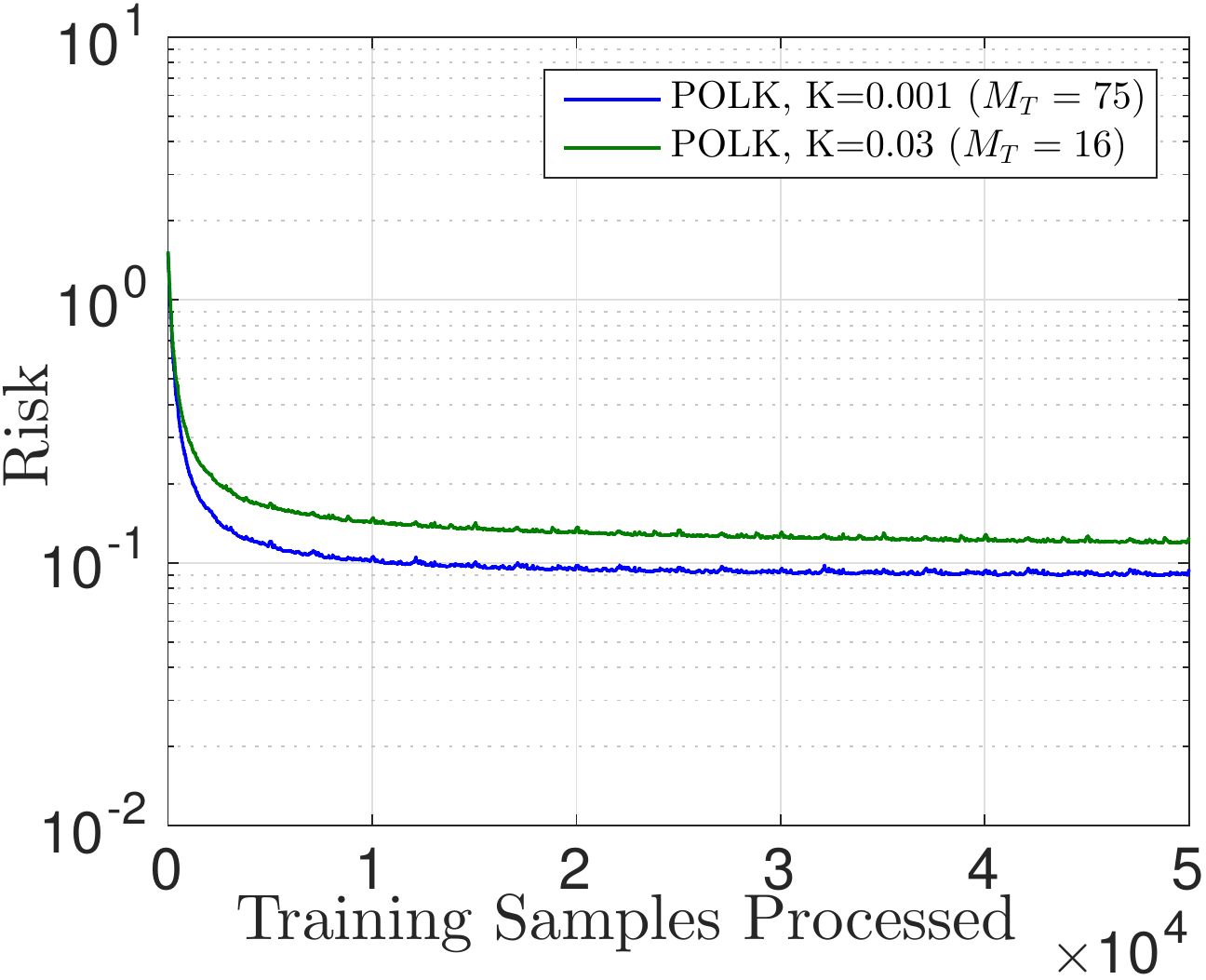}\label{subfig:objective_gmm_klr}}	
	\subfigure[Error rate]{\includegraphics[width=0.32\linewidth,height=3.5cm]{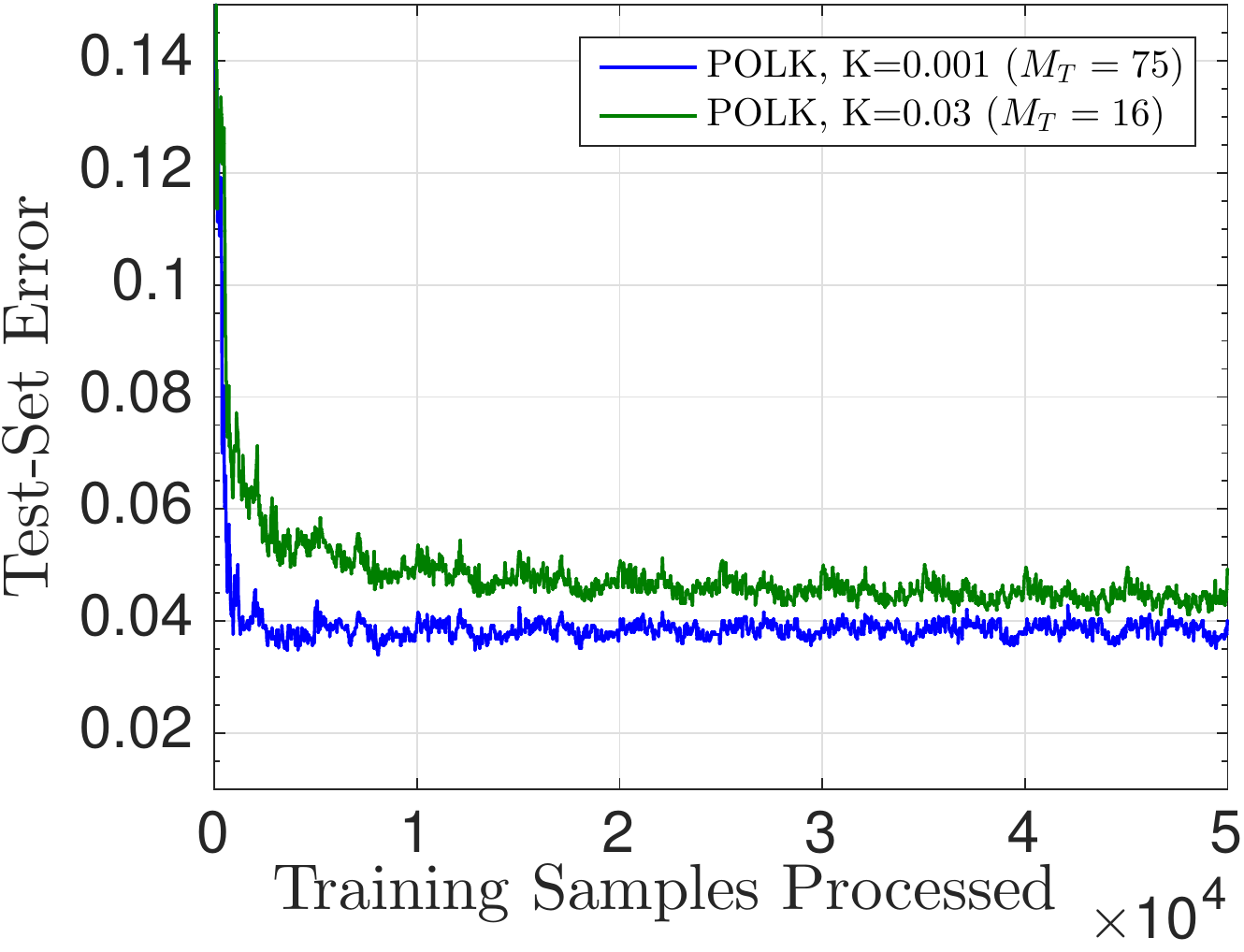}\label{subfig:error_gmm_klr}}
	\subfigure[Model order $M_t$]{\includegraphics[width=0.32\linewidth,height=3.5cm]{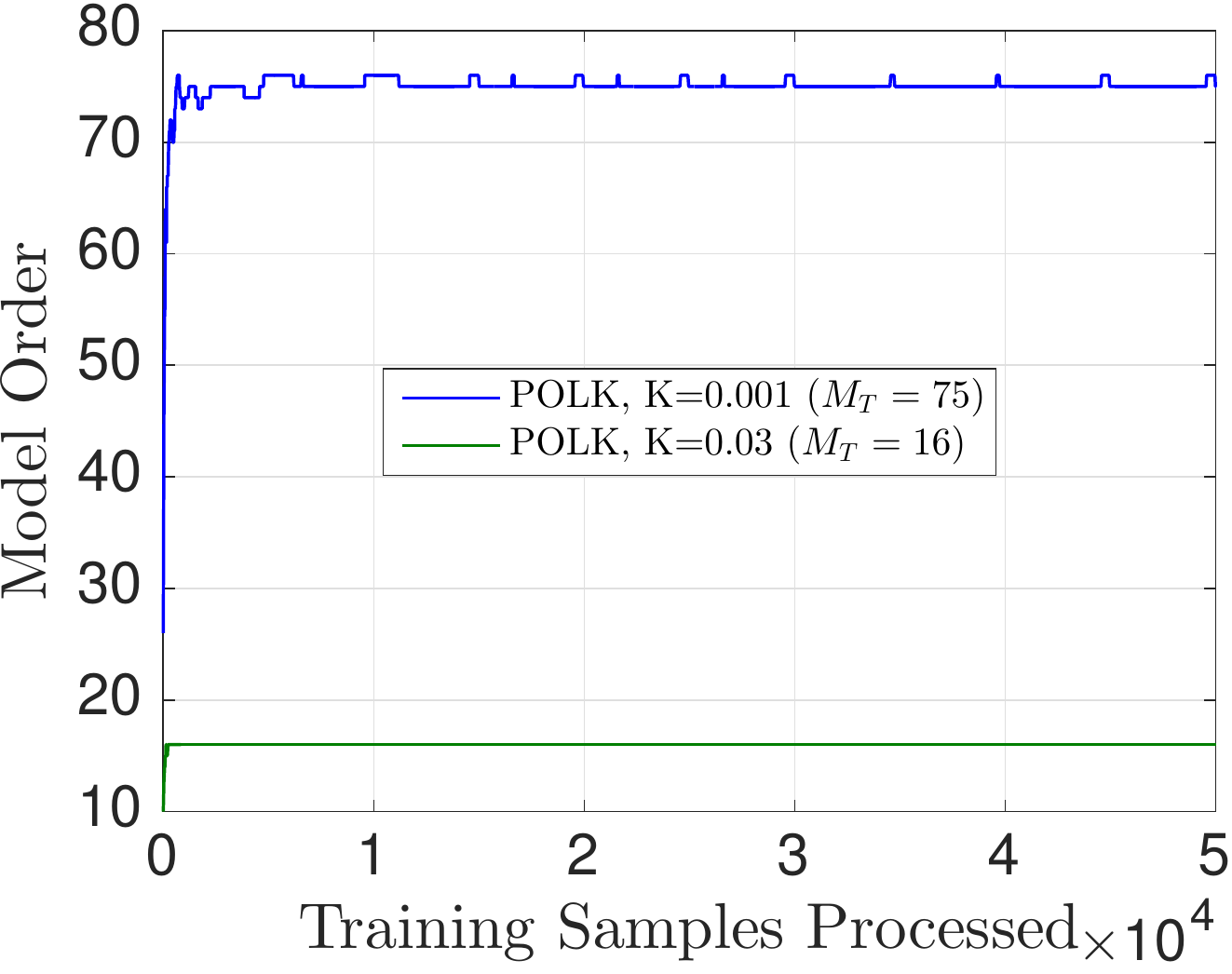}\label{subfig:order_gmm_klr}}
	\caption{Empirical behavior of the POLK algorithm applied to the \texttt{multidist} data set for the Multi-Logistic task. Observe that the algorithm converges to a low risk value ($R(f_t)<10^{-1}$) and achieves test set accuracy between $4\%$ and $5\%$ depending on choice of parsimony constant $K$, which respectively corresponds to a model order between $75$ and $16$.}\vspace{-4mm}
	\label{fig:klr_gmm}\vspace{-4mm}
\end{figure}

\begin{table*}[t]\centering
\begin{tabular}{ | c | c | c |}
\hline
                   & \multicolumn{2}{|c|}{\texttt{multidist}}  \\
\hline
\textbf{Algorithm}                    & Multi-KSVM                      & Multi-Logistic    \\
                   & \tiny{(risk/error/model order)} & \tiny{(risk/error/model order)} \\
\hline
LIBSVM 			   & $-/3.92/656$                   & $-/-/-$ \\ 
\hhline{|=|==|}
L-BFGS             & $0.0854/4.08/5000$             & $0.0854/4.04/5000$ \\
\hhline{|=|==|}
IVM                & $-/-/-$                        & $0.0894/4.08/16$ \\
\hhline{|=|==|}
BSGD               & $0.385/21.8/16$                & $-/-/-$ \\
\hline
POLK               & $0.0919/3.98/16$               & $0.120/4.36/16$ \\
\hline
\end{tabular}
\caption{Comparison of POLK, BSGD, IVM, L-BFGS, and LIBSVM results on the \texttt{multidist} data set.  Reported risk and error values for POLK and BSGD were averaged over the final 5\% of processed training examples.  Dashes indicate where the method could not be used to generate results because it is not defined for that task.  LIBSVM is used as a baseline, but note that it uses a fundamentally different model for multi-class problems (a separate one-vs-all classifier is trained for each class, and then at test time, a majority vote is executed), and so a comparable risk value can not computed.}
\label{tab:multidist} \vspace{-5mm}
\end{table*}

For the Multi-Logisitic task on this data set, we were able to generate results for each method except BSGD and LIBSVM, which are specifically tailored to the SVM task.  For POLK, we used the following parameter values: Gaussian kernel with bandwidth $\tilde{\sigma}^2 = 0.6$, constant learning rate $\eta = 6.0$, parsimony constant $K \in \{0.001, 0.03\}$, and regularization constant $\lambda = 10^{-6}$.  As in Multi-KSVM, we processed the streaming samples in mini-batches of size $32$.  The empirical behavior of POLK for the Multi-Logistic task can be seen in Figure \ref{fig:klr_gmm} and the final decision surface is presented in Figure \ref{subfig:decision_klr}. Observe that POLK is exhibits comparable convergence to the SVM problem, but a smoother descent due to the differentiability of the multi-logistic loss. In Table \ref{tab:multidist} we present final accuracy and risk values on the logistic task, and note that it performs comparably, or in some cases, favorably, to the batch techniques (IVM and L-BFGS), while processing streaming data.

\begin{figure}
	\centering 
	\subfigure[Expected risk $R(\bbf_t)$]{\includegraphics[width=0.32\linewidth,height=3.5cm]{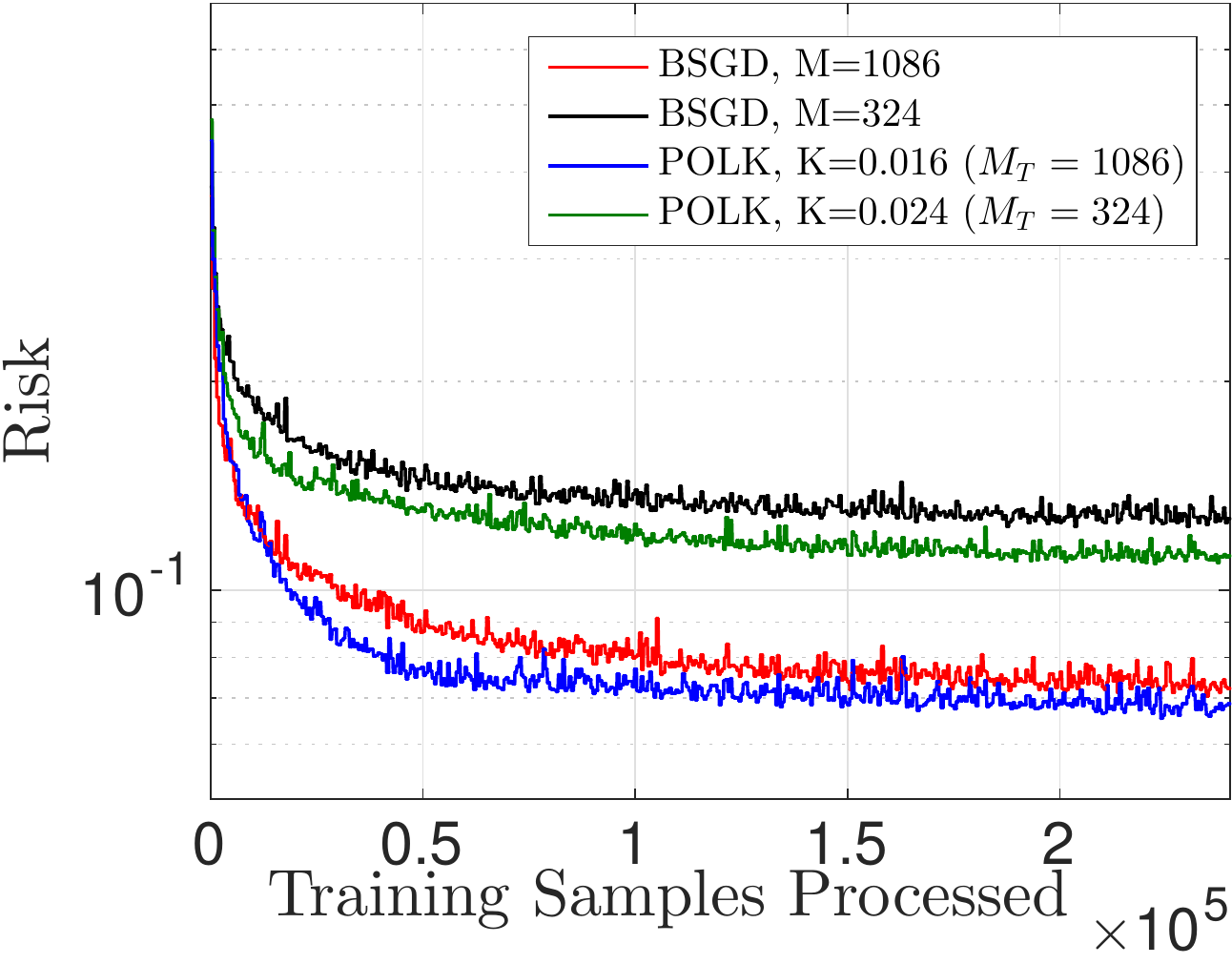}\label{subfig:objective_mnist_ksvm}}
	\subfigure[Error rate]{\includegraphics[width=0.32\linewidth,height=3.5cm]{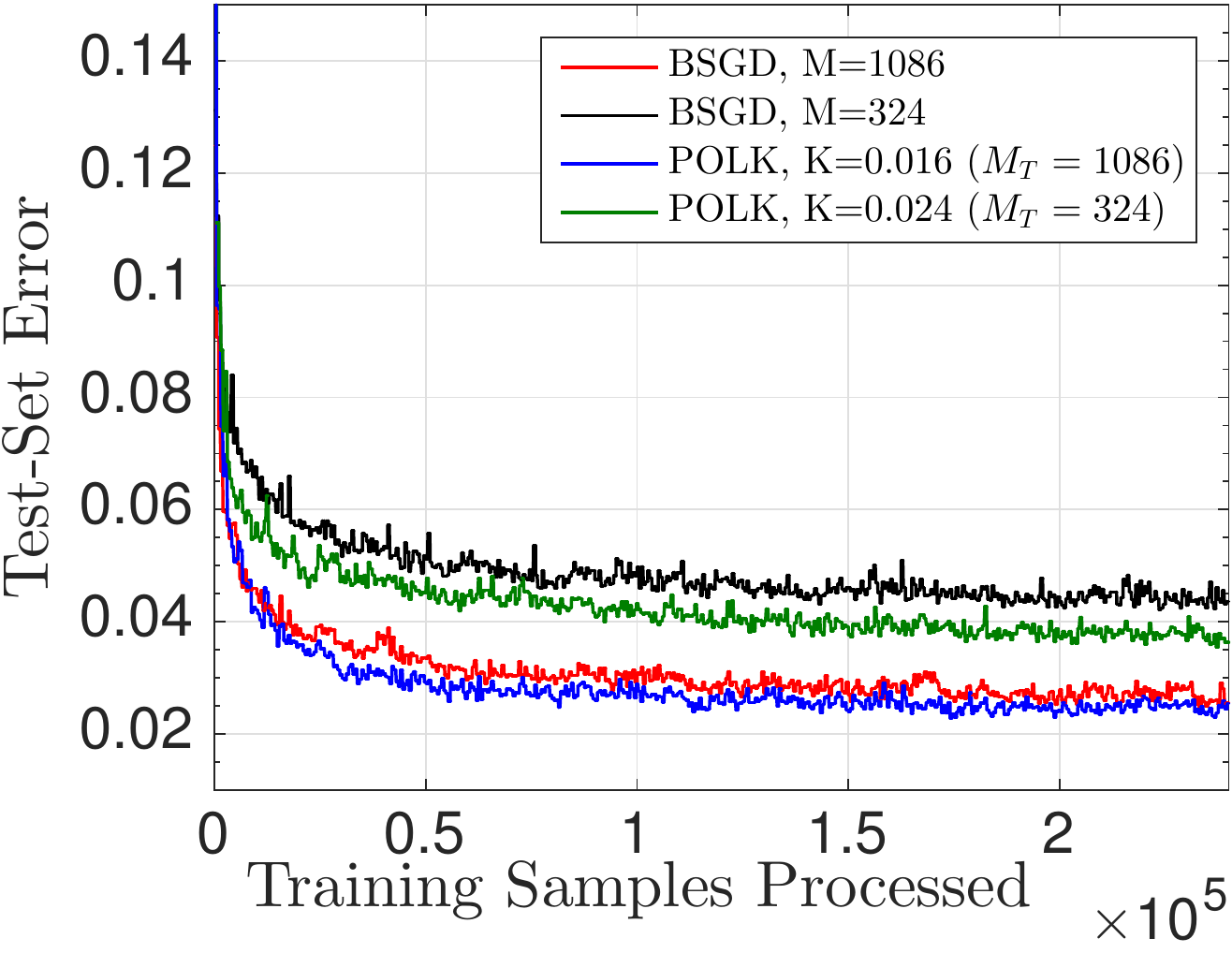}\label{subfig:error_mnist_ksvm}}
	\subfigure[Model order $M_t$]{\includegraphics[width=0.32\linewidth,height=3.5cm]{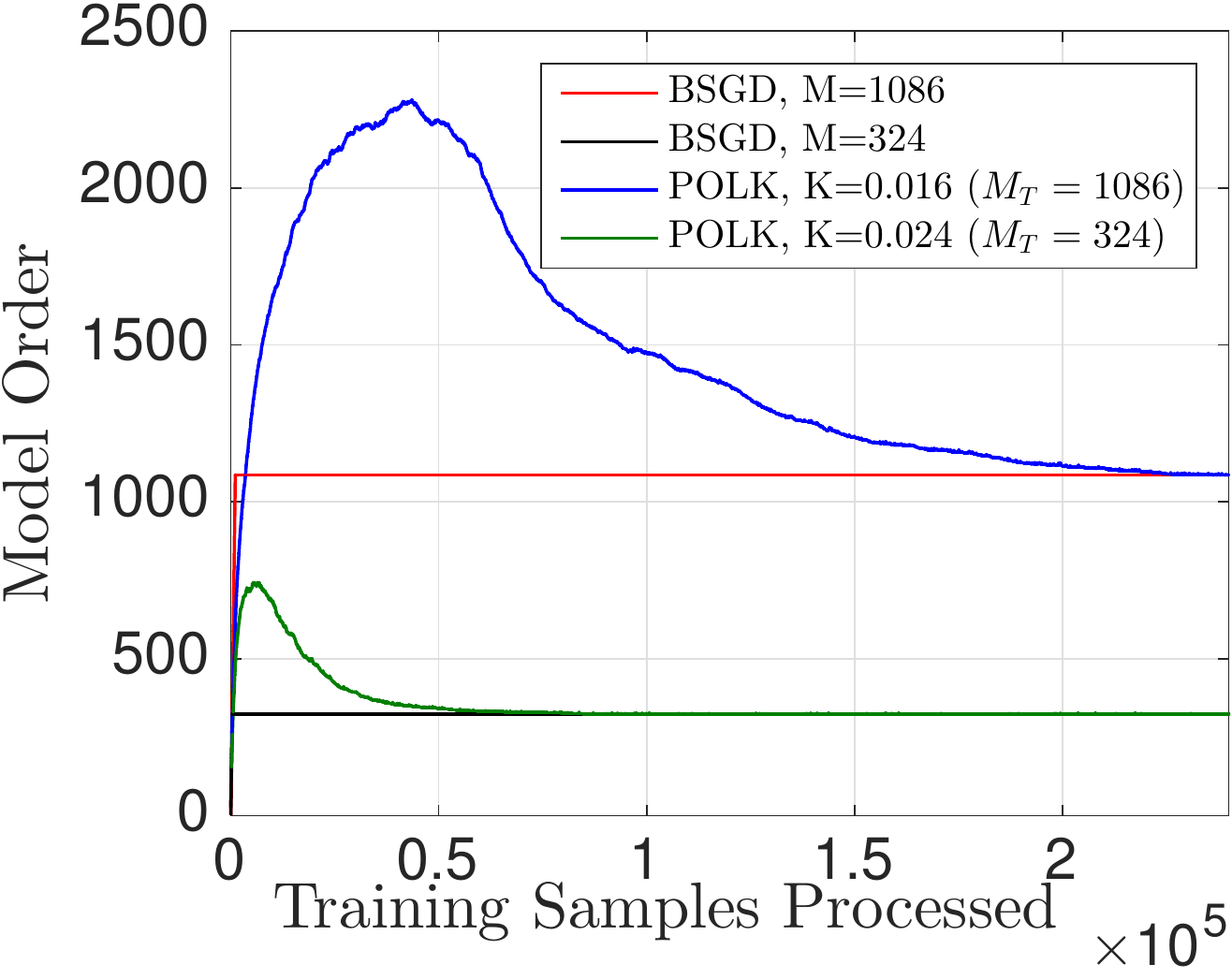}\label{subfig:order_mnist_ksvm}}
	\caption{Comparison of POLK and BSGD on \texttt{mnist} data set for the Multi-KSVM task. Observe that POLK achieves lower risk and higher accuracy on this task, and extracts a model order directly from the feature space that yields convergence.}
	\label{fig:ksvm_mnist}\vspace{-5mm}
\end{figure}

\begin{figure}
	\centering 
	\subfigure[Expected risk $R(\bbf_t)$]{\includegraphics[width=0.32\linewidth,height=3.5cm]{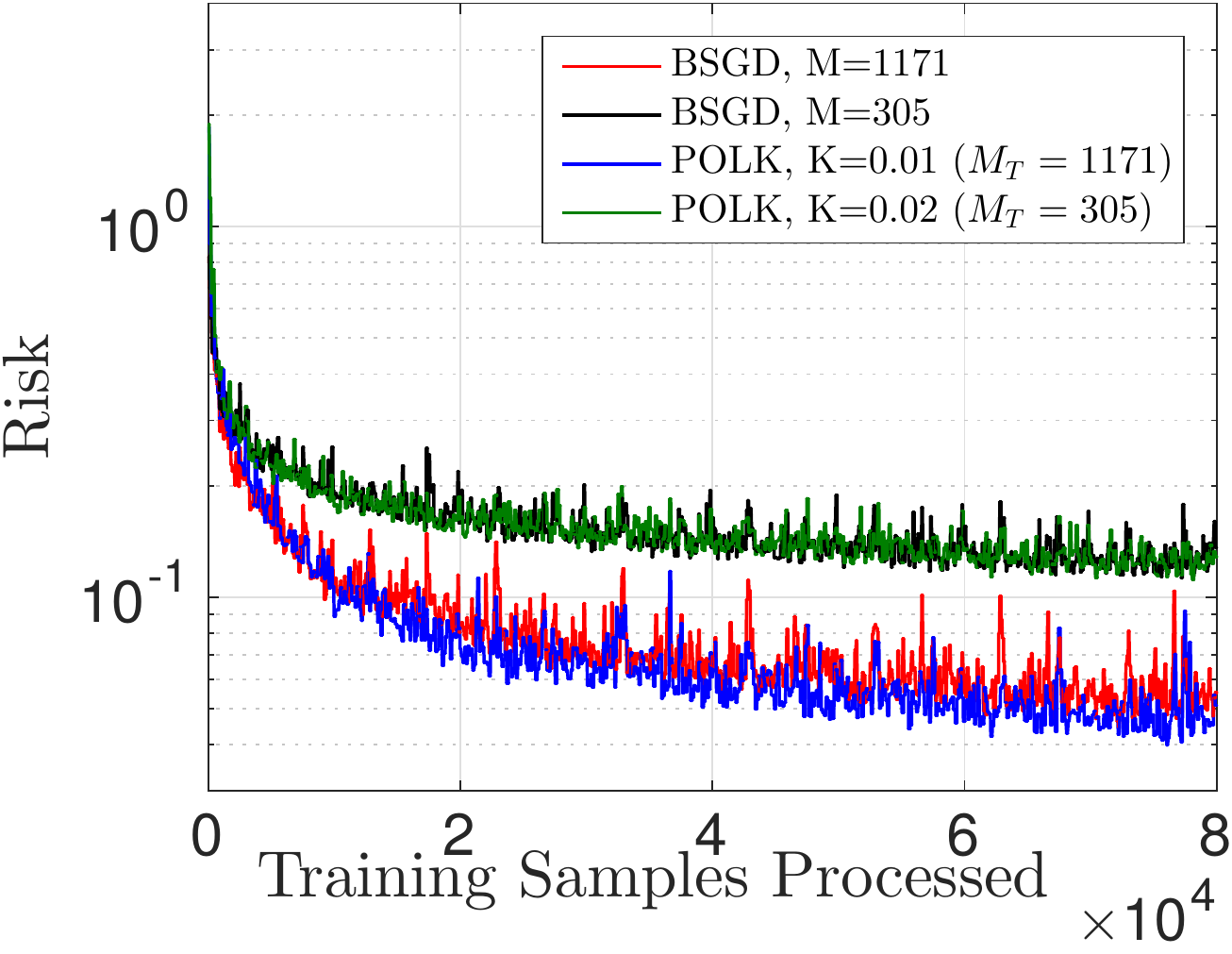}\label{subfig:objective_brodatz_ksvm}}	
	\subfigure[Error rate]{\includegraphics[width=0.32\linewidth,height=3.5cm]{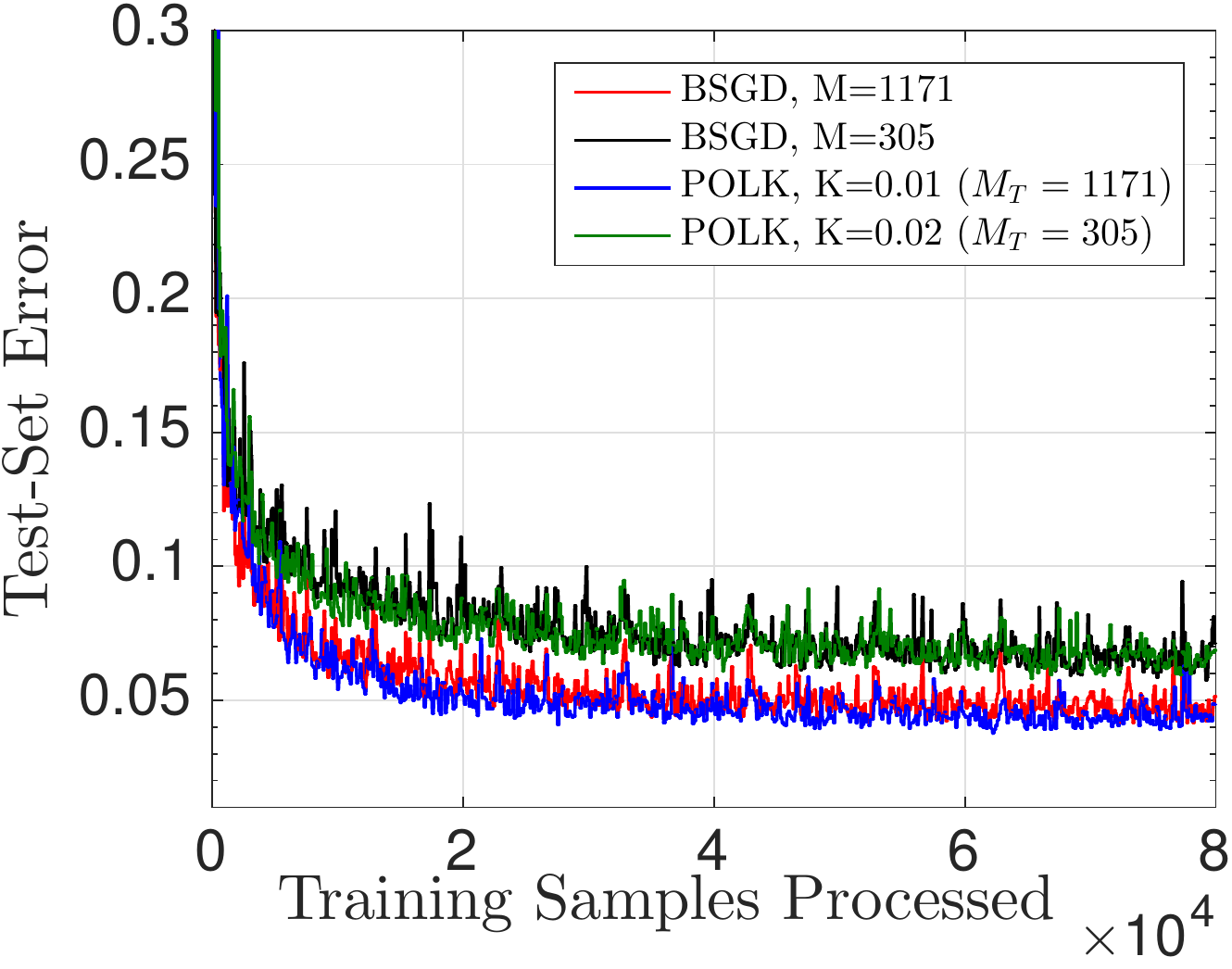}\label{subfig:error_brodatz_ksvm}}
	\subfigure[Model order $M_t$]{\includegraphics[width=0.32\linewidth,height=3.5cm]{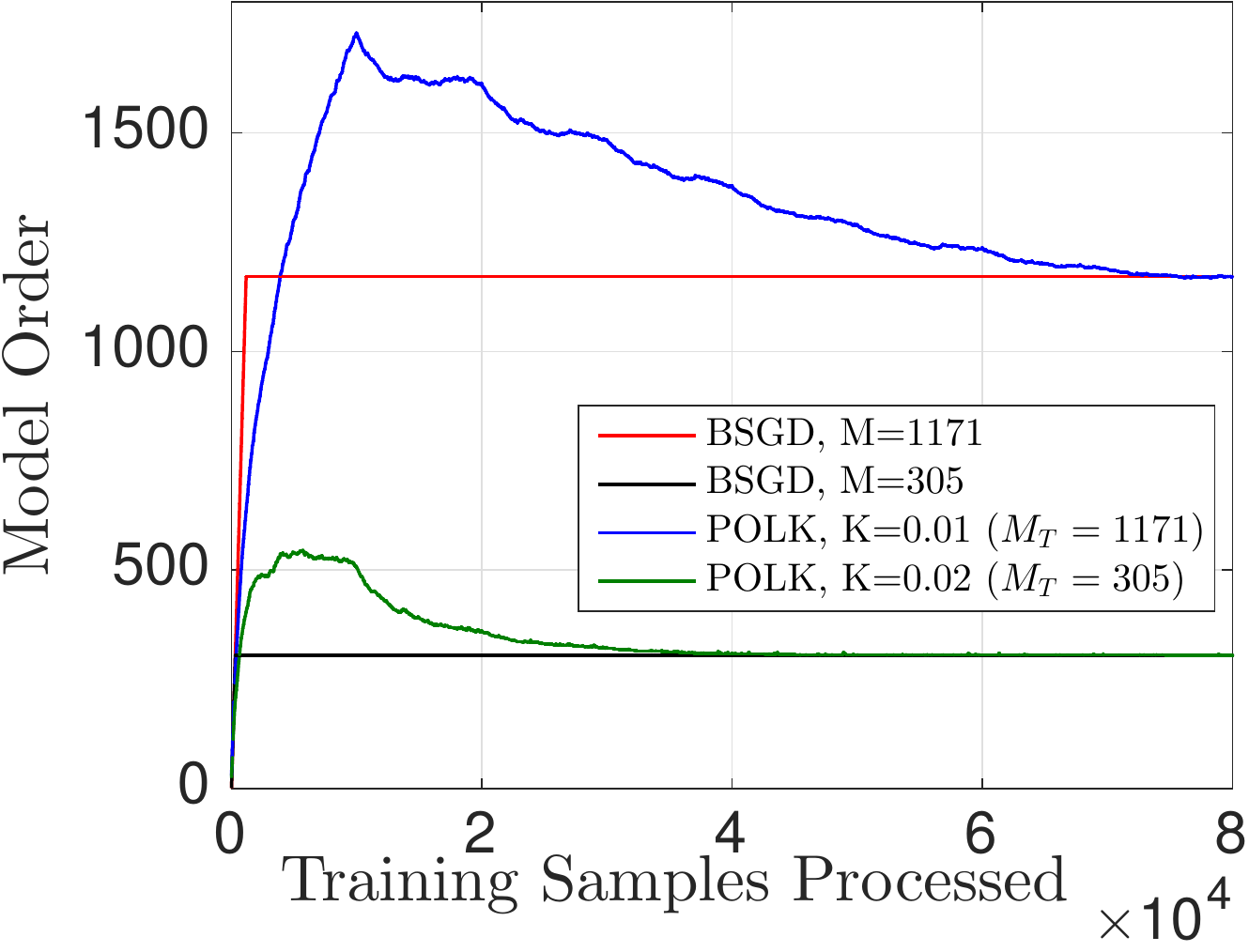}\label{subfig:order_brodatz_ksvm}}
	\caption{Comparison of POLK and BSGD on  \texttt{brodatz} data set for the Multi-KSVM task. We observe that POLK behaves similarly to BSGD for this task, stabilizing at an accuracy near $96\%$. For this dense data domain, larger model orders are needed to achieve convergence. }
	\label{fig:ksvm_brodatz}\vspace{-5mm}
\end{figure}

\begin{figure}
	\centering 
	\subfigure[Expected risk $R(\bbf_t)$]{\includegraphics[width=0.32\linewidth,height=3.5cm]{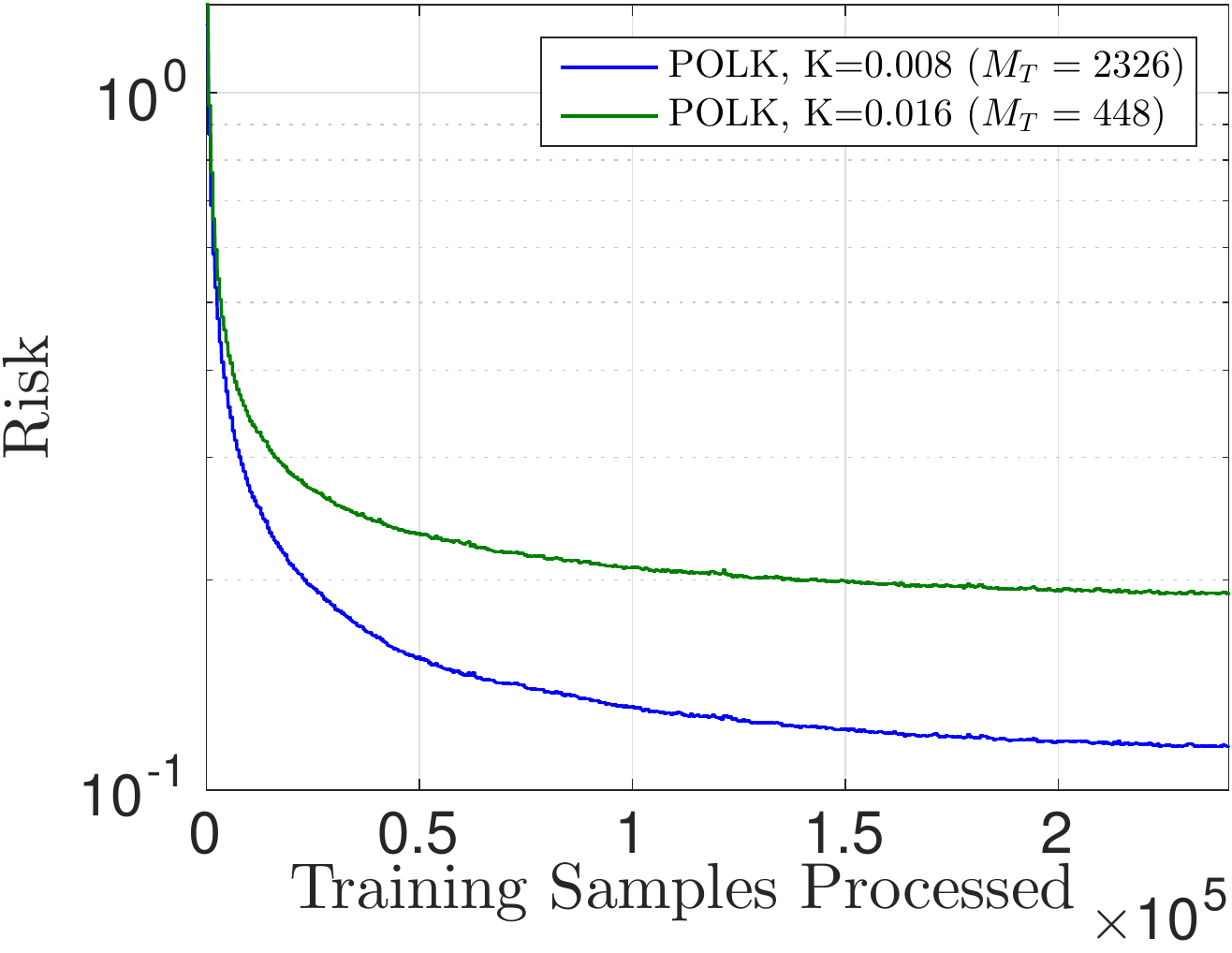}\label{subfig:objective_mnist_klr}}	
	\subfigure[Error rate]{\includegraphics[width=0.32\linewidth,height=3.5cm]{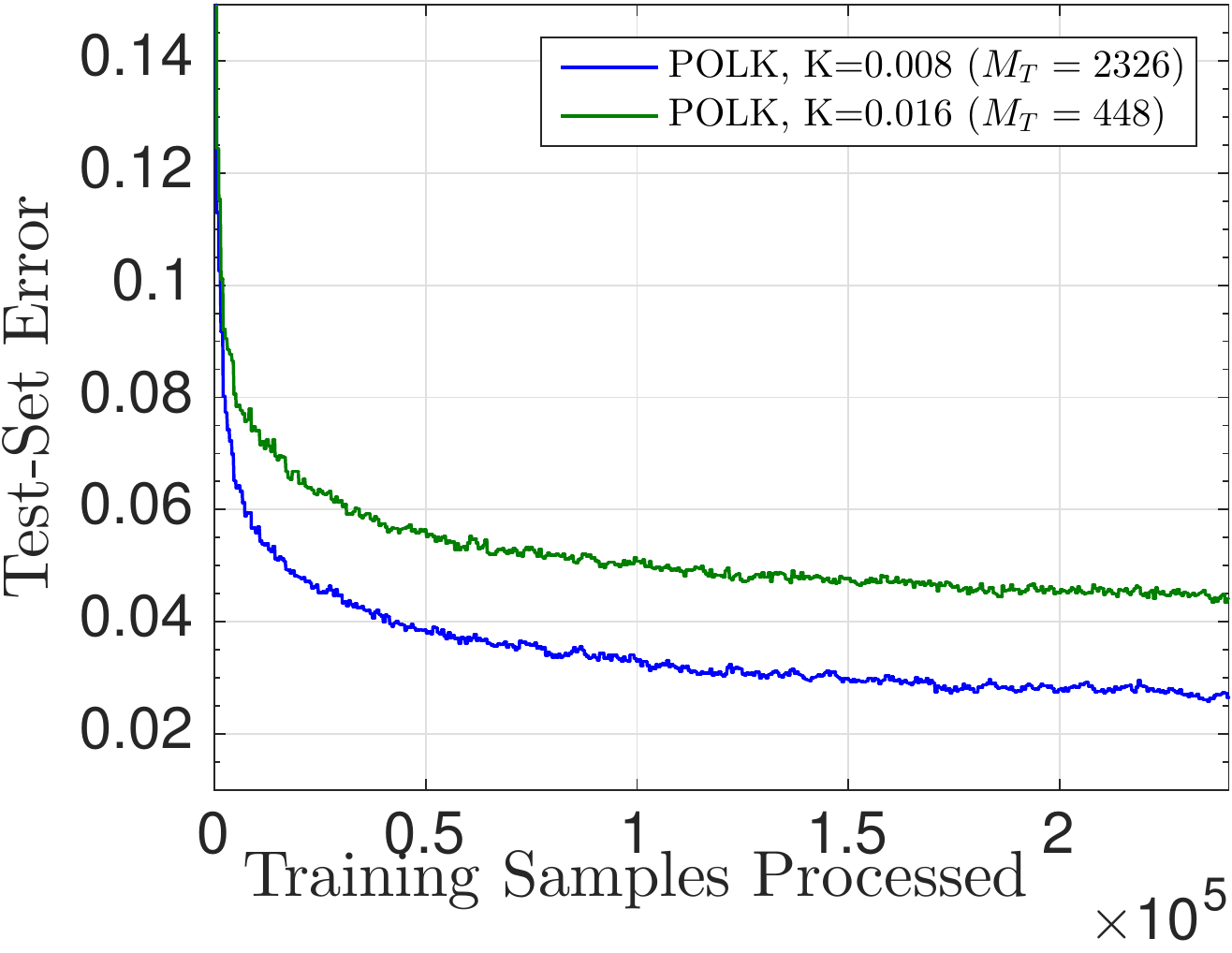}\label{subfig:error_mnist_klr}}
	\subfigure[Model order $M_t$]{\includegraphics[width=0.32\linewidth,height=3.5cm]{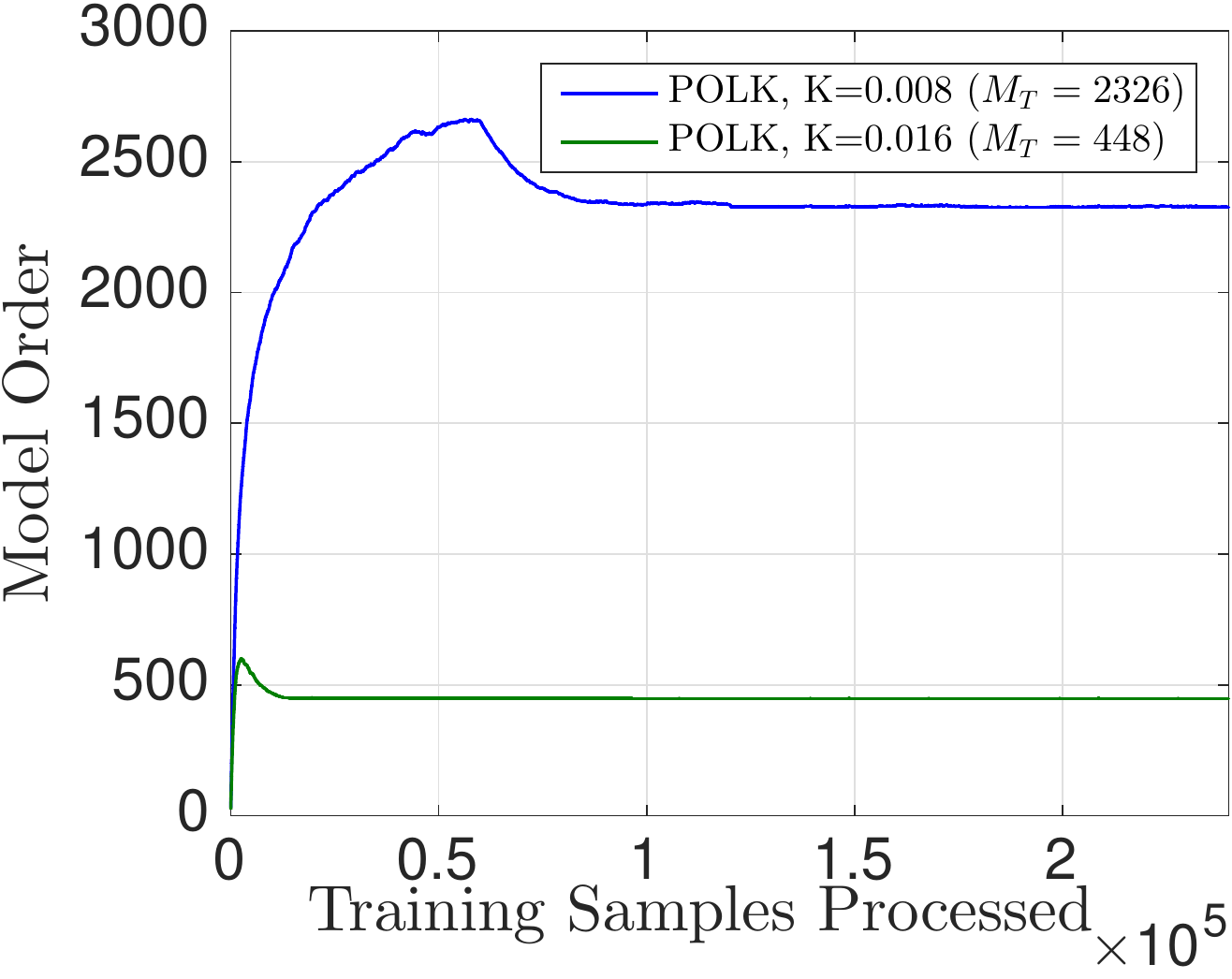}\label{subfig:order_mnist_klr}}
	\caption{Empirical behavior of the POLK algorithm applied to \texttt{mnist} data set for the Multi-Logistic task. The algorithm exhibits smoother convergence due to the differentiability of the logistic loss, and achieves asymptotic test error $2.6\%$. We again observe due to the dense data domain, larger model orders are needed to exhibit competitive classification performance.}\vspace{-4mm}
	\label{fig:klr_mnist}
\end{figure}

\begin{figure}
	\centering 
	\subfigure[Expected risk $R(\bbf_t)$]{\includegraphics[width=0.32\linewidth,height=3.5cm]{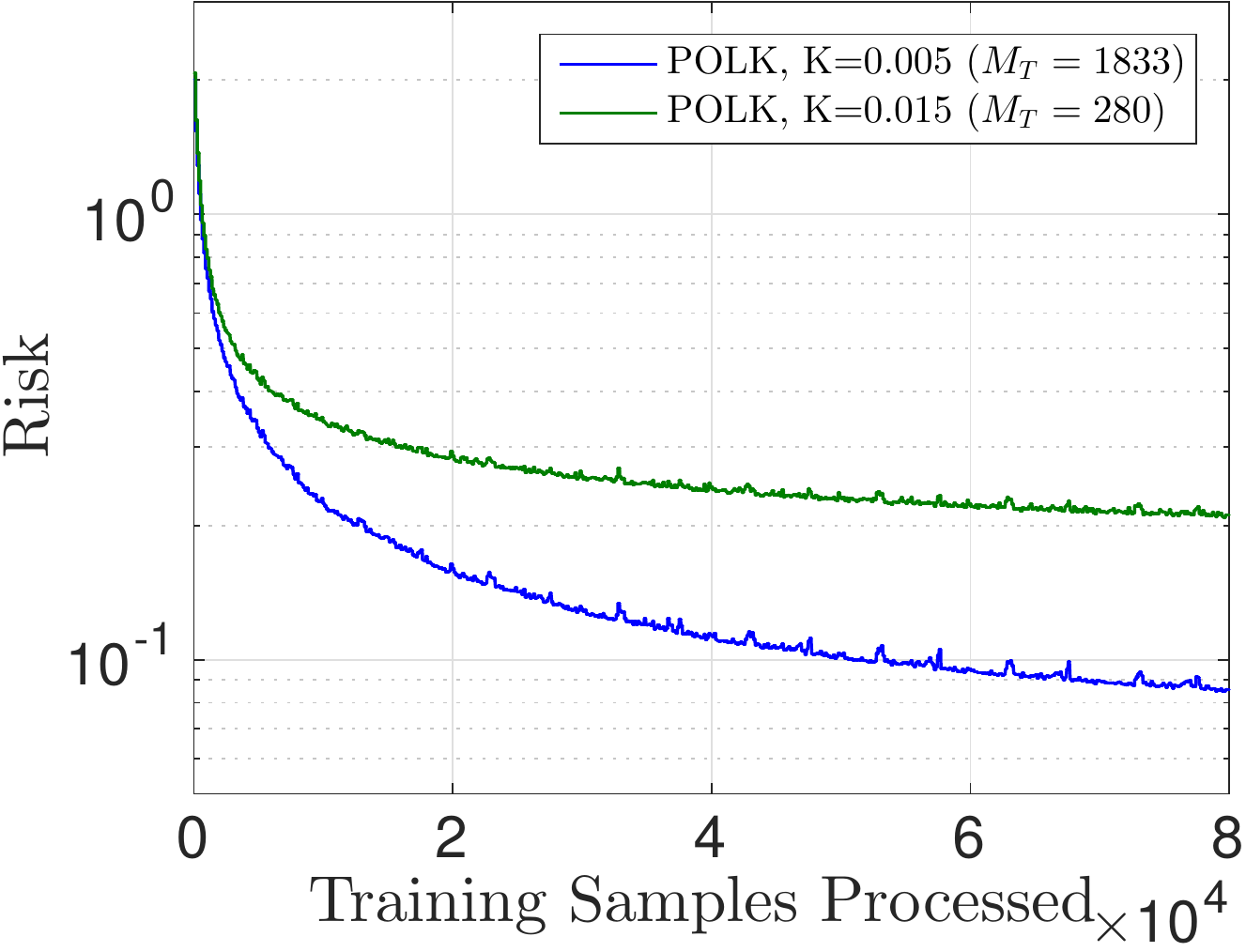}\label{subfig:objective_brodatz_klr}}	
	\subfigure[Error rate]{\includegraphics[width=0.32\linewidth,height=3.5cm]{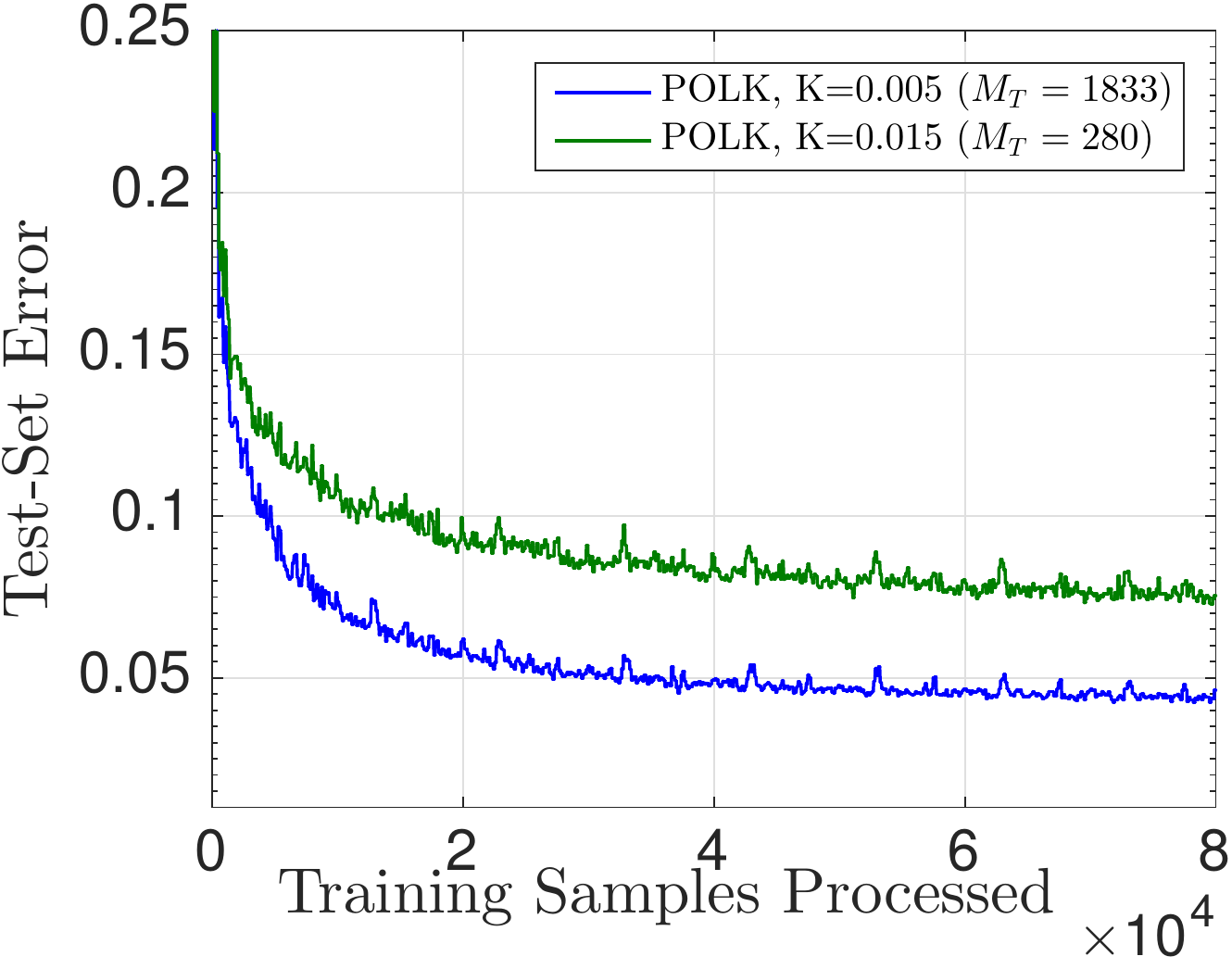}\label{subfig:error_brodatz_klr}}
	\subfigure[Model order $M_t$]{\includegraphics[width=0.32\linewidth,height=3.5cm]{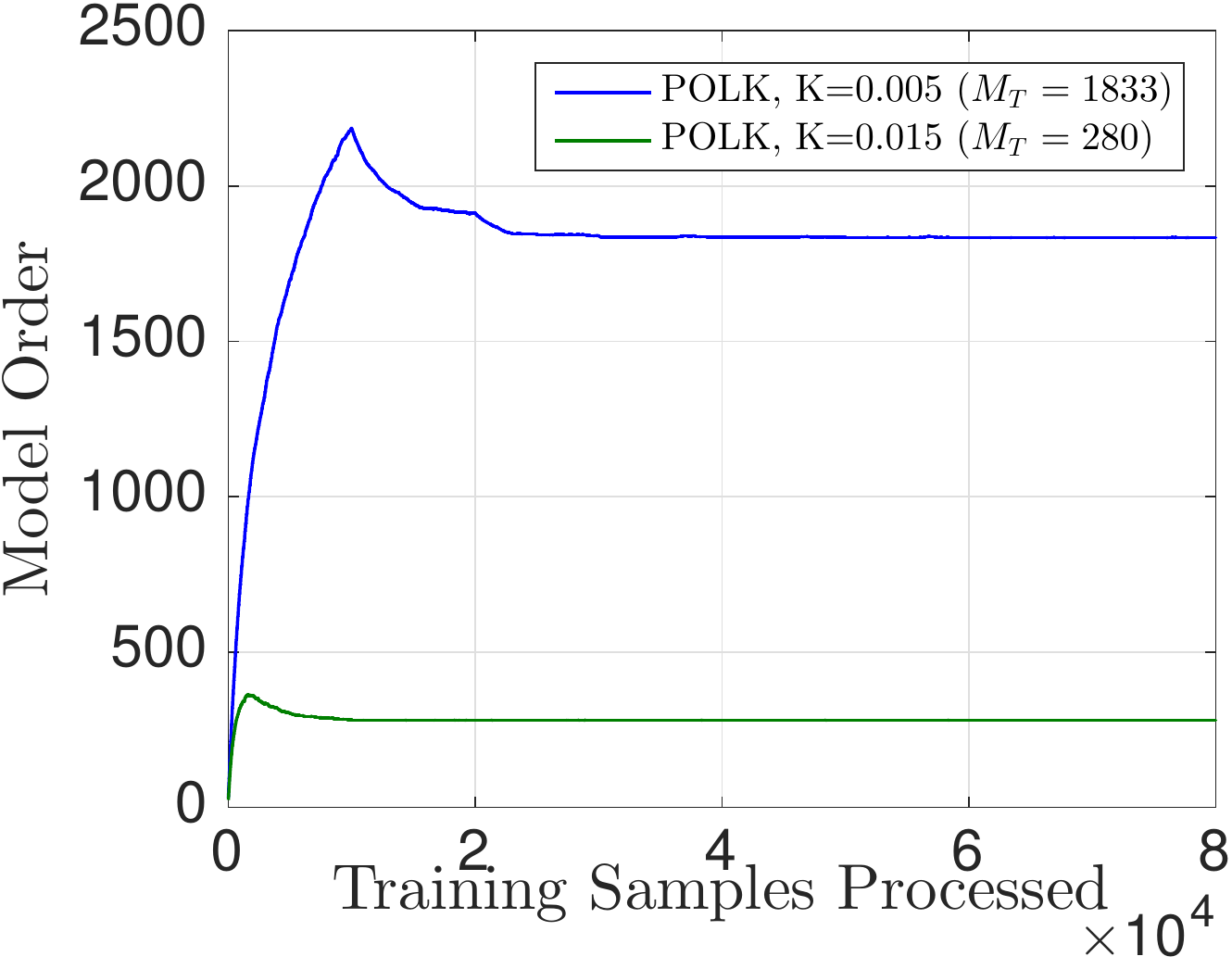}\label{subfig:order_brodatz_klr}}
	\caption{Empirical behavior of the POLK algorithm applied to the \texttt{brodatz} data set for the Multi-Logistic task. We observe convergent behavior, and a clear trade off between higher model order and increased accuracy. Due to this data domain being a more challenging task than the \texttt{mnist} digits, we observe asymptotic test accuracy of approximately $4.4\%$. }
	\label{fig:klr_brodatz}\vspace{-4mm}
\end{figure}

\vspace{0.2cm}

\noindent{\bf \texttt{mnist} and \texttt{brodatz} Results }\\
By construction, the \texttt{multidist} data set above yields optimal activation functions that are themselves sparse (i.e., $\bbf^*$ has a low model order due to the marginal feature density).  Here, we analyze the performance of POLK on more realistic data sets where the optimal solutions are not sparse, i.e., where one might desire a sparse approximation.  Due to the increased size and dimensionality of these data sets, we were unable to generate results for \texttt{mnist} using the batch L-BFGS technique, and unable to generate results for  either data set using IVM.

For Multi-KSVM on \texttt{mnist}, we used the following parameter values for POLK: Gaussian kernel with bandwidth $\tilde{\sigma}^2 = 4.0$, constant learning rate $\eta = 24.0$, parsimony constant $K \in \{0.16, 0.24\}$, and regularization constant $\lambda = 10^{-6}$.  We again processed data in mini-batches of size $32$.  For \texttt{brodatz}, we used identical parameters except for changing the kernel bandwith $\tilde{\sigma}^2 = 0.1$ and parsimony constant $K \in \{0.01, 0.02\}$.  For BSGD, we again found $\eta = 1.0$ to yield the best results on both datasets, and pre-specified model orders of $\{324,1086\}$ and $\{305,1171\}$ on \texttt{mnist} and \texttt{brodatz}, respectively, for comparison to POLK.

In Figures \ref{fig:ksvm_mnist} and \ref{fig:ksvm_brodatz} we plot the empirical results of these experiments for POLK and BSGD.  We observe that POLK is able to outperform the comparable BSGD trial in terms of convergence speed and steady-state risk and test-set error.  The strength of the proposed technique is further demonstrated in Table \ref{tab:mnistbrodatz}, where we can see that POLK is able to achieve test-set error within $1$-$2\%$ of LIBSVM while requiring a number of support vectors (model points) that is significantly-less than LIBSVM, \emph{while} adding the ability to process streaming data.

\begin{table*}[!]\centering
\begin{tabular}{ | c | c | c | c | c |}
\hline
                   & \multicolumn{2}{|c|}{\texttt{mnist}}                              & \multicolumn{2}{|c|}{\texttt{brodatz}} \\
\hline
\textbf{Algorithm}                   & Multi-KSVM                      & Multi-Logistic                  & Multi-KSVM                      & Multi-Logistic \\
                   & \tiny{(risk/error/model order)} & \tiny{(risk/error/model order)} & \tiny{(risk/error/model order)} & \tiny{(risk/error/model order)} \\
\hline
LIBSVM 				& $-/1.50/16118$                 & $-/-/-$                         & $-/3.72/4777$                   & $-/-/-$ \\ 
\hhline{|=|====|}
L-BFGS              & $-/-/-$                        & $-/-/-$                         & $0.0319/4.44/10000$             & $0.0572/4.00/10000$ \\
\hhline{|=|====|}
BSGD                & $0.0731/2.67/1086$             & $-/-/-$                         & $0.0560/4.72/1171$              & $-/-/-$ \\
\hline
POLK                & $0.0684/2.46/1086$             & $0.116/2.68/2326$               & $0.0507/4.53/1171$              & $0.0871/4.41/1833$ \\
\hline
\end{tabular}
\caption{Comparison of POLK, BSGD, IVM, L-BFGS, and LIBSVM results on the \texttt{mnist} and \texttt{brodatz} data sets.  Reported risk and error values for POLK and BSGD were averaged over the final 5\% of processed training examples.  Dashes indicate where the method could not be used to generate results either because it is not defined for the task or because the size of the problem was too large for that data set.  For these reasons, IVM was not able to generate results for these data sets on either task, and so is omitted here.  LIBSVM is used as a baseline, but note that it uses a fundamentally different model for multi-class problems (1v1 + majority vote), and so a comparable risk value can not be computed.}
\label{tab:mnistbrodatz}\vspace{-.5cm}
\end{table*}  
For the Multi-Logistic task on \texttt{mnist}, we ran POLK using a Gaussian kernel with bandwidth $\tilde{\sigma}^2 = 4.0$, constant learning rate $\eta = 24.0$, parsimony constant $K \in \{0.08, 0.16\}$, and regularization constant $\lambda = 10^{-6}$.  Data was processed in mini-batches of size $32$ here as well.  For \texttt{brodatz}, we again change the kernel bandwidth $\tilde{\sigma}^2 = 0.1$ and used different parsimony constants $K = \{0.005, 0.015\}$.  
The empirical behavior of POLK on this task can be seen in Figures \ref{fig:klr_mnist} and \ref{fig:klr_brodatz}. Observe that for this task the descent is smoother due to the differentiability of the logistic loss, although the asymptotic test accuracy is lower than that of KSVM.  

The overall performance is summarized in Table \ref{tab:mnistbrodatz}.  Note that the only other technique that was able to generate results for this task was L-BFGS, and even there only on the \texttt{brodatz} data set, since the complexity bottleneck in the sample size for \texttt{mnist} is prohibitive for batch optimization.  We see from this comparison that POLK yields a test-set error within $0.5\%$ of the batch solution while using an order of magnitude fewer model points.  Additionally, POLK is able to run \emph{online}, with streaming data, whereas L-BFGS requires all the data points to be operated on at each step.


\section{Discussion}\label{sec:discussion}
Over the past several years, parametric function approximation has largely dominated the machine learning landscape.
Deep learning is perhaps currently the most prominent parametric paradigm \citep{haykin1994neural}. One must first specify a network structure, thereby fixing the parametric representation of the function to be learned, before proceeding to determine the coefficients linking neurons in different layers.
Given this parametric representation, training techniques proceed by searching over the predefined parameter space for the optimal parameter values that minimize the error between the function and observed input-output pairs.
The main reason for the popularity of parametric function approximation is its success in solving practical problems, but there are other factors that have fostered their adoption.
One such factor is the availability of workable, if not necessarily efficient, optimization techniques for the determination of the optimal parameter values, in the form of stochastic gradient descent and its variants. Parametric stochastic gradient descent processes training examples sequentially and has a per-iteration complexity that is linear on the number of parameters but independent of the size of the training set.

Despite the success of parametric techniques, nonparametric function approximation has the advantage of expressive power in the sense that they are allowed to select the approximating function from a more general set of functions than those that admit a parametric form chosen \emph{a priori}.
This advantage is seen, e.g., in the improved classification accuracy of (nonparametric) kernel support vector machines (SVMs) relative to (parametric) linear SVMs \citep{evgeniou2000regularization}.
This is not to say that nonparametric methods are necessarily better. Neural networks, e.g., have proven to be very adept parametric representations in image classification problems \citep{krizhevsky2012imagenet}.
However, it is nonetheless true that the better expressive power of nonparametric representations is of importance in some applications.

The importance of expressiveness notwithstanding, nonparametric approaches are relatively less popular.
This is partly explained by the fact that, contrary to parametric approaches, workable algorithms for the minimization of functional costs are not as well-developed.
Indeed, nonparametric models involve function representations that depend on an infinite number of parameters.
This is a challenge not only because optimal function descriptions can become computationally intractable but, more importantly, because finding these optimal representations is itself intractable.
%
%

This work represents the first attempt at comprehensively addressing this intractability.
In particular, we have proposed solving general convex expected risk minimization problems over a Hilbert space that defines nonparametric regression functions in a way that guarantees the model order of the learned function does not grow unnecessarily large.
 In doing so, we addressed challenges (\ref{list1}) - (\ref{list2}) as follows: we considered the generalization of stochastic gradient descent to the kernelized expected risk setting and we compressed the learned decision function in a way that guarantees stochastic descent by tuning a greedy sparse approximation error criterion to the underlying optimization sequence.
The result is an almost-sure convergent function sequence with moderate complexity that is able to operate in true online settings.

Indeed, our experiments have shown that POLK performs comparably to batch kernel methods in terms of accuracy, while its model complexity is reduced by orders of magnitude.
Additionally, we observe state-of-the-art performance in terms of test-set accuracy relative to the number of samples processed.
Such performance is key to achieving reasonable performance in many applications of interest, e.g., when learning on robotics platforms operating in unknown environments.
In this case, the online nature of the problem is intrinsic and due to a lack of prior information on their operating domain \citep{cKoppelEtal16a}.

On the other hand, it must be noted that POLK, and even batch kernel methods, for certain large-scale supervised learning tasks, have not met the high bar of asymptotic test set accuracy set forth by batch approaches to deep learning \citep{krizhevsky2012imagenet}.
We believe this discrepancy is on account of the single-layer nature of the nonparametric regressor, which is tied to the choice of reproducing kernel used in our experiments.
Of course, more complicated multi-layer composite kernels may be used, based on the fact that a composition and positive linear combination of kernels is still a kernel \citep[Ch. 11]{theodoridis2015machine}.
However, the scalable development of online nonparametric methods based on such composite kernels is not straight-forward, and left to future work. 

\appendix

\section{Technical Assumptions and Auxiliary Results}\label{apx_assumptions}
Before analyzing the proposed method developed in Section \ref{sec:algorithm}, we define key quantities to simplify the analysis and introduce standard assumptions which are necessary to establish convergence. First, define the regularized stochastic functional gradient as
\begin{equation}\label{eq:stoch_reg_grad}
\hat{\nabla}_f\ell(f_{t}(\bbx_t),y_t) =\nabla_f\ell(f_{t}(\bbx_t),y_t)  + \lambda f_t
\end{equation}
Further define the projected stochastic functional gradient associated with the update in \eqref{eq:projection_hat} as
\begin{equation}\label{eq:proj_grad}
\tilde{\nabla}_f\ell(f_{t}(\bbx_t),y_t) =\Big( f_{t} - \ccalP_{ \ccalH_{\bbD_{t+1}}} \Big[
 f_t 
- {\eta}_t \hat{\nabla}_f\ell(f_{t}(\bbx_t),y_t) \Big]\Big)/\eta_t
\end{equation}
such that the Hilbert space update of Algorithm \ref{alg:soldd} [cf. \eqref{eq:projection_hat}] may be expressed as a stochastic descent using projected functional gradients
\begin{equation}\label{eq:iterate_tilde}
f_{t+1} = f_t - \eta_t \tilde{\nabla}_f\ell(f_{t}(\bbx_t),y_t) \; .
\end{equation}

The definitions \eqref{eq:proj_grad} - \eqref{eq:stoch_reg_grad} will be used to analyze the convergence behavior of the algorithm. Before doing so, observe that the stochastic functional gradient in \eqref{eq:stoch_reg_grad}, based upon the fact that $(\bbx_t, y_t)$ are independent and identically distributed realizations of the random pair $(\bbx, y)$, is an unbiased estimator of the true functional gradient of the regularized expected risk $R(f)$ in \eqref{eq:kernel_stoch_opt}, i.e.
\begin{equation}\label{eq:unbiased}
\mathbb{E}[\hat{\nabla}_f\ell(f_{t}(\bbx_t),y_t) \given \ccalF_t ] =\nabla_f R(f_{t})
\end{equation}
for all $t$. Next, we formally state technical conditions on the loss functions, data domain, and stochastic approximation errors that are necessary to establish convergence.
%
\begin{assumption}\label{as:first}
The feature space $\ccalX\subset\reals^p$ and target domain $\ccalY\subset\reals$ are compact, and the reproducing kernel map may be bounded as
\begin{equation}\label{eq:bounded_kernel}
\sup_{\bbx\in\ccalX} \sqrt{\kappa(\bbx, \bbx )} = X < \infty
\end{equation}
\end{assumption}

\begin{assumption}\label{as:2}
The instantaneous loss $\ell: \ccalH \times \ccalX \times \ccalY \rightarrow \reals$ is uniformly $C$-Lipschitz continuous for all $z \in \reals $ for a fixed $y\in\ccalY$
\begin{equation}\label{eq:lipschitz}
| \ell(z, y) - \ell( z', y ) | \leq C |z - z'|
\end{equation}
\end{assumption}
%
%
\begin{assumption}\label{as:3}
The loss function $\ell(f(\bbx), y)$ is convex and differentiable with respect to its first (scalar) argument $f(\bbx)$ on $\reals$ for all $\bbx\in\ccalX$ and $y\in\ccalY$.  
\end{assumption}
%
%
\begin{assumption}\label{as:last}
Let $\ccalF_t$ denote the sigma algebra which measures the algorithm history for times $u\leq t$, i.e. $\ccalF_t=\{\bbx_u, y_u, u_u\}_{u=1}^t$. The projected functional gradient of the regularized instantaneous risk in \eqref{eq:stoch_reg_grad} has finite conditional second moments for each $t$, that is,
\begin{equation}\label{eq:stochastic_grad_var}
\mathbb{E} [ \| \tilde{\nabla}_f\ell(f_{t}(\bbx_t),y_t)\|^2_{\ccalH} \mid \ccalF_t ] \leq \sigma^2
\end{equation}
\end{assumption}
Assumption \ref{as:first} holds in most practical settings by the data domain itself, and justifies the bounding of the loss in Assumption \ref{as:2}. Taken together, these conditions permit bounding the optimal function $f^*$ in the Hilbert norm, and imply that the worst-case model order is guaranteed to be finite. Variants of Assumption \ref{as:2} appear in the analysis of stochastic descent methods in the kernelized setting \citep{Pontil05erroranalysis,1715525}. Assumption \ref{as:3} is satisfied for supervised learning problems such as logistic regression, support vector machines with the square-hinge-loss, the square loss, among others. Moreover, it is a standard condition in the analysis of descent methods (see \cite{Boyd2004}). Assumption \ref{as:last} is common in stochastic approximation literature, and ensures that the variance of the stochastic approximation error is finite. 

Next we establish an auxiliary result needed to prove Theorems \ref{theorem_diminishing} and \ref{theorem_constant} which bounds the magnitude of the iterates of Algorithm \ref{alg:soldd} in the Hilbert norm.
\begin{proposition}\label{prop_bounded}
Let Assumptions \ref{as:first}-\ref{as:last} hold and denote $\{f_t \}$ as the sequence generated by Algorithm \ref{alg:soldd} with $f_0 = 0$. Further denote $f^*$ as the optimum defined by \eqref{eq:kernel_stoch_opt}. Both quantities are bounded by the constant $K:={C X}/{\lambda}$ in Hilbert norm for all $t$ as
\begin{equation}\label{eq:prop_bounded}
\| f_t \|_{\ccalH} \leq \frac{C X}{\lambda} \; , \qquad \|f^* \|_{\ccalH} \leq \frac{C X}{\lambda}
\end{equation}
\end{proposition}
\begin{myproof} 
First, since we repeatedly use the Cauchy-Schwartz inequality together with the reproducing kernel property in the following analysis, we here note that for all $g\in\ccalH$,  $|g(\bbx_t) | \leq | \langle g, \kappa(\bbx_t, \cdot) \rangle_{\ccalH} | \leq X \| g \|_{\ccalH}$. Now, consider the magnitude of $f_1$ in the Hilbert norm, given $f_0 = 0$
\begin{align}\label{eq:iterate_opt_bound}
\| f_1 \|_{\ccalH} 
&= \eta_0 \|  \tilde{\nabla}_f\ell(0,y_0) \|_{\ccalH}
=  \Big\| \ccalP_{ \ccalH_{\bbD_{1}}} \Big[
 \nabla_f\ell(0,y_0) \Big] \Big\|_{\ccalH} \nonumber \\
 &\leq \eta_0 \|  \nabla_f\ell(0,y_0) \|_{\ccalH}
  \leq \eta_0 |\ell'(0, y_0)| \|\kappa(\bbx_0, \cdot)\|_{\ccalH} \nonumber \\
&  \leq \eta_0 C X < \frac{C X}{\lambda}
\end{align}
The first equality comes from substituting in $f_0=0$ and the second is obtained by using the definition of the projected stochastic functional gradient in \eqref{eq:proj_grad}. On the second line, the first inequality comes from the definition of optimality condition of the projection operator, and the third uses the derivation of the functional stochastic gradient in \eqref{eq:stochastic_grad} with the Cauchy-Schwartz inequality. Lastly, we make use of Assumptions \ref{as:first} and \ref{as:2} to bound the scalar derivative $\ell'$ using the Lipschitz constant, and the boundedness of the kernel map [cf. \eqref{eq:bounded_kernel}]. The final strict inequality in \eqref{eq:iterate_opt_bound} comes from applying the step-size condition $\eta_0 < 1/\lambda$.

Now we consider the induction step. Given the induction hypothesis $\|f_t \|_{\ccalH} \leq CX/\lambda$, consider the magnitude of the iterate at the time $t+1$ as
\begin{align}\label{eq:iterate_opt_bound_induction1}
\|f_{t+1} \|_{\ccalH}& = \Big\|\ccalP_{ \ccalH_{\bbD_{t+1}}} \Big[
(1-\eta_t \lambda) f_t 
- \eta_t \nabla_f\ell(f_{t}(\bbx_t),y_t) \Big] \Big\|_{\ccalH} \nonumber \\
&\leq \|(1-\eta_t \lambda) f_t 
- \eta_t \nabla_f\ell(f_{t}(\bbx_t),y_t) \|_{\ccalH} \nonumber \\
&\leq (1-\eta_t \lambda)\| f_t \|
+ \eta_t \| \nabla_f\ell(f_{t}(\bbx_t),y_t) \|_{\ccalH} \; ,
\end{align}
where we have applied the non-expansion property of the projection operator for the first inequality on the right-hand side of \eqref{eq:iterate_opt_bound_induction1}, and the triangle inequality for the second. Now, apply the induction hypothesis $\|f_t \|_{\ccalH} \leq CX/\lambda$ to the first term on the right-hand side of \eqref{eq:iterate_opt_bound_induction1}, and the chain rule together with the triangle inequality to the second to obtain
\begin{align}\label{eq:iterate_opt_bound_induction2}
\|f_{t+1} \|_{\ccalH}
&\leq (1-\eta_t \lambda) \frac{C X}{\lambda}
+ \eta_t | \ell'(f_{t}(\bbx_t),y_t) | \|\kappa(\bbx_t, \cdot)\|_\ccalH
\nonumber \\
&\leq (\frac{1}{\lambda} -\eta_t ) CX 
+ \eta_t C X = \frac{C X}{\lambda}
\end{align}
where we have made use of Assumptions \ref{as:first} and \ref{as:2} to bound the scalar derivative $\ell'$ using the Lipschitz constant, and the boundedness of the kernel map [cf. \eqref{eq:bounded_kernel}] as in the base case for $f_1$, as well as the fact that $\eta_t < 1/\lambda$. The same bound holds for $f^*$ by applying the result of Section V-B of \cite{Kivinen2004} with $m\rightarrow \infty$.

 \end{myproof}

Next we introduce a proposition which quantifies the error due to our sparse stochastic projection scheme in terms of the ratio of the sparse approximation budget to the algorithm step-size.
%
\begin{proposition}\label{prop1}
Given independent identical realizations $(\bbx_t, y_t)$ of the random pair $(\bbx, y)$, the difference between the projected stochastic functional gradient and the stochastic functional gradient of the regularized instantaneous risk defined by \eqref{eq:proj_grad} and \eqref{eq:stoch_reg_grad}, respectively, is bounded for all $t$ as
\begin{equation}\label{eq:prop1}
 \| \tilde{\nabla}_f\ell(f_{t}(\bbx_t),y_t) - \hat{\nabla}_f\ell(f_{t}(\bbx_t),y_t) \|_{\ccalH} \leq \frac{\eps_t}{\eta_t}
\end{equation}
where $\eta_t>0$ denotes the algorithm step-size and $\eps_t>0$ is the approximation budget parameter of Algorithm \ref{alg:komp}.
\end{proposition}
\begin{myproof}
Consider the square-Hilbert-norm difference of $ \tilde{\nabla}_f\ell(f_{t}(\bbx_t),y_t) $  and $\hat{\nabla}_f\ell(f_{t}(\bbx_t),y_t) $ defined in \eqref{eq:stoch_reg_grad} and \eqref{eq:proj_grad}, respectively,
\begin{align}\label{eq:norm_stoch_grad_diff}
 \| \tilde{\nabla}_f&\ell(f_{t}(\bbx_t),y_t) - \hat{\nabla}_f\ell(f_{t}(\bbx_t),y_t) \|_{\ccalH}^2  \\
& = \Big\| \Big( f_{t} -\ccalP_{ \ccalH_{\bbD_{t+1}}} \Big[
  f_t 
- \eta_t \hat{\nabla}_f\ell(f_{t}(\bbx_t),y_t) \Big]\Big)/\eta_t 
 - \hat{\nabla}_f\ell(f_{t}(\bbx_t),y_t) \Big\|_{\ccalH}^2 \nonumber
\end{align}
Multiply and divide $\hat{\nabla}_f\ell(f_{t}(\bbx_t),y_t)$, the last term, by $\eta_t$, and reorder terms to write
\begin{align}\label{eq:norm_stoch_grad_expand}
\Big\| \Big( f_{t} -\ccalP_{ \ccalH_{\bbD_{t+1}}} \Big[  f_t 
&- \eta_t \hat{\nabla}_f\ell(f_{t}(\bbx_t),y_t) \Big]\Big)/\eta_t 
 - \hat{\nabla}_f\ell(f_{t}(\bbx_t),y_t) \Big\|_{\ccalH}^2 \nonumber \\
& = \Big\| \frac{1}{\eta_t}\left( f_{t} -\eta_t \hat{\nabla}_f\ell(f_{t}(\bbx_t),y_t)\right) -\frac{1}{\eta_t}\ccalP_{ \ccalH_{\bbD_{t+1}}} \Big[  f_t - \eta_t \hat{\nabla}_f\ell(f_{t}(\bbx_t),y_t) \Big]  \Big\|_{\ccalH}^2 \nonumber \\
&=\frac{1}{\eta_t^2}\| \tilde{f}_{t+1} - f_{t+1} \|_{\ccalH}^2 
  \end{align}
  where we have substituted the definition of $\tilde{f}_{t+1}$ and $f_{t+1}$ in  \eqref{eq:sgd_tilde} and \eqref{eq:proximal_hilbert_dictionary}, respectively, and pulled the nonnegative scalar $\eta_t$ outside the norm. Now, observe that the KOMP residual stopping criterion in Algorithm \ref{alg:komp} is $\lVert \tilde{f}_{t+1} - f_{t+1} \rVert_{\ccalH} \leq \epsilon_t$, which we may apply to the last term on the right-hand side of \eqref{eq:norm_stoch_grad_expand} to conclude \eqref{eq:prop1}.

\end{myproof}
\begin{lemma}(Stochastic Descent) \label{lemma1}
Consider the sequence generated $\{f_t \}$ by Algorithm \ref{alg:soldd} with $f_0 = 0$. Under Assumptions \ref{as:first}-\ref{as:last}, the following expected descent relation holds.
\begin{align}\label{eq:lemma1}
\E{\| f_{t+1} - f^* \|_{\ccalH}^2 \given \ccalF_t}
	 &\leq\| f_t - f^* \|_{\ccalH}^2
	 - 2 \eta_t [R(f_t) -R(f^*)]  + 2 \eps_t \| f_t - f^* \|_{\ccalH} 
	+ \eta_t^2 \sigma^2 \; . 
\end{align}
\end{lemma}
\begin{myproof} 
Begin by considering the square of the Hilbert-norm difference between $f_{t+1}$ and $f^*$ defined by \eqref{eq:kernel_stoch_opt}, and expand the square to write
\begin{align}\label{eq:iterate_square_expand}
\| f_{t+1} - f^* \|_{\ccalH}^2 
	& =\| f_t - \eta_t \tilde{\nabla}_f\ell(f_{t}(\bbx_t),y_t) \|_{\ccalH}^2 \nonumber \\
	& =\|  f_t - f^* \|_{\ccalH}^2 - 2 \eta_t \langle f_t - f^*, \tilde{\nabla}_f\ell(f_{t}(\bbx_t),y_t) \rangle_{\ccalH} + \eta_t^2 \| \tilde{\nabla}_f\ell(f_{t}(\bbx_t),y_t) \|_{\ccalH}^2
\end{align}
Add and subtract the gradient of the regularized instantaneous risk $\hat{\nabla}_f\ell(f_{t}(\bbx_t),y_t)$ defined in \eqref{eq:stoch_reg_grad} to the second term on the right-hand side of \eqref{eq:iterate_square_expand} to obtain
\begin{align}\label{eq:iterate_stoch_grad}
\| f_{t+1} - f^* \|_{\ccalH}^2 
	 &=\| f_t - f^* \|_{\ccalH}^2
	 - 2 \eta_t \langle f_t - f^*, \hat{\nabla}_f\ell(f_{t}(\bbx_t),y_t) \rangle_{\ccalH}  \\
	 &\quad- 2 \eta_t \langle f_t - f^*, \tilde{\nabla}_f\ell(f_{t}(\bbx_t),y_t) - \hat{\nabla}_f\ell(f_{t}(\bbx_t),y_t) \rangle_{\ccalH} 
	+ \eta_t^2 \| \tilde{\nabla}_f\ell(f_{t}(\bbx_t),y_t) \|_{\ccalH}^2 \nonumber
\end{align}
We deal with the third term on the right-hand side of \eqref{eq:iterate_stoch_grad}, which represents the directional error associated with the sparse stochastic projections, by applying the Cauchy-Schwartz inequality together with Proposition \ref{prop1} to obtain
\begin{align}\label{eq:iterate_prop1_subst}
\| f_{t+1} - f^* \|_{\ccalH}^2 
	 &=\| f_t - f^* \|_{\ccalH}^2
	 - 2 \eta_t \langle f_t - f^*, \hat{\nabla}_f\ell(f_{t}(\bbx_t),y_t) \rangle_{\ccalH}  \\
	 &\quad+ 2 \eps_t \| f_t - f^* \|_{\ccalH} 
	+ \eta_t^2 \| \tilde{\nabla}_f\ell(f_{t}(\bbx_t),y_t) \|_{\ccalH}^2 \nonumber
\end{align}
Now compute the expectation of \eqref{eq:iterate_prop1_subst} conditional on the algorithm history $\ccalF_t$
\begin{align}\label{eq:expectation_square_expand}
\E{\| f_{t+1} - f^* \|_{\ccalH}^2 \given \ccalF_t}
	 =\| f_t - f^* \|_{\ccalH}^2
	 - 2 \eta_t \langle f_t - f^*, {\nabla}_f R(f_{t}) \rangle_{\ccalH}  + 2 \eps_t \| f_t - f^* \|_{\ccalH} 
	+ \eta_t^2 \sigma^2 \; , 
\end{align}
where we have applied the fact that the stochastic functional gradient in \eqref{eq:stoch_reg_grad} is an unbiased estimator [cf. \eqref{eq:unbiased}] for the functional gradient of the expected risk in \eqref{eq:kernel_stoch_opt}, as well as the fact that the variance of the functional projected stochastic gradient is finite stated in \eqref{eq:stochastic_grad_var} (Assumption \ref{as:last}). 
Observe that since $R(f)$ is an expectation of a convex function, it is also convex, which allows us to write
\begin{equation}\label{eq:convexity}
R(f_t) -R(f^*) \leq \langle f_t - f^*, {\nabla}_f R(f_{t}) \rangle_{\ccalH}  \;,
\end{equation}
which we substitute into the second term on the right-hand side of the relation given in \eqref{eq:expectation_square_expand} to obtain
\begin{align}\label{eq:expectation_convexity}
\E{\| f_{t+1} - f^* \|_{\ccalH}^2 \given \ccalF_t}
	 &\leq\| f_t - f^* \|_{\ccalH}^2
	 - 2 \eta_t [R(f_t) -R(f^*)]  + 2 \eps_t \| f_t - f^* \|_{\ccalH} 
	+ \eta_t^2 \sigma^2 \; . 
\end{align}
Thus the claim in Lemma \ref{lemma1} is valid.
%
 \end{myproof}

\section{Proof of Theorem \ref{theorem_diminishing}}\label{apx_theorem_diminishing}
Apply the iterate bound stated in Proposition \ref{prop_bounded} to the third term on the right-hand side of \eqref{eq:lemma1} (Lemma \ref{lemma1}) to write
\begin{equation}\label{eq:expectation_convexity2}
\E{\| f_{t+1} - f^* \|_{\ccalH}^2 \given \ccalF_t}
	 \leq\| f_t - f^* \|_{\ccalH}^2
	 - 2 \eta_t [R(f_t) -R(f^*)]  +  \eta_t^2 \left(\frac{4  C X}{\lambda}
	+  \sigma^2\right) \; . 
\end{equation}
where we also have applied the approximation budget condition $\eps_t = \eta_t^2$. We use the relation in \eqref{eq:expectation_convexity2} to construct a martingale difference sequence. In particular, define the nonnegative stochastic processes $\alpha_t$ and $\beta_t$ as
\begin{align}\label{eq:alpha_beta}
\alpha_t = \| f_t - f^* \|_{\ccalH}^2 + \left(\frac{4  C X}{\lambda}
	+  \sigma^2\right)\sum_{u=t}^\infty \eta_u^2 
	 \; ,\qquad \beta_t = 	2 \eta_t [R(f_t) -R(f^*)] 
\end{align}
Observe that $\alpha_t$ is finite almost surely, since $\sum_{u=t}^\infty \eta_u^2 \leq \sum_{u=0}^\infty \eta_u^2$. Given the definitions of $\alpha_t$ and $\beta_t$ in \eqref{eq:alpha_beta}, we may write
\begin{equation}\label{eq:martingale_diff1}
\E{\alpha_{t+1} \given \ccalF_t } \leq \alpha_t - \beta_t \; ,
\end{equation}
together with the fact that $\alpha_t$ and $\beta_t$ are nonnegative, whereby the conditions of the Supermartingale Convergence Theorem \citep{victor1995adaptive} are satisfied. Therefore, we obtain that (i) $\alpha_t$ has a finite limit almost surely; and (ii) the series $\sum_{t=1}^\infty \beta_t < \infty$ is almost surely finite. The later result, taken together with the non-summability of the step-sizes stated in \eqref{eq:dim_stepsize}, implies that the almost surely the limit infimum of $R(f_t) - R(f^*)$ is null, i.e.
\begin{equation}\label{eq:liminf_objective_convergence}
\liminf_{t\rightarrow \infty} R(f_t) - R(f^*) = 0 \qquad\text{ a.s.}
\end{equation}
Now, using the consequence of the Supermartingale Convergence Theorem, $\alpha_t$ almost surely has a limit. Observe that the sum $\sum_{u=t}^\infty$ is a deterministic quantity whose limit is null (we sum over less and less terms over time, asymptotically summing over zero terms). Taken with the finiteness of the limit of $\alpha_t$, we conclude
\begin{equation}\label{eq:sequence_conv}
\lim_{t\rightarrow \infty} \| f_t - f^*\|_{\ccalH}^2 = 0 \qquad \text{ a.s.}
\end{equation}
%
\section{Proof of Theorem \ref{theorem_constant}}\label{apx_theorem_constant}
\begin{myproof}
The use of the regularizing term $(\lambda/2)\|f \|^2_\ccalH$ in \eqref{eq:kernel_stoch_opt} implies that the regularized expected risk is $\lambda$-strongly convex with respect to $f\in\ccalH$, which allows us to write
\begin{equation}\label{eq:strong_cvx}
 \frac{\lambda}{2} \|f_t - f^* \|_{\ccalH}^2 \leq R(f_t) - R(f^*)
\end{equation}
Substituting the relation \eqref{eq:strong_cvx} into the second term on the right-hand side of the expected descent relation stated in Lemma \ref{lemma1}, with constant step-size $\eta_t=\eta$ and approximation budget $\eps_t=\eps$, yields
\begin{align}\label{eq:expected_descent}
\E{\| f_{t+1} - f^* \|_{\ccalH}^2 \given \ccalF_t}
	 &\leq (1 - \eta \lambda )\| f_t - f^* \|_{\ccalH}^2
	 + 2 \eps \| f_t - f^* \|_{\ccalH} 
	+ \eta^2 \sigma^2 \; . 
\end{align}
We use the expression in \eqref{eq:expected_descent} to construct a stopping stochastic process, which tracks the suboptimality of $\| f_t - f^* \|_{\ccalH}^2$ until it reaches a specific threshold. In doing so, we obtain convergence to a neighborhood. We aim to define a stochastic process $\delta_t$ that qualifies as a supermartingale, i.e. $\E{ \delta_{t+1} \given \ccalF_t } \leq \delta_t$. To do so, consider \eqref{eq:expected_descent} and solve for the appropriate threshold by analyzing when the following holds true
\begin{align}\label{eq:martingale_construct1}
\E{\| f_{t+1} - f^* \|_{\ccalH}^2 \given \ccalF_t}
	 &\leq (1 - \eta \lambda )\| f_t - f^* \|_{\ccalH}^2
	 + 2 \eps \| f_t - f^* \|_{\ccalH} 
	+ \eta^2 \sigma^2 \leq \| f_t - f^* \|^2_{\ccalH} \; . 
\end{align}
Re-arrange the above expression to obtain the sufficient condition
\begin{align}\label{eq:martingale_construct2}
  - \eta \lambda \| f_t - f^* \|_{\ccalH}^2
	 + 2 \eps \| f_t - f^* \|_{\ccalH} 
	+ \eta^2 \sigma^2 \leq 0\; . 
\end{align}
Observe that \eqref{eq:martingale_construct2} defines a quadratic polynomial in $\| f_t - f^* \|_{\ccalH}$, which, using the quadratic formula, has roots
\begin{align}\label{eq:quadratic_formula}
\| f_t - f^* \|_{\ccalH}= \frac{2\eps \pm \sqrt{4\eps^2 - ( - 4 \lambda \eta)(\eta^2 \sigma^2)}}{-2 \lambda \eta} = \frac{\eps \pm \sqrt{\eps^2 + \lambda \eta^3  \sigma^2 }}{\lambda \eta} 
\end{align}
The quadratic polynomnial defined by \eqref{eq:martingale_construct2} opens downward, and $\|f_t - f^* \|_{\ccalH} \geq 0$, so we focus on the positive root, substituting the approximation budget selection $\eps=K \eta^{3/2}$ to define the radius of convergence as 
\begin{align}\label{eq:radius}
\Delta:= \frac{\eps + \sqrt{\eps^2 + \lambda \eta^3  \sigma^2 }}{\lambda \eta} = \frac{\sqrt{\eta}}{\lambda}\Big(K +  \sqrt{K^2 + \lambda  \sigma^2 }\Big)
\end{align}
The definition \eqref{eq:radius} allows us to construct a stopping process. In particular, define the stochastic process $\delta_t$ as
\begin{align}\label{eq:delta}
\delta_t = \|f_t - f^* \|_\ccalH \mathbbm{1} \Big\{\min_{u\leq t}   - \eta \lambda \| f_u - f^* \|_{\ccalH}^2
	 + 2 \eps \| f_u - f^* \|_{\ccalH} 
	+ \eta^2 \sigma^2    > \Delta \Big\} 
\end{align}
where $\mathbbm{1}\{E \}$ denotes the indicator process of event $E \in \ccalF_t$. Note that $\delta_t\geq 0$ for all $t$, since both $\| f_t - f^* \|_{\ccalH}$ and the indicator function are nonnegative. Observe that, given the definition \eqref{eq:delta}, either 
$\min_{u\leq t}   - \eta \lambda \| f_u - f^* \|_{\ccalH}^2
	 + 2 \eps \| f_u - f^* \|_{\ccalH} 
	+ \eta^2 \sigma^2    > \Delta$ holds, in which case we may compute the square root of the condition  in \eqref{eq:martingale_construct1} to write
\begin{align}\label{eq:delta_martingale}
\mathbb{E}[\delta_{t+1} \given \ccalF_t ] \leq \delta_t
\end{align}
Alternatively, $\min_{u\leq t}   - \eta \lambda \| f_u - f^* \|_{\ccalH}^2
	 + 2 \eps \| f_u - f^* \|_{\ccalH} 
	+ \eta^2 \sigma^2    \leq \Delta$, in which case the indicator function is null for all subsequent times, due to the use of the minimum inside the indicator in the definition of \eqref{eq:delta}. Thus in either case, \eqref{eq:delta_martingale} holds, which implies that $\delta_t$ converges almost surely to null, which, as a consequence we obtain
	%
the fact that either %
$\lim_{t\rightarrow \infty }\|f_t - f^* \|_\ccalH - \Delta = 0$ or
the indicator function is null for large $t$, i.e. 
$\lim_{t\rightarrow \infty} \mathbbm{1} \{\min_{u\leq t}   - \eta \lambda \| f_u - f^* \|_{\ccalH}^2
	 + 2 \eps \| f_u - f^* \|_{\ccalH} 
	+ \eta^2 \sigma^2    > \Delta \} = 0$ almost surely.
Therefore, we obtain that 
\begin{align}\label{eq:liminf_constant}
\liminf_{t\rightarrow \infty} \|f_t - f^* \|_\ccalH \leq \Delta = \frac{\sqrt{\eta}}{\lambda}\Big(K +  \sqrt{K^2 + \lambda  \sigma^2 }\Big)\qquad\text{ a.s. }
\end{align}
which is as stated in Theorem \ref{theorem_constant}.
\end{myproof}
%
\section{Proofs Leading to Theorem \ref{theorem_model_order}}\label{apx_theorem_model_order1}
Before proving Theorem \ref{apx_theorem_model_order1}, we present a lemma which allows us to relate the stopping criterion of our sparsification procedure to a Hilbert subspace distance.
%
\begin{lemma}\label{lemma_subspace_dist}
Define the distance of an arbitrary feature vector $\bbx$ evaluated by the feature transformation $\phi(\bbx) = \kappa(\bbx, \cdot)$ to $\ccalH_{\bbD}=\text{span}\{\kappa(\bbd_n, \cdot) \}_{n=1}^M$, the subspace of the Hilbert space spanned by a dictionary $\bbD$ of size $M$,  as
\begin{equation}\label{eq:hilbert_subspace_dist}
\text{dist}( \kappa(\bbx, \cdot) , \ccalH_{\bbD}) 
= \min_{f\in\ccalH_{\bbD}} \| \kappa(\bbx, \cdot) - \bbv^T \bbkappa_{\bbD}(\cdot) \|_{\ccalH} \; .
\end{equation}
This set distance simplifies to following least-squares projection when $\bbD \in \reals^{p\times M}$ is fixed
\begin{equation}\label{eq:hilbert_subspace_dist_ls}
\text{dist}( \kappa(\bbx, \cdot) , \ccalH_{\bbD}) 
= \Big\|  \kappa(\bbx,\cdot) 
  - [\bbK_{\bbD, \bbD}^{-1} \bbkappa_{\bbD}(\bbx)]^T
   \bbkappa_{\bbD}(\cdot) \Big\|_{\ccalH} \; .
\end{equation}
\end{lemma}



\begin{myproof}
The distance to the subspace $\ccalH_{\bbD}$ is defined as
\begin{align}\label{eq:subspace_dist}
\text{dist}( \kappa(\bbx, \cdot) , \ccalH_{\bbD_t}) 
= \min_{f\in\ccalH_{\bbD}} \| \kappa(\bbx, \cdot) - \bbv^T \bbkappa_{\bbD}(\cdot) \|_{\ccalH} 
= \min_{\bbv\in \reals^{M}} \| \kappa(\bbx, \cdot) - \bbv^T \bbkappa_{\bbD}(\cdot) \|_{\ccalH} \; ,
\end{align}
where the first equality comes from the fact that the dictionary $\bbD$ is fixed, so $\bbv\in \reals^M$ is the only free parameter. Now plug in the minimizing weight vector $\tbv^*=\bbK_{\bbD_t, \bbD_t}^{-1}\bbkappa_{\bbD_t}(\bbx_t)$ into \eqref{eq:subspace_dist} which is obtained in an analogous manner to the logic which yields \eqref{eq:proximal_hilbert_representer} - \eqref{eq:hatparam_update}. Doing so simplifies \eqref{eq:subspace_dist} to the following
\begin{align}\label{eq:subspace_dist2}
\text{dist}(\kappa(\bbx_t, \cdot) , \ccalH_{\bbD_t}) =  \Big\|  \kappa(\bbx_t,\cdot) 
  - [\bbK_{\bbD_t, \bbD_t}^{-1} \bbkappa_{\bbD_t}(\bbx_t)]^T
   \bbkappa_{\bbD_t}(\cdot) \Big\|_{\ccalH} \; .
\end{align}
\end{myproof}
\subsection{Proof of Theorem \ref{theorem_model_order}}\label{apx_theorem_model_order}
\begin{myproof}
The proof proceeds by the following logic. We begin by considering the model order at two arbitrary subsequent iterates of Algorithm \ref{alg:soldd}, and reduce model order growth at a given time to a criterion involving the approximation error $\gamma_{M_t+1}$ associated with removing the most recent feature vector $\bbx_t$, and then analyze the conditions under which this simplified criterion is not satisfied for all subsequent times, meaning that the model order does not grow beyond a certain point. To do so, we prove that this quantity reduces to a weighted set distance to the Hilbert subspace $\ccalH_{\bbD_t}$ defined by dictionary $\bbD_t$, and thus we are able to invoke point-set topological properties of the compact feature space $\ccalX$, specifically, its packing number, which guarantee that the number of dictionary elements remains finite, in a manner similar to the proof of Theorem 3.1 in \cite{1315946}.

Consider the model order of the function iterates $f_t$ and $f_{t+1}$ generated by Algorithm \ref{alg:soldd} denoted by $M_t$ and $M_{t+1}$, respectively, at two arbitrary subsequent times $t$ and $t+1$. Assume a constant algorithm step-size $\eta$ has been chosen such that $\eta<1/\lambda$ and the approximation budget $\eps$ satisfies $\eps=K \eta^{3/2}$ for some positive scalar $K>0$.
Suppose the model order of the function $f_{t+1}$ is less than or equal to that of $f_t$, i.e. $M_{t+1} \leq M_t$. This relation holds when the stopping criterion of KOMP (Algorithm \ref{alg:komp}), stated as $\min_{j=1,\dots,{M_t + 1}} \gamma_j > \eps$, \emph{is not} satisfied for the kernel dictionary matrix with the newest sample point $\bbx_t$ appended: $\tbD_{t+1} = [\bbD_t ; \bbx_t ]$ [cf. \eqref{eq:param_tilde}], which is of size $M_t + 1$. 
Thus, the negation of the termination condition of Algorithm \ref{alg:komp} must hold for this case, stated as
\begin{equation}\label{eq:komp_no_terminate}
\min_{j=1,\dots,{M_t + 1}} \gamma_j \leq \eps \; .
\end{equation}
Observe that the left-hand side of \eqref{eq:komp_no_terminate} lower bounds the approximation error $\gamma_{M_t + 1}$ of removing the most recent feature vector $\bbx_t$ due to the minimization over $j$, that is, $\min_{j=1,\dots,{M_t + 1}} \gamma_j \leq\gamma_{M_t + 1} $. Consequently, if $\gamma_{M_t + 1} \leq \eps$, then \eqref{eq:komp_no_terminate} holds and the model order does not grow. Thus it suffices to consider $\gamma_{M_t + 1}$.

The definition of $\gamma_{M_t + 1}$ with the substitution of $\tilde{f}_{t+1}$ in \eqref{eq:sgd_tilde} allows us to write
\begin{align}\label{eq:min_gamma_expand}
\gamma_{M_t+1}
			&=\min_{\bbu\in\reals^{{M_t}}} \Big\|(1-\eta\lambda){f}_t - \eta\ell'({f}_t(\bbx_t),\bby_t)\kappa(\bbx_t,\cdot) -\sum_{k \in \ccalI \setminus \{M_t + 1\}} u_k \kappa(\bbd_k, \cdot) \Big\|_{\ccalH} \\
			&=\min_{\bbu\in\reals^{{M_t}}} \Big\|(1 - \eta \lambda) \!\!\! \!\!\!\sum_{k \in \ccalI \setminus \{M_t + 1\}} \!\!\!  \!\!\!w_k \kappa(\bbd_k, \cdot)- \eta\ell'({f}_t(\bbx_t),\bby_t)\kappa(\bbx_t,\cdot)   -\!\!\! \sum_{k \in \ccalI \setminus \{M_t + 1\}} \!\!\! \!\!\! u_k \kappa(\bbd_k, \cdot)\Big\|_{\ccalH} \; 
			, \nonumber
\end{align}
where we denote the $k^\text{th}$ column of $\bbD_t$ as $\bbd_k$. The minimal error is achieved by considering the square of the expression inside the minimization and expanding terms to obtain
\begin{align}\label{eq:error_expansion}
\Big\|(1 -& \eta \lambda) \!\!\! \!\!\!\sum_{k \in \ccalI \setminus \{M_t + 1\}} \!\!\!  \!\!\!w_k \kappa(\bbd_k, \cdot)- \eta\ell'({f}_t(\bbx_t),\bby_t)\kappa(\bbx_t,\cdot)   -\!\!\! \sum_{k \in \ccalI \setminus \{M_t + 1\}} \!\!\! \!\!\! u_k \kappa(\bbd_k, \cdot)\Big\|_{\ccalH}^2  \\
&= (1\!-\!\eta \lambda\!)^2 \bbw^T \bbK_{\bbD_t, \bbD_t} \bbw 
+ \eta^2 \ell'({f}_t(\bbx_t),\bby_t)^2\kappa(\bbx_t,\bbx_t)
+ \bbu^T \bbK_{\bbD_t, \bbD_t} \bbu  \nonumber \\
&\quad\!-\! 2(1\!-\!\eta \lambda\!) \eta\ell'\!({f}_t(\bbx_t),\bby_t)^2 \bbw^T\! \bbkappa_{\bbD_t}\!(\bbx_t) 
\!+\! 2 \eta^2 \ell'\!({f}_t(\bbx_t),\bby_t\!)\bbu^T\!\bbkappa_{\bbD_t}\!(\bbx_t\!) \!
- \!\!2 (\!1\!-\!\eta \lambda) \bbw^T\! \bbK_{\bbD_t, \bbD_t}\! \bbu  . \nonumber
\end{align}
 To obtain the minimum, we compute the stationary solution of \eqref{eq:error_expansion} with respect to $\bbu \in \reals^{M_t}$ and solve for the minimizing $\tbu^*$, which in a manner similar to the logic in \eqref{eq:proximal_hilbert_representer} - \eqref{eq:hatparam_update}, is given as
 \begin{align}\label{eq:minimal_weights}
 \tbu^* = (1-\eta \lambda) \bbw - \eta \ell'({f}_t(\bbx_t),\bby_t) \bbK_{\bbD_t, \bbD_t}^{-1} \bbkappa_{\bbD_t}(\bbx_t) \; .
\end{align}
Plug $\tbu^*$ in \eqref{eq:minimal_weights} into the expression in \eqref{eq:min_gamma_expand} and  using the short-hand notation ${f}_{t}(\cdot)=\bbw^T \bbkappa_{\bbD_t}(\cdot)$ and $\sum_k u_k \kappa(\bbd_k, \cdot)= \bbu^T \bbkappa_{\bbD_t}(\cdot)$. Doing so simplifies \eqref{eq:min_gamma_expand} to
\begin{align}\label{eq:min_gamma_optimal_weights}
&\Big\| (1-\eta \lambda) \bbw^T\bbkappa_{\bbD_t}(\cdot) - \eta\ell'({f}_t(\bbx_t),\bby_t)\kappa(\bbx_t,\cdot) -  \bbu^T \bbkappa_{\bbD_t}(\cdot) \Big\|_{\ccalH} \\
&\quad =\! \Big\|(1\!-\!\eta\lambda\!)\bbw^T\! \bbkappa_{\bbD_t}(\cdot) \!
 -\! \eta \ell'\!({f}_t(\bbx_t),\bby_t)\kappa(\bbx_t,\!\cdot\!) \! \nonumber \\
&\qquad - [(1\!-\!\eta\lambda)\bbw \!- \!\eta \ell'\!({f}_t(\bbx_t),\bby_t)\bbK_{\bbD_t, \bbD_t}^{-1} \bbkappa_{\bbD_t}(\bbx_t)]^T 
\!\!\bbkappa_{\bbD_t}(\cdot) \! \Big\|_{\ccalH}  .\nonumber
\end{align}
The above expression may be simplified by cancelling like terms $(1-\eta\lambda)\bbw^T\! \bbkappa_{\bbD_t}(\cdot)$ and pulling out a common factor of $\eta |\ell'({f}_t(\bbx_t),\bby_t) |$ outside the norm as
\begin{align}\label{eq:min_gamma_optimal_weights2}
  \Big\| -\eta \ell'({f}_t(\bbx_t),\bby_t) \kappa(\bbx_t,\cdot) 
&  -\eta \ell'({f}_t(\bbx_t),\bby_t) [\bbK_{\bbD_t, \bbD_t}^{-1} \bbkappa_{\bbD_t}(\bbx_t)]^T
   \bbkappa_{\bbD_t}(\cdot) \Big\|_{\ccalH} \nonumber \\
&  = \eta|\ell'({f}_t(\bbx_t),\bby_t) | \Big\|  \kappa(\bbx_t,\cdot) 
  - [\bbK_{\bbD_t, \bbD_t}^{-1} \bbkappa_{\bbD_t}(\bbx_t)]^T
   \bbkappa_{\bbD_t}(\cdot) \Big\|_{\ccalH} \; .
\end{align}
Notice that the right-hand side of \eqref{eq:min_gamma_optimal_weights2} may be identified as the distance to the subspace $\ccalH_{\bbD_t}$ in \eqref{eq:subspace_dist2} defined in Lemma \ref{lemma_subspace_dist} scaled by a factor of $\eta |\ell'({f}_t(\bbx_t),\bby_t) | $. Using this identification, we transform the sufficient condition for the stopping criterion of KOMP to be violated, stated as $\gamma_{M_t + 1} \leq \eps$, into a criterion on $\text{dist}(\kappa(\bbx_t,\cdot),\ccalH_{\bbD_t})$, the subspace distance of $\kappa(\bbx_t,\cdot)$ to the span of kernel evaluations of the current dictionary $\ccalH_{\bbD_t}$.
%
%
Substituting the definition \eqref{eq:subspace_dist2} into $\gamma_{M_t + 1} \leq \eps$ and dividing both sides by $\eta |\ell'({f}_t(\bbx_t),\bby_t) |$ yields
\begin{align}\label{eq:newest_gamma_rescaled}
\text{dist}(\kappa(\bbx_t,\cdot),\ccalH_{\bbD_t}) \leq \frac{\epsilon}{\eta|\ell'(f_t(\bbx_t),\bby_t)|} \; .
\end{align}
Now use the approximation budget selection in terms of the learning rate $\eta$ as $\epsilon=K\eta^{3/2}$. Furthermore, the $C$-Lipschitz continuity of $\ell$ [cf. \eqref{eq:lipschitz}] in Assumption \ref{as:2} allows us to bound the instantaneous gradient by this same constant. Inverting this expression yields $1/|\ell'(f_t(\bbx_t),\bby_t)| \geq 1/C$. Substituting in this lower bound and selection of $\eps$, we obtain that if
\begin{align}\label{eq:newest_gamma_rescaled2}
\text{dist}(\kappa(\bbx_t,\cdot),\ccalH_{\bbD_t}) \leq \frac{K\sqrt{\eta}}{C}
\end{align}
holds, then \eqref{eq:newest_gamma_rescaled} holds, and consequently $M_{t+1} \leq M_t$. 
The contrapositive of the aforementioned logic tells us that growth in the model order ($M_{t+1} = M_t + 1$) implies that the condition
\begin{align}
\label{eq:min_gamma2}
\text{dist}(\kappa(\bbx_t,\cdot),\ccalH_{\bbD_t}) > \frac{K\sqrt{\eta}}{C}
\end{align} 
holds.  Therefore, each time a new point is added to the model, the corresponding kernel function is guaranteed to be at least a distance of $\frac{K\sqrt{\eta}}{C}$ from every other kernel function in the current model, i.e., for distinct dictionary points $\bbd_k$ and $\bbd_j$  for $j,k\in\{1,\dots,M_t\}$, $\|\phi(\bbd_j) - \phi(\bbd_k) \|_2 > \frac{K\sqrt{\eta}}{C}$.  We shall now proceed in a manner similar to the proof of Theorem 3.1 in \cite{1315946}.  Since $\ccalX$ is compact and $\kappa$ is continuous, the range $\phi(\ccalX) $ (where $\phi(\bbx)=\kappa(\bbx,\cdot)$ for $\bbx \in \ccalX$) of the kernel transformation of feature space $\ccalX$ is compact. Therefore, the number of balls of radius $\delta$ (here, $\delta = \frac{K\sqrt{\eta}}{C}$) needed to cover the set $\phi(\ccalX)$ is finite (see, e.g., \cite{anthony2009neural}).  Therefore, for some finite $M^\infty$, if $M_t = M^\infty$, the left-hand side of \eqref{eq:newest_gamma_rescaled2} holds, which implies \eqref{eq:komp_no_terminate} is true for all $t$. We conclude that  $M_{t} \leq M^\infty$ for all $t$.
\end{myproof}



\bibliography{bibliography}

\end{document}